\documentclass[11pt]{article}

\usepackage[final]{acl}

\usepackage{times}
\usepackage{latexsym}
\usepackage{comment}
\usepackage{amsmath}
\usepackage{stfloats}

\usepackage[T1]{fontenc}

\usepackage[utf8]{inputenc}

\usepackage{microtype}

\usepackage{inconsolata}

\usepackage{graphicx}
\usepackage{subcaption}
\usepackage{booktabs}
\graphicspath{{images/}}

\title{Does Item Response Theory Generalize Across Languages? \\Evidence from Cross-Lingual LLM Benchmarks}

\author{Ruchira Dhar$^{\dagger}$ \ \ 
        Qiwei Peng$^{\dagger}$ \ \ 
        \textbf{Anders S{\o}gaard}$^{\dagger}$ \\
        $^{\dagger}$Department of Computer Science, University of Copenhagen \\ 
        }

\begin{document}
\maketitle

\begin{abstract}


The application of Item Response Theory (IRT) for data selection in benchmarks has mostly been limited to monolingual settings. In this work, we ask: how does IRT generalize when the same benchmark is translated across languages? We conduct experiments on Multilingual GSM and Global MMLU, covering over 100 language models and 11 languages.  Fitting 1PL, 2PL, and 3PL IRT models per language, we assess cross-lingual consistency of parameters and whether IRT-filtered subsets preserve model rankings across languages: item \emph{difficulty} emerges as the most stable parameter followed by \emph{guessing} and then \emph{discrimination}. A differential item analysis also shows that lack of stability can be tracked to specific items shifting under translation across languages. Our findings suggest IRT-based selection in multilingual settings must be both parameter-aware and item-aware, validating model choice and screening individual items before assuming cross-lingual generalization.

\end{abstract}

\section{Introduction}

Item Response Theory (IRT), a psychometric framework that characterizes each item in a test with parameters estimated from response patterns across many subjects, has its origins in clinical psychology \cite{harvey1999item} and educational testing \cite{yen2006item, lord2012applications}. In empirical NLP research, IRT has been used as a tool for selecting data points by filtering test-sets for model evaluation since not all evaluations examples turn out to be equally informative \cite{lalor2016building, natesan2016bayesian, lalor2019learning, rodriguez2022clustering}. More recently, it has also been shown that fitting IRT models to LLM responses, we can select data points while preserving ranking validity of the full benchmark \cite{hofmann2025fluid, zhou2026lost}. However, the vast majority of this work has been conducted in monolingual settings, primarily English (see \autoref{app:related} for related work). Today a key focus of NLP is cross-lingual evaluation \cite{hu2020xtreme, doddapaneni2025cross, chollampatt2025cross} where translation is a popular strategy for constructing multilingual benchmarks \cite{shi2023language, singh2025global}. This raises a natural but largely untested concern: how does IRT generalize cross-lingually when the same benchmark is translated across languages? To our knowledge, this is the first systematic study to attempt to address this.

We evaluate 102 LMs on MGSM \cite{shi2023language} (11 languages giving 11 slices) and 103 LMs on Global MMLU \cite{singh2025global} (7 languages, 6 domains giving 42 slices), yielding over 568,000 item-response pairs across 53 benchmark slices. We fit three IRT variants with incrementally more parameters and increasing complexity: 1PL (\textit{difficulty} only), 2PL (adds \textit{discrimination}), and 3PL (adds \textit{guessing}), independently to each benchmark slice, yielding 159 fitted models.
We ask three questions: 1) are IRT parameters consistent across translated versions of a benchmark; 2) do IRT-filtered subsets preserve model rankings across languages?; and 3) can cross-lingual instability in IRT parameters be traced to specific items rather than treated as a language-wide effect?  Our results indicate that item \emph{difficulty} shows highest concordance across languages for all IRT models, and \emph{guessing} performs better than \emph{discrimination}. This parameter hierarchy directly determines subset validity: 1PL is the most reliable cross-lingual variant. A differential item analysis also confirms this instability is item-level and not language-wide effects. We further observe that for MMLU, scientific domains are consistently more concordant across languages as compared to non-science domains. 

Taken together, our results indicate that  IRT does not generalise uniformly across languages: which parameters and subsets transfer depends on the IRT model, the individual item, and the domain of evaluation.


\section{Methodology} 
\label{sec:methodology}


\subsection{LLM Evaluations}

We use two multilingual translated benchmarks: Multilingual GSM (MGSM) \cite{shi2023language} across 11 languages (\texttt{bn}, \texttt{de}, \texttt{en}, \texttt{es}, \texttt{fr}, \texttt{ja}, \texttt{ru},
\texttt{sw}, \texttt{te}, \texttt{th}, \texttt{zh}), yielding  \textbf{250 } items per language and \textbf{2{,}750} items per model; Global MMLU \cite{singh2025global} across 7 languages (\texttt{de}, \texttt{en}, \texttt{es},
\texttt{fr}, \texttt{ja}, \texttt{sw}, \texttt{zh}) and 6 subject domains
(\textit{business}, \textit{humanities}, \textit{medical}, \textit{other},
\textit{social sciences}, \textit{stem}), totaling \textbf{400 }items per language and \textbf{2{,}800 }items per model under the standard 7-language configuration. For models, we use the HuggingFace Open LLM Leaderboard to select the top instruction-tuned models under 20 billion parameters that are compatible with the \verb|lm-eval-harness| framework \cite{eval-harness}, yielding 102 and 103 compatible models for MGSM and Global MMLU respectively. In total, our corpus comprises \textbf{568{,}900} evaluated item--response pairs, each consisting of the model's generated response to a benchmark question and a binary correctness label (implementation details in \autoref{app:techdetails}). These item--response pairs are used for IRT model training and for all downstream analysis\footnote{The code is publicly available at
\href{https://github.com/ruchiradhar/benchfilter}{this GitHub Repo}.}.

\subsection{IRT Model Training}

The response data, consisting of binary correctness indicators $X_{ij} \in \{0, 1\}$ representing whether a specific language model $i$ answered a benchmark item $j$ correctly, are organised into one response matrix per language (for MGSM) and one per language--domain combination (for Global MMLU), yielding 53 matrices in total (11 MGSM $\times$ 1 + 7 MMLU languages $\times$ 6 domains). We fit three IRT model variants to each matrix independently (implementation details in \autoref{app:techdetails}). For 1PL, we use Stochastic Variational Inference (SVI) as implemented in \texttt{py-irt} \cite{natesan2016bayesian} on top of \texttt{Pyro} \cite{bingham2019pyro}. For 2PL and 3PL, we use MAP-based implementation in \texttt{PyTorch} \footnote{We retain \texttt{py-irt}'s SVI for 1PL, where SVI and MAP estimates are highly concordant, confirming inference method does not affect results. For 2PL and 3PL, SVI's prior dominates gradient updates at LLM-scale and MAP avoids this. See \autoref{app:svi_vs_map} the full comparison and justification.}. We fit the following IRT model variants:

\begin{figure*}[t]
  \centering
  \begin{subfigure}[t]{0.32\linewidth}
    \includegraphics[width=\linewidth]{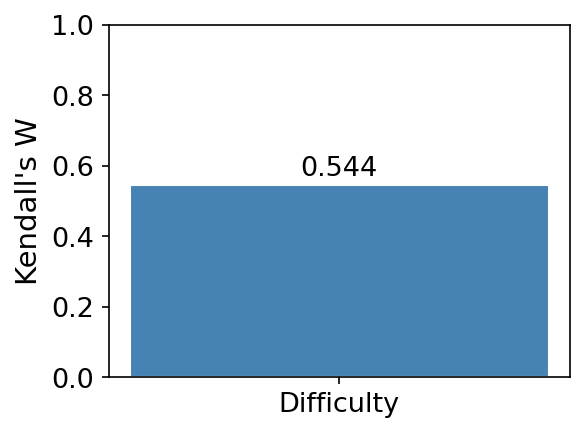}
    \caption{MGSM 1PL}\label{fig:gsm_1pl_kw}
  \end{subfigure}\hfill
  \begin{subfigure}[t]{0.32\linewidth}
    \includegraphics[width=\linewidth]{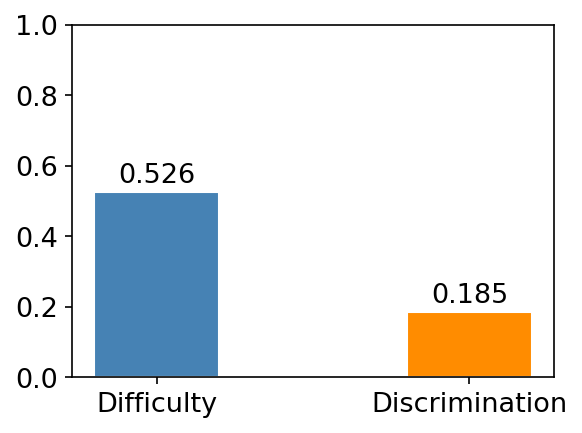}
    \caption{MGSM 2PL}\label{fig:gsm_2pl_kw}
  \end{subfigure}\hfill
  \begin{subfigure}[t]{0.32\linewidth}
    \includegraphics[width=\linewidth]{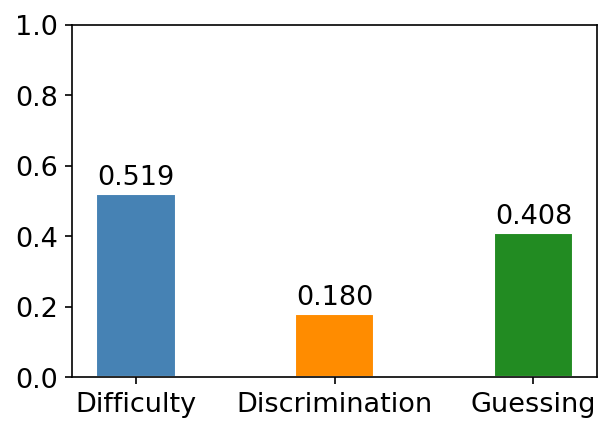}
    \caption{MGSM 3PL}\label{fig:gsm_3pl_kw}
  \end{subfigure}
  \begin{subfigure}[t]{0.32\linewidth}
    \includegraphics[width=\linewidth]{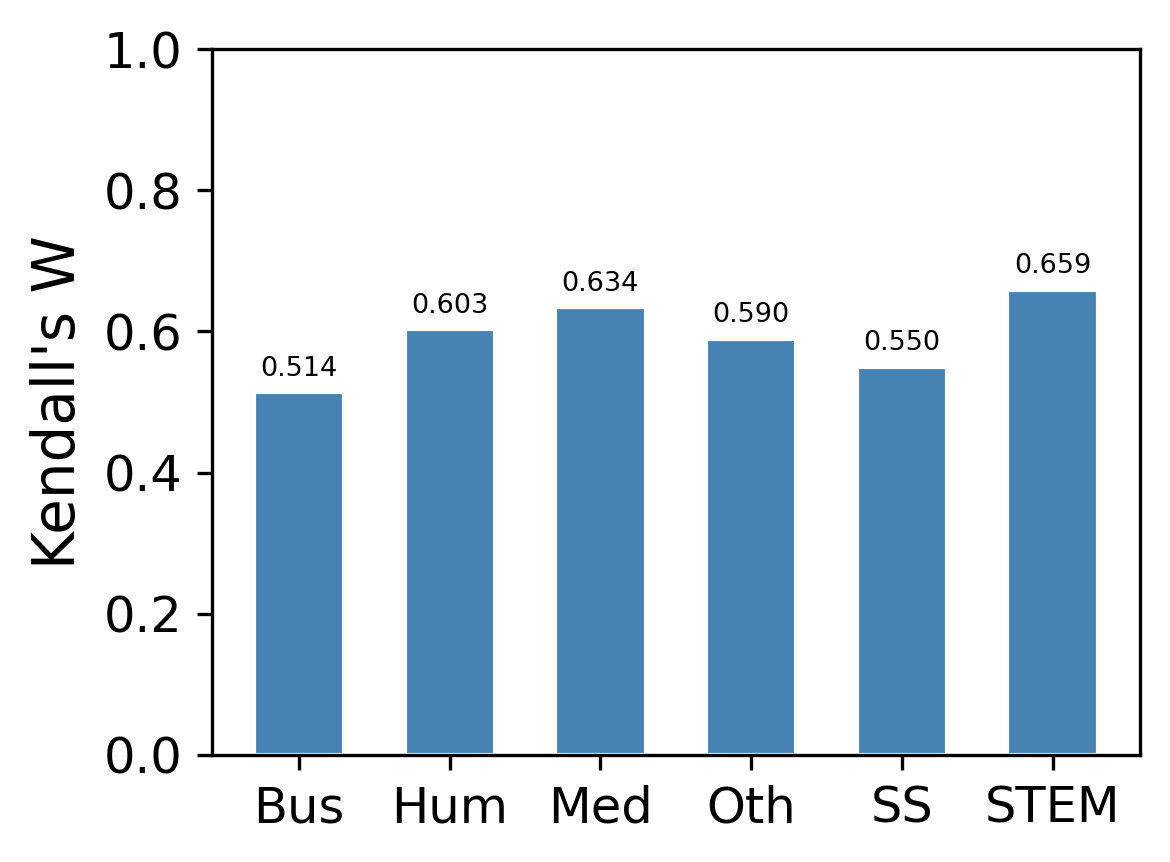}
    \caption{MMLU 1PL}\label{fig:mmlu_1pl_kw}
  \end{subfigure}\hfill
  \begin{subfigure}[t]{0.32\linewidth}
    \includegraphics[width=\linewidth]{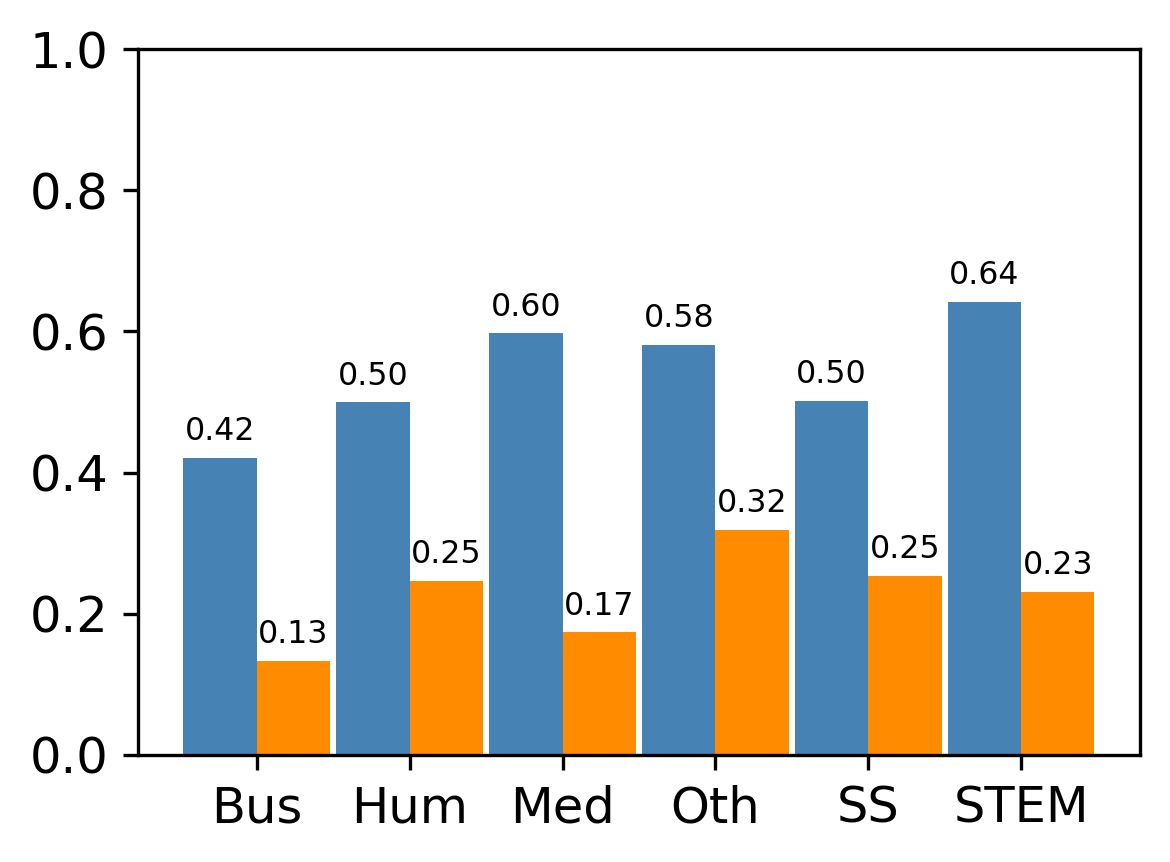}
    \caption{MMLU 2PL}\label{fig:mmlu_2pl_kw}
  \end{subfigure}\hfill
  \begin{subfigure}[t]{0.32\linewidth}
    \includegraphics[width=\linewidth]{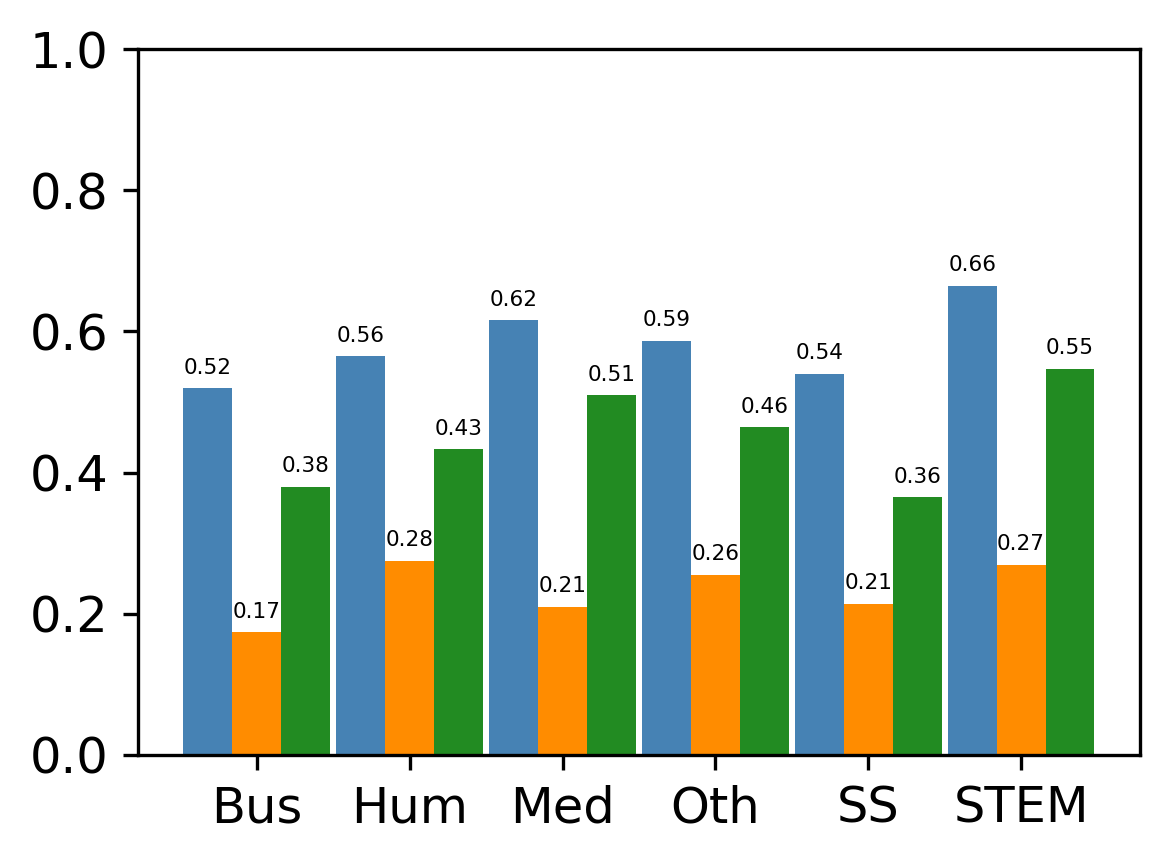}
    \caption{MMLU 3PL}\label{fig:mmlu_3pl_kw}
  \end{subfigure}
  \caption{Kendall's $W$ concordance across languages for MGSM (top row) and MMLU
  (bottom row) under 1PL, 2PL, and 3PL models. For MGSM, each bar is one parameter type; for MMLU, each bar is one domain (Bus: Business, Hum: Humanities, Med: Medical, SS: Social Sciences, Oth: Others).}
  \label{fig:kendalls_w}
\end{figure*}

\begin{itemize}

  \item \textbf{1PL (Rasch model).}
  The response probability is
  $P(X_{ij}=1) = \sigma(\theta_i - b_j)$,
  where $\theta_i$ is the ability of model $i$ and $b_j$ is the difficulty of item $j$. 

  \item \textbf{2PL.}
  The response probability is
  $P(X_{ij}=1) = \sigma\!\left(a_j(\theta_i - b_j)\right)$,
  adding a per-item discrimination parameter $a_j > 0$ that scales how sharply the
  item distinguishes between high- and low-ability models.

  \item \textbf{3PL.}
  The response probability is
  $P(X_{ij}=1) = c_j + (1 - c_j)\,\sigma\!\left(a_j(\theta_i - b_j)\right)$,
  where $c_j \in [0,1]$ is a lower asymptote (guessing parameter) that
  models the probability of a correct response by a minimally-capable model, capturing
  chance-level or knowledge-independent success on an item.

\end{itemize}

\noindent

For 1PL, we use \texttt{py-irt}'s SVI with dropout ($p=0.5$), vague priors, and learning rate $\eta=0.1$, training for 2{,}000 epochs (selected at the point of ELBO stabilisation) and retaining parameters at the minimum-ELBO epoch. For 2PL and 3PL, we use MAP with Adam ($\eta=0.05$) for 4{,}000 and 6{,}000 epochs respectively (selected at the point of MAP objective stabilisation, with longer schedules reflecting increased model complexity), retaining the best-objective state across training.

\subsection{IRT Analysis}

We conduct three sets of experiments to assess how IRT parameters generalize across language versions of these benchmarks .

\textbf{Cross-lingual Consistency Analysis} (§\ref{sec:cross_lingual}) examines whether item parameters are stable across languages using four measures\footnote{All IRT variants achieve monotonically improving within-language fit across majority of slices (3PL $>$ 2PL $>$ 1PL; \autoref{app:withinlangfits}), suggesting that cross-lingual parameter differences in following analyses are not merely artifacts of poor within-language fit.}:
(i)~Kendall's $W \in [0,1]$, the coefficient of concordance across all $|\mathcal{L}|$ languages simultaneously ($W=1$: identical item rankings; $W\approx 0$: chance);
(ii)~Spearman Correlations $\rho \in [-1,1]$, the monotonic rank correlation between each language pair, displayed as a $|\mathcal{L}|\times|\mathcal{L}|$ heatmap;
(iii)~Per-Item Rank Deviation, the standard deviation of each item's difficulty rank across languages, where high values flag unstable items;
(iv)~Language Clustering (average linkage on Spearman distance $1-\rho$) grouping languages by similarity of their IRT profiles, with alignment to accuracy-based clusters quantified via Mantel test.

\textbf{Subset Analysis} (§\ref{sec:iif}) evaluates IRT filtering along two dimensions. (i)~ Cross-Model IIF Concordance: items are ranked by their Item Information Function (IIF) at $\theta=0$ under each IRT model variant, and pairwise Jaccard similarity of the top-50\% subsets is measured across benchmark slices to assess whether model choice affects item selection; (ii)~Subset Ranking Fidelity: Spearman $\rho$ between model accuracy on the IRT-selected subset and on the full benchmark, evaluated within-language and cross-lingually under in-sample and held-out (80/20) model splits.

\textbf{DIF Analysis} (§\ref{sec:dif}) asks whether cross-lingual instability in IRT parameters is a language-wide effect or is concentrated in specific items.  Thus, for each item we ask: once we account for the fact that some languages are simply harder overall, does this specific item still get noticeably harder or easier when translated? An item where the answer is yes is exhibiting \emph{differential item functioning} (DIF) and language-specific bias in how the item behaves, independent of overall difficulty \cite{holland2012differential, allalouf1999identifying}. To test this directly, we conduct an anchor-item DIF analysis relative to English. For each language, the bottom quartile of items by cross-lingual rank standard deviation serves as an anchor set to equate the 1PL difficulty scale between English and each focal language, avoiding the equal-ability assumption that otherwise blocks a formal DIF test in our setting. For each remaining item we compute the anchor-corrected difficulty shift $\Delta b_j$, test significance via Fisher-information standard errors, and control the false discovery rate at $q=0.05$ with Benjamini--Hochberg.

\section{Results}
\label{sec:results}

\subsection{Cross-Lingual Consistency Analysis}
\label{sec:cross_lingual}

\paragraph{Kendall's $W$ Concordance}

\begin{table}[t]
  \centering
  \small
  \begin{tabular}{lcccc}
    \toprule
    & \multicolumn{2}{c}{\textbf{Science}} & \multicolumn{2}{c}{\textbf{Non-science}} \\
    \cmidrule(lr){2-3}\cmidrule(lr){4-5}
    \textbf{Model} & $W$ & $\bar{\rho}$ & $W$ & $\bar{\rho}$ \\
    \midrule
    1PL & 0.647 & 0.588 & 0.564 & 0.492 \\
    2PL & 0.620 & 0.559 & 0.501 & 0.420 \\
    3PL & 0.640 & 0.581 & 0.553 & 0.479 \\
    \bottomrule
  \end{tabular}
  \caption{ Kendall's $W$ and Mean Pairwise Spearman $\bar{\rho}$ for item
    \emph{difficulty} across languages, averaged over science (STEM + Medical)
    and non-science (Business + Humanities + Other + Social Sciences) domains
    of MMLU, for each IRT model variant.}
  \label{tab:science_vs_nonsci}
\end{table}

\autoref{fig:kendalls_w} shows that difficulty concordance is maintained across model complexities, while discrimination parameters are substantially less cross-lingually stable. Under 1PL, difficulty concordances are: MGSM at $W = 0.544$ and MMLU domains ranging from $W \in [0.514, 0.659]$. Under 2PL, difficulty concordance is maintained: MGSM $W_\text{diff} = 0.526$ (barely below 1PL), MMLU $W_\text{diff} \in [0.421, 0.642]$. Under 3PL, difficulty concordance is similarly preserved (MGSM $W_\text{diff} = 0.519$; MMLU $W_\text{diff} \in [0.520, 0.664]$). The discrimination parameter, by contrast, shows substantially lower cross-lingual agreement across both 2PL (MGSM $W_\text{disc} = 0.185$; MMLU $W_\text{disc} \in [0.134, 0.319]$) and 3PL (MGSM $W_\text{disc} = 0.180$; MMLU $W_\text{disc} \in [0.174, 0.275]$). The guessing parameter $c$ under 3PL retains concordance on both MGSM ($W = 0.408$) and MMLU ($W \in [0.365, 0.547]$). To rule out that this concordance is inflated by rater redundancy among typologically similar languages, we recompute $W$ after removing the tight cluster of Indo-European languages and find that concordance does not decrease under any IRT variant (details in \autoref{app:correlatedraters}).

A cross-domain analysis (\autoref{tab:science_vs_nonsci}) of item \textit{difficulty} groups MMLU domains into science (STEM + Medical) and non-science (Business + Humanities + Other + Social Sciences), revealing a persistent science advantage that holds across all model variants. 
Scientific domains, where mathematical and factual content is more likely to be language-invariant, yield more consistent IRT parameters across languages.


\begin{figure}[t]
  \centering
  \begin{subfigure}[t]{0.32\linewidth}
    \includegraphics[width=\linewidth]{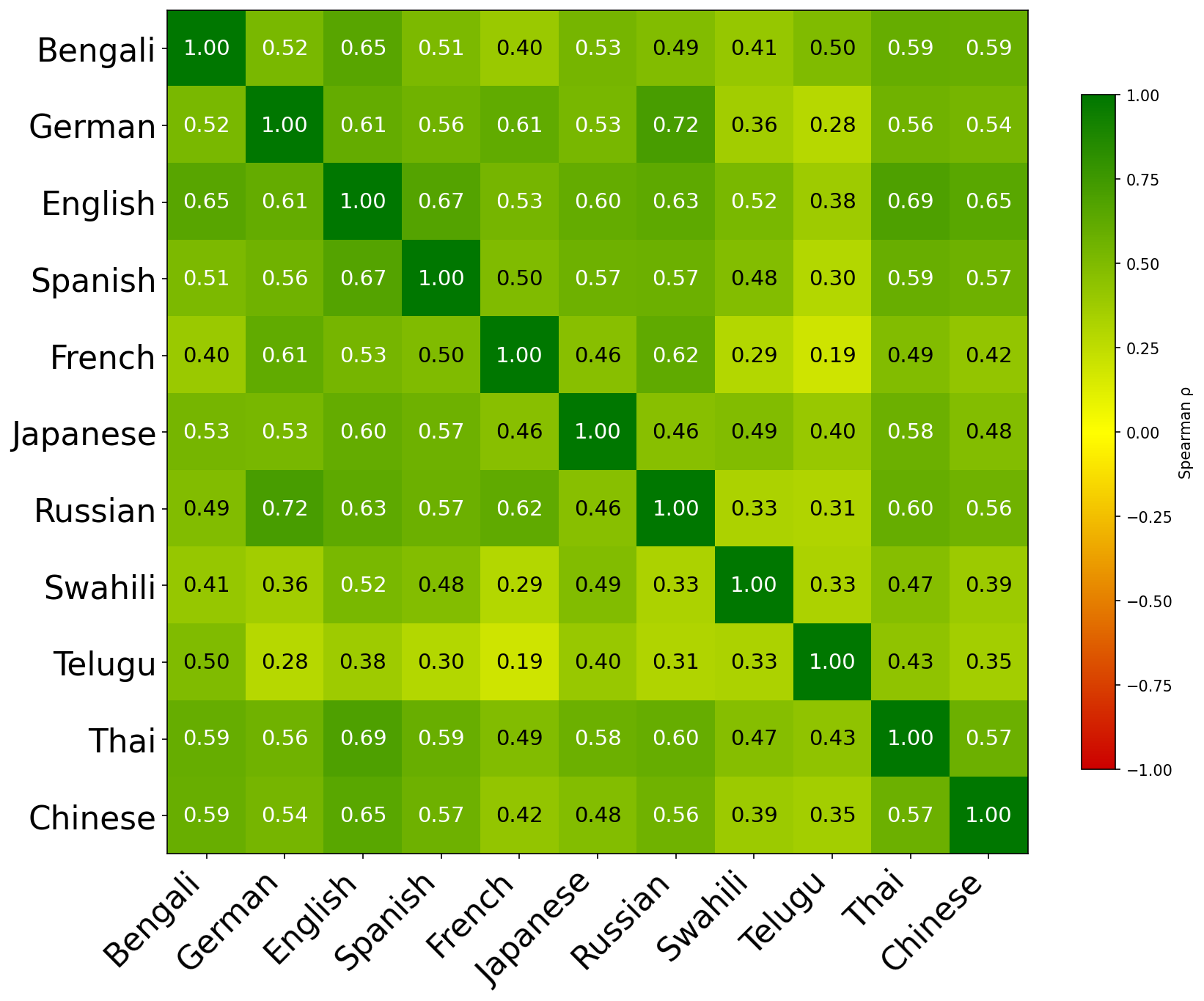}
    \caption{MGSM 1PL}\label{fig:gsm_1pl_sp}
  \end{subfigure}\hfill
  \begin{subfigure}[t]{0.32\linewidth}
    \includegraphics[width=\linewidth]{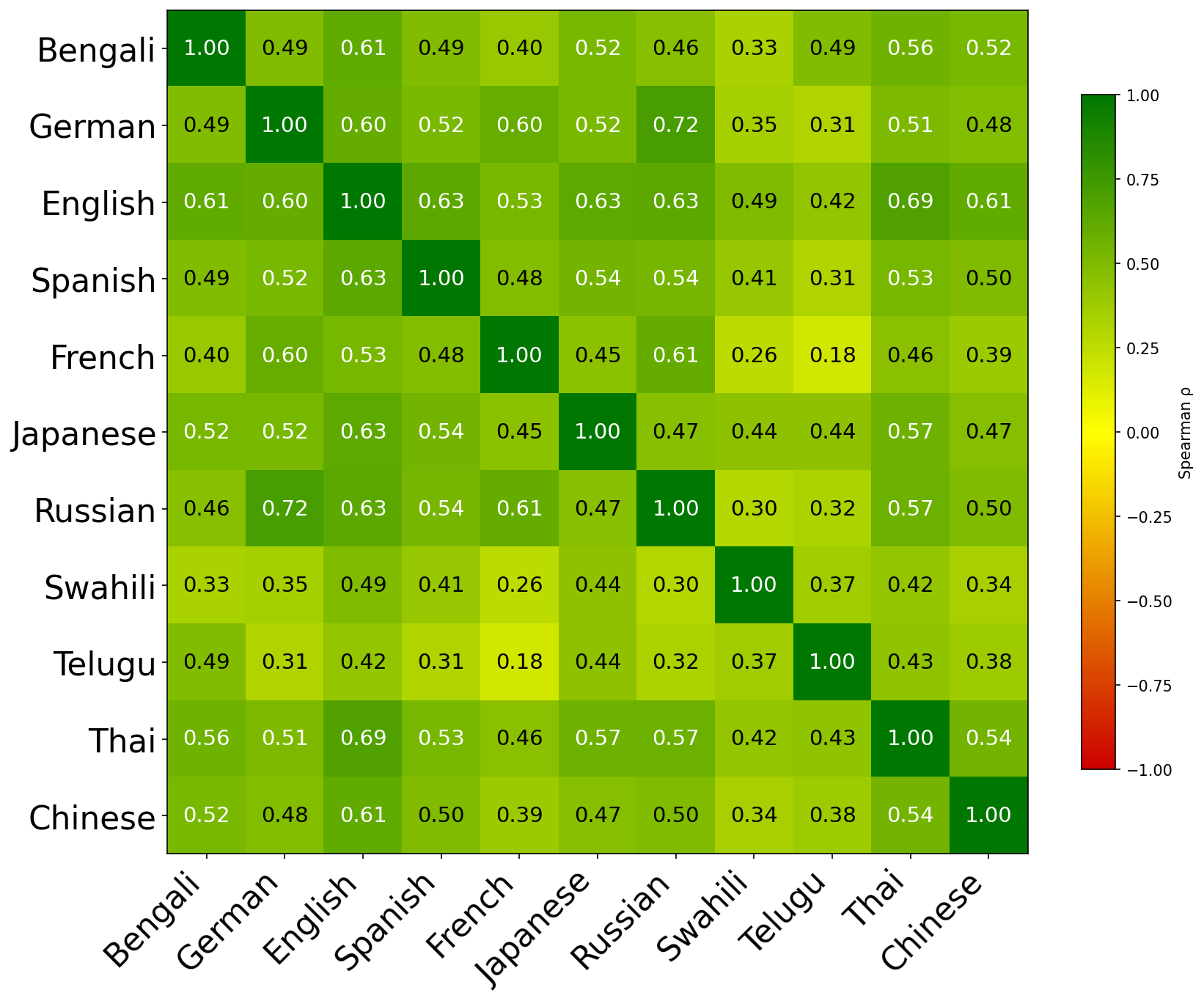}
    \caption{MGSM 2PL}\label{fig:gsm_2pl_sp}
  \end{subfigure}\hfill
  \begin{subfigure}[t]{0.32\linewidth}
    \includegraphics[width=\linewidth]{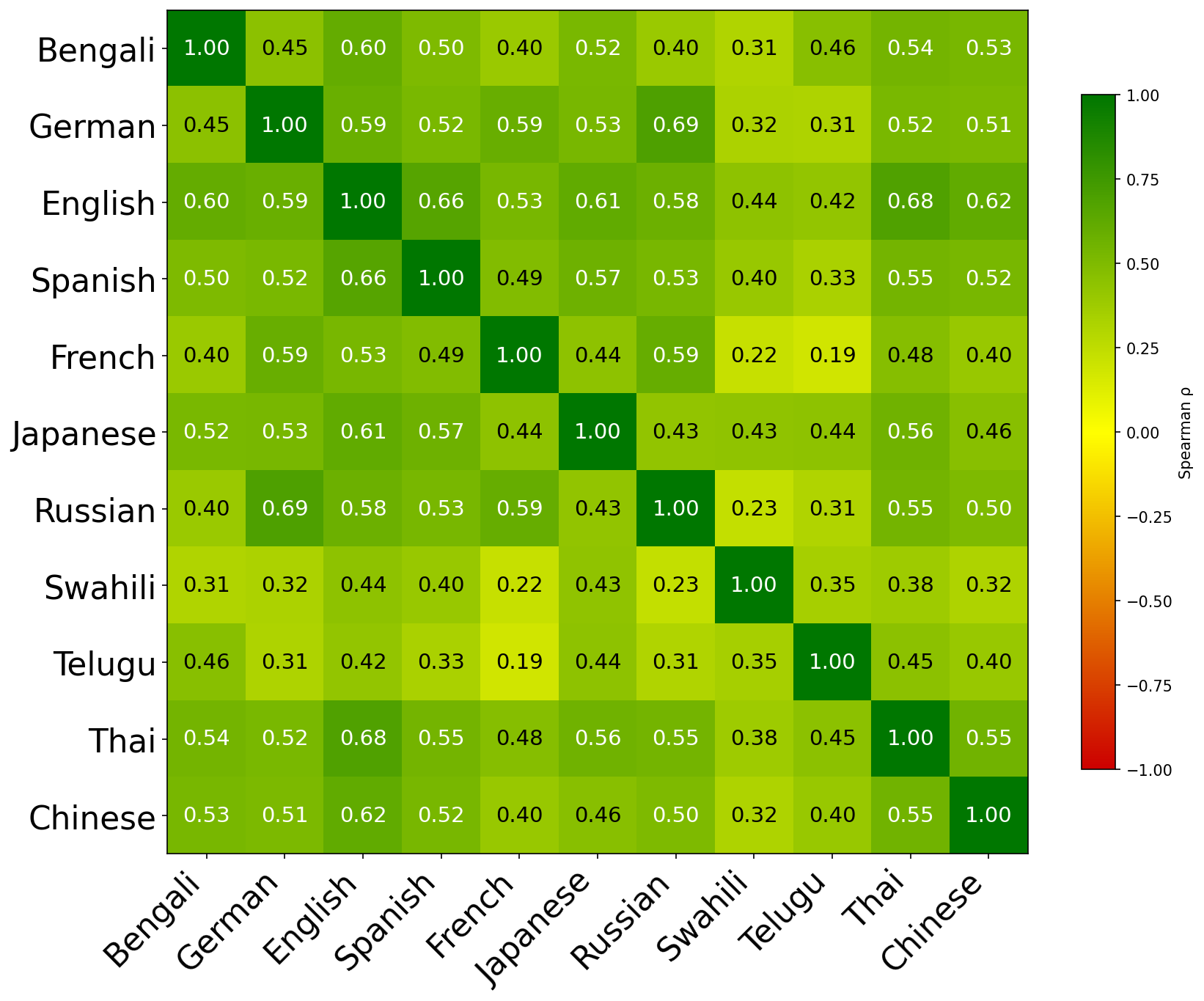}
    \caption{MGSM 3PL}\label{fig:gsm_3pl_sp}
  \end{subfigure}
  \begin{subfigure}[t]{0.32\linewidth}
    \includegraphics[width=\linewidth]{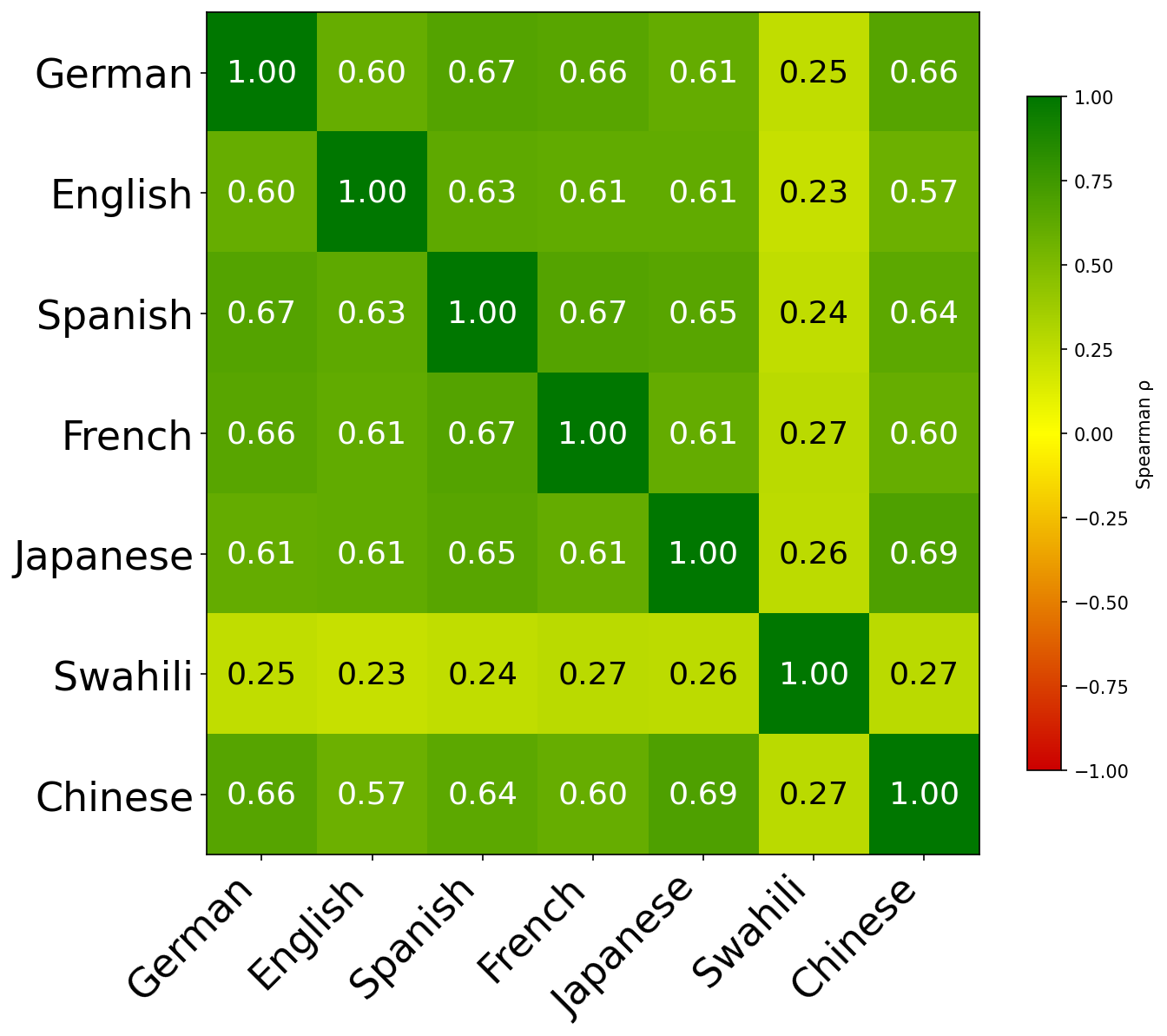}
    \caption{MMLU 1PL}\label{fig:mmlu_1pl_sp}
  \end{subfigure}\hfill
  \begin{subfigure}[t]{0.32\linewidth}
    \includegraphics[width=\linewidth]{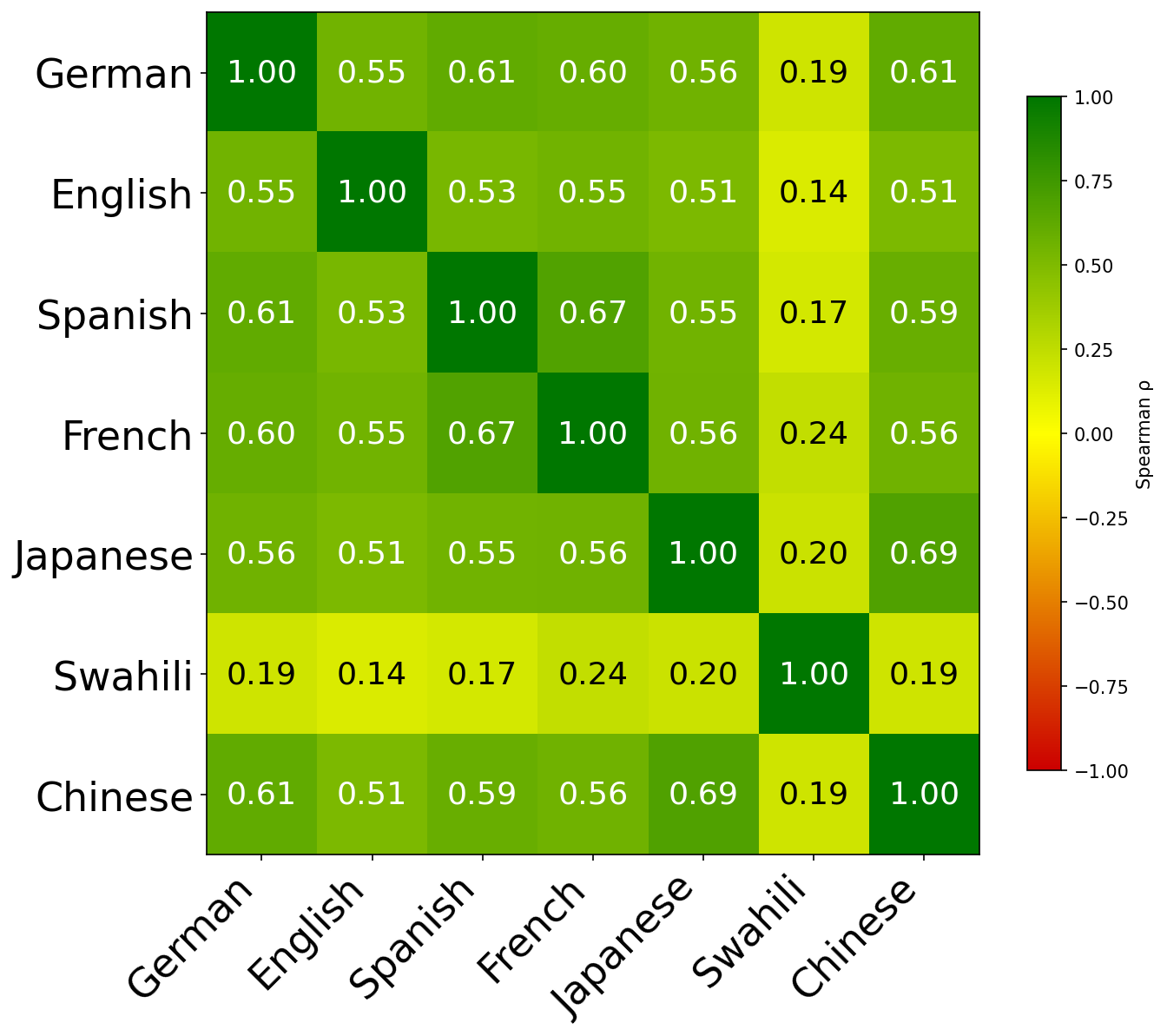}
    \caption{MMLU 2PL}\label{fig:mmlu_2pl_sp}
  \end{subfigure}\hfill
  \begin{subfigure}[t]{0.32\linewidth}
    \includegraphics[width=\linewidth]{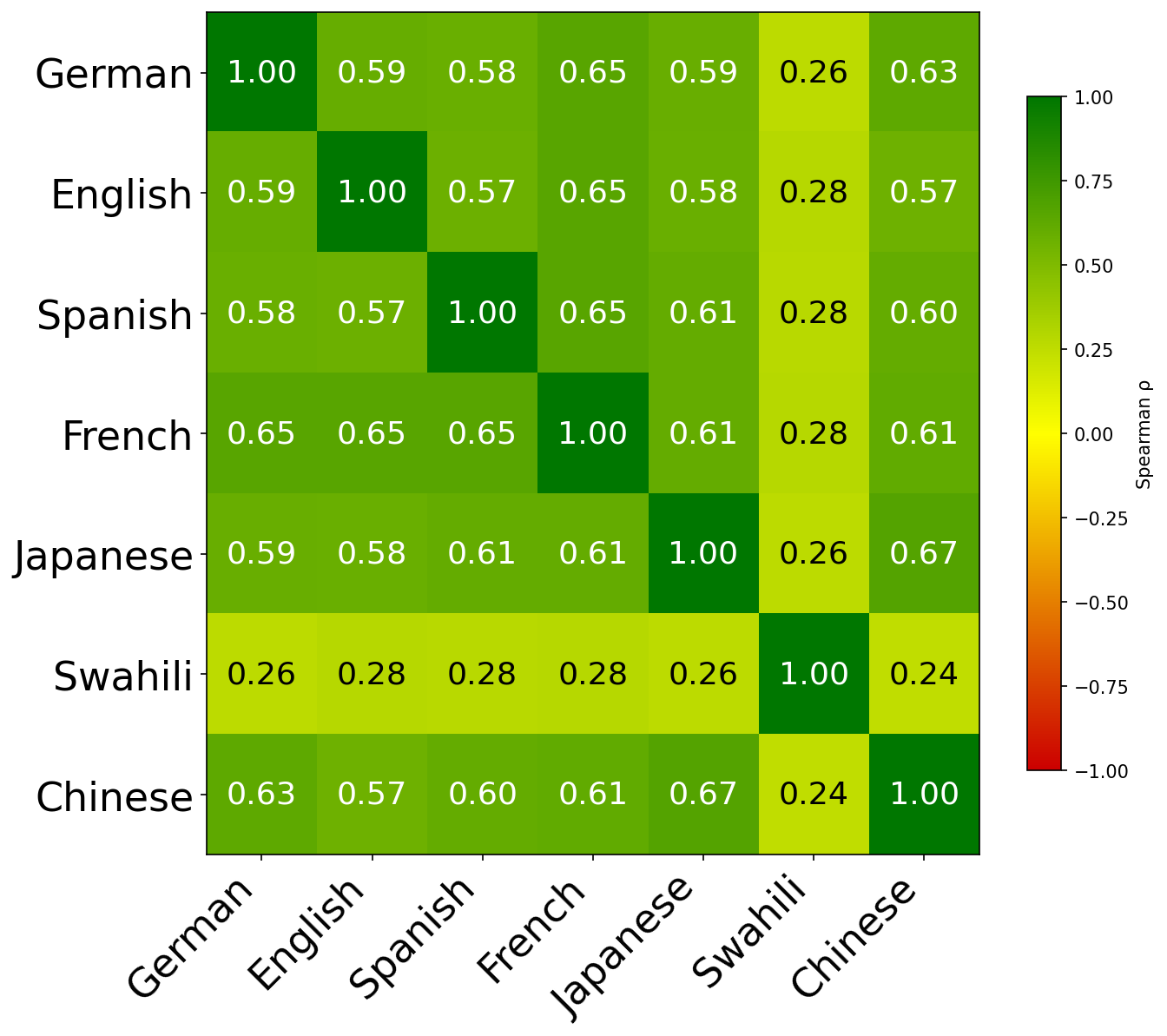}
    \caption{MMLU 3PL}\label{fig:mmlu_3pl_sp}
  \end{subfigure}
  \caption{Pairwise Spearman rank correlation of item \emph{difficulty} across
    languages for 1PL, 2PL, and 3PL IRT models on MGSM (top row) and MMLU
    (bottom row; each cell is the element-wise average over six domains).
    Colour encodes $\rho \in [-1, +1]$: red $= -1$, yellow $= 0$, green $= +1$.}
  \label{fig:spearman_heatmaps}
\end{figure}

\begin{figure}[t]
  \centering
  \begin{subfigure}[t]{0.32\linewidth}
    \includegraphics[width=\linewidth]{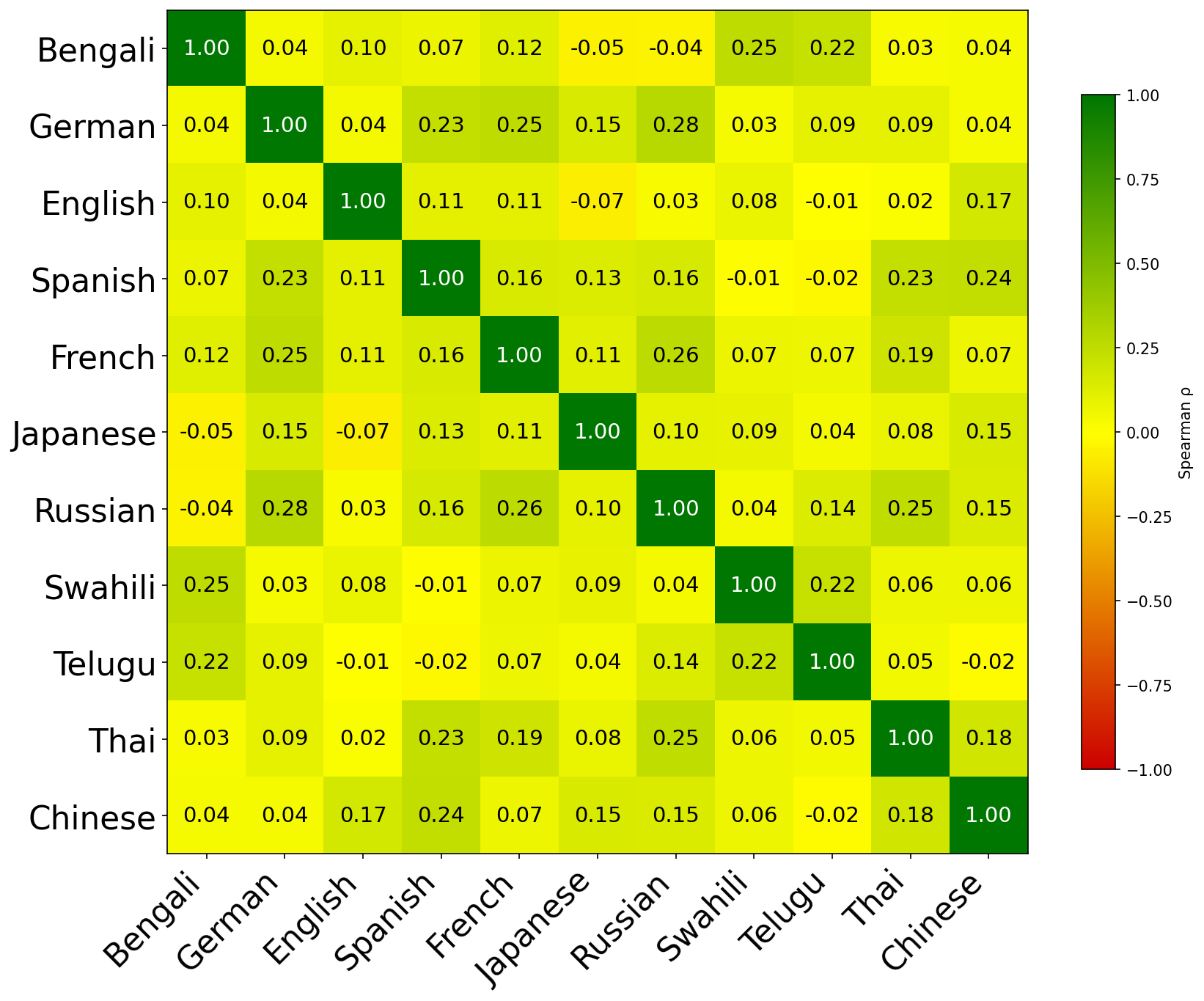}
    \caption{MGSM 2PL}\label{fig:gsm_2pl_disc}
  \end{subfigure}\hfill
  \begin{subfigure}[t]{0.32\linewidth}
    \includegraphics[width=\linewidth]{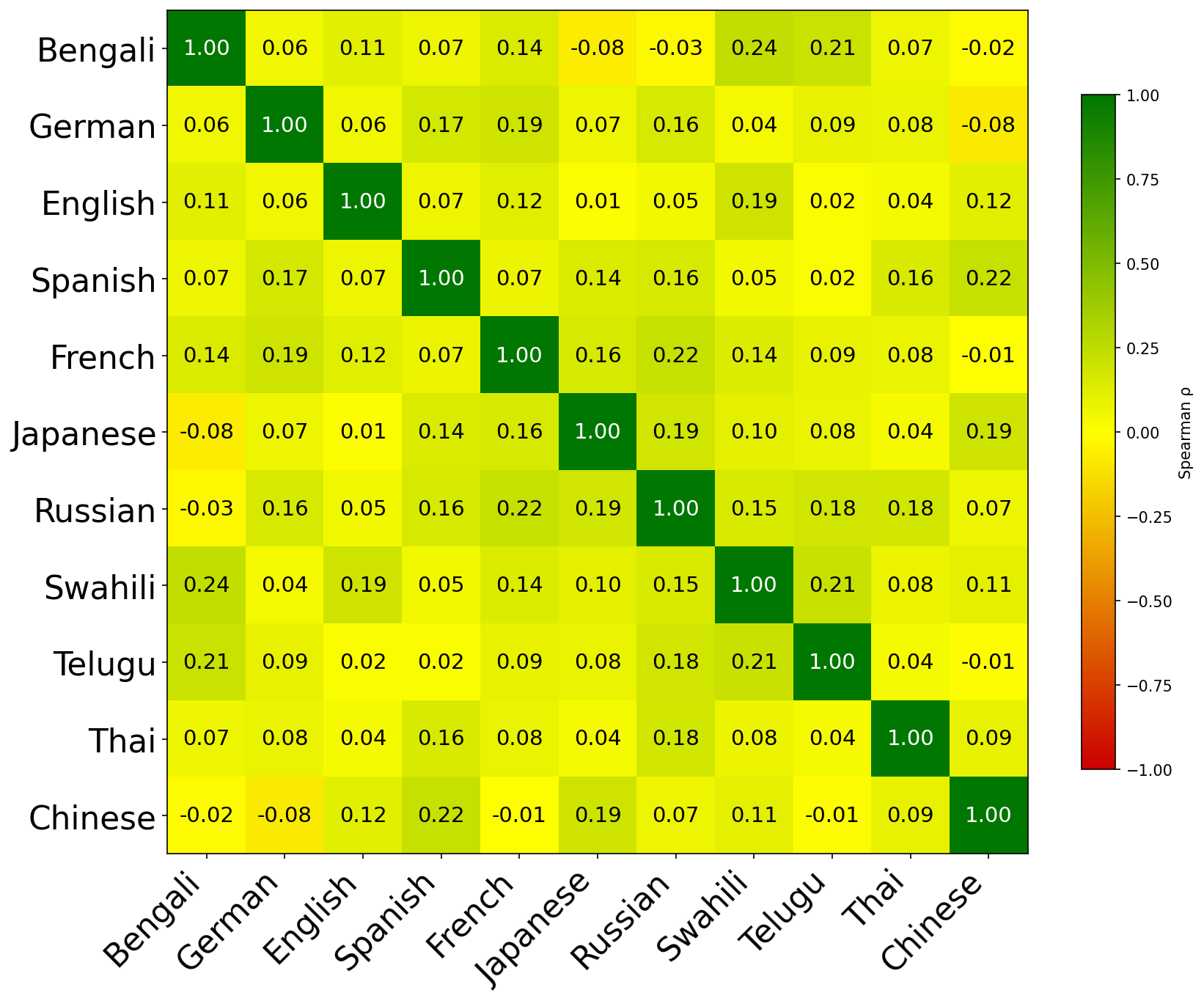}
    \caption{MGSM 3PL}\label{fig:gsm_3pl_disc}
  \end{subfigure}\hfill
  \begin{subfigure}[t]{0.32\linewidth}
    \includegraphics[width=\linewidth]{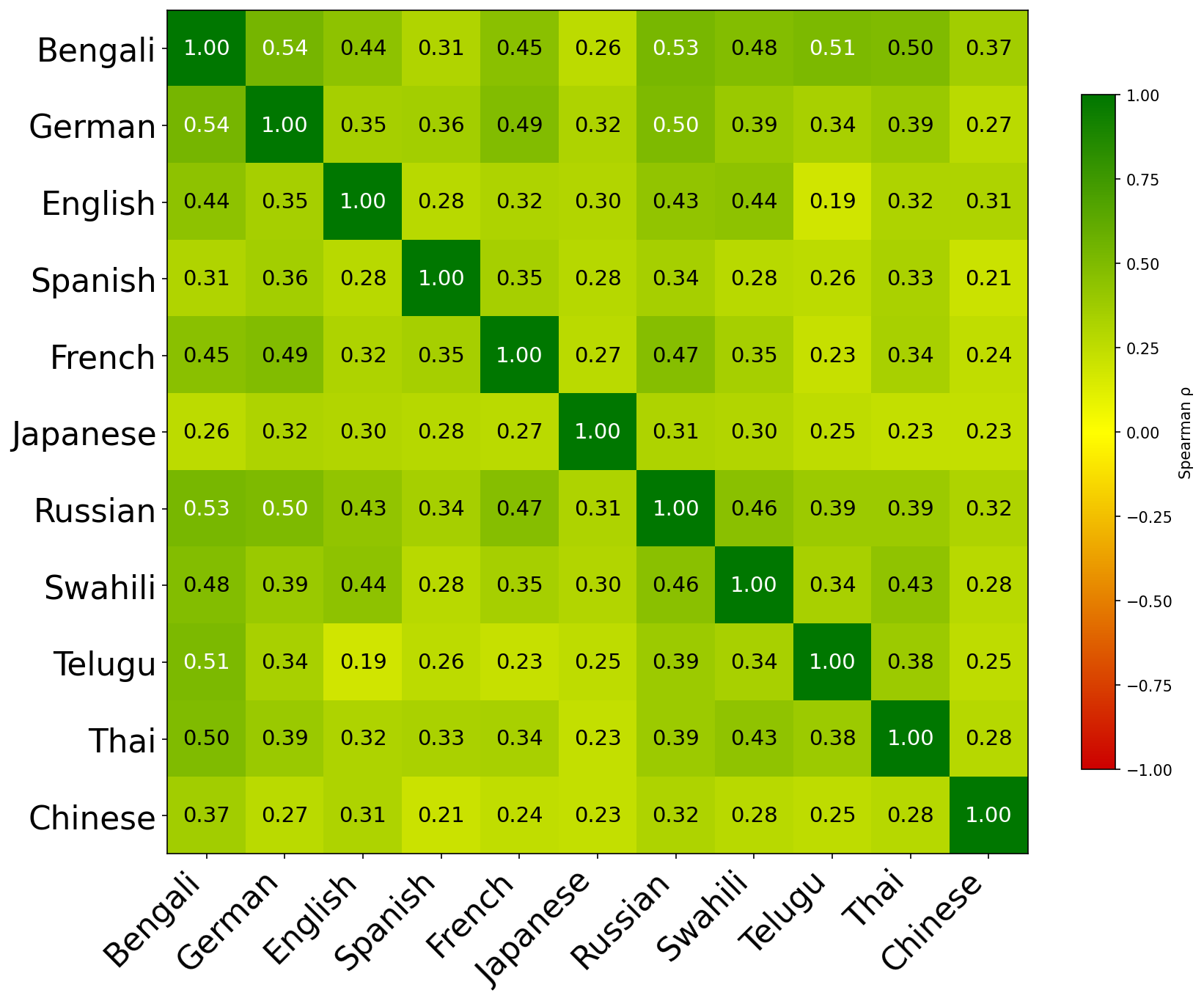}
    \caption{MGSM 3PL}\label{fig:gsm_3pl_guess}
  \end{subfigure}
  \begin{subfigure}[t]{0.32\linewidth}
    \includegraphics[width=\linewidth]{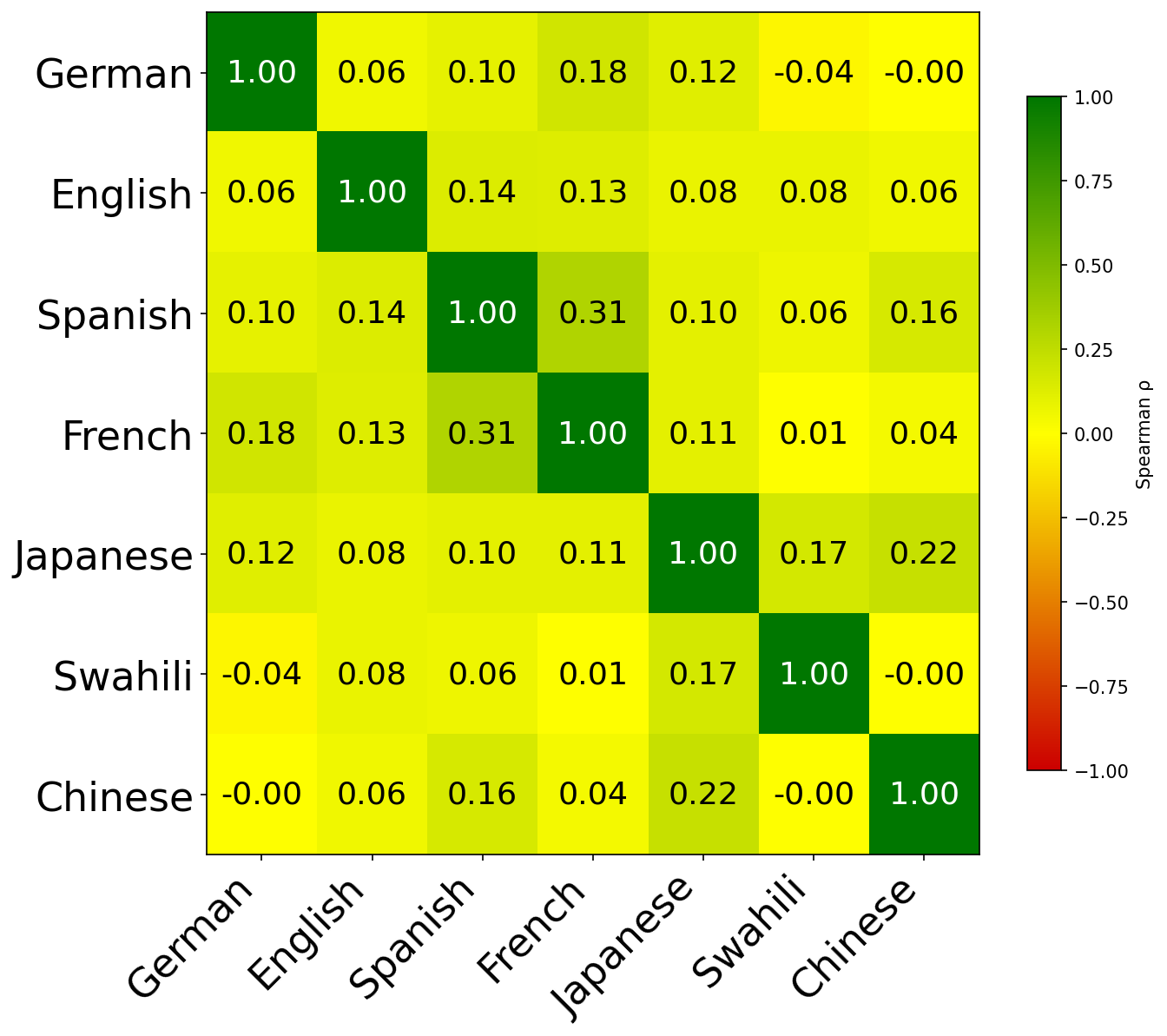}
    \caption{MMLU 2PL }\label{fig:mmlu_2pl_disc}
  \end{subfigure}\hfill
  \begin{subfigure}[t]{0.32\linewidth}
    \includegraphics[width=\linewidth]{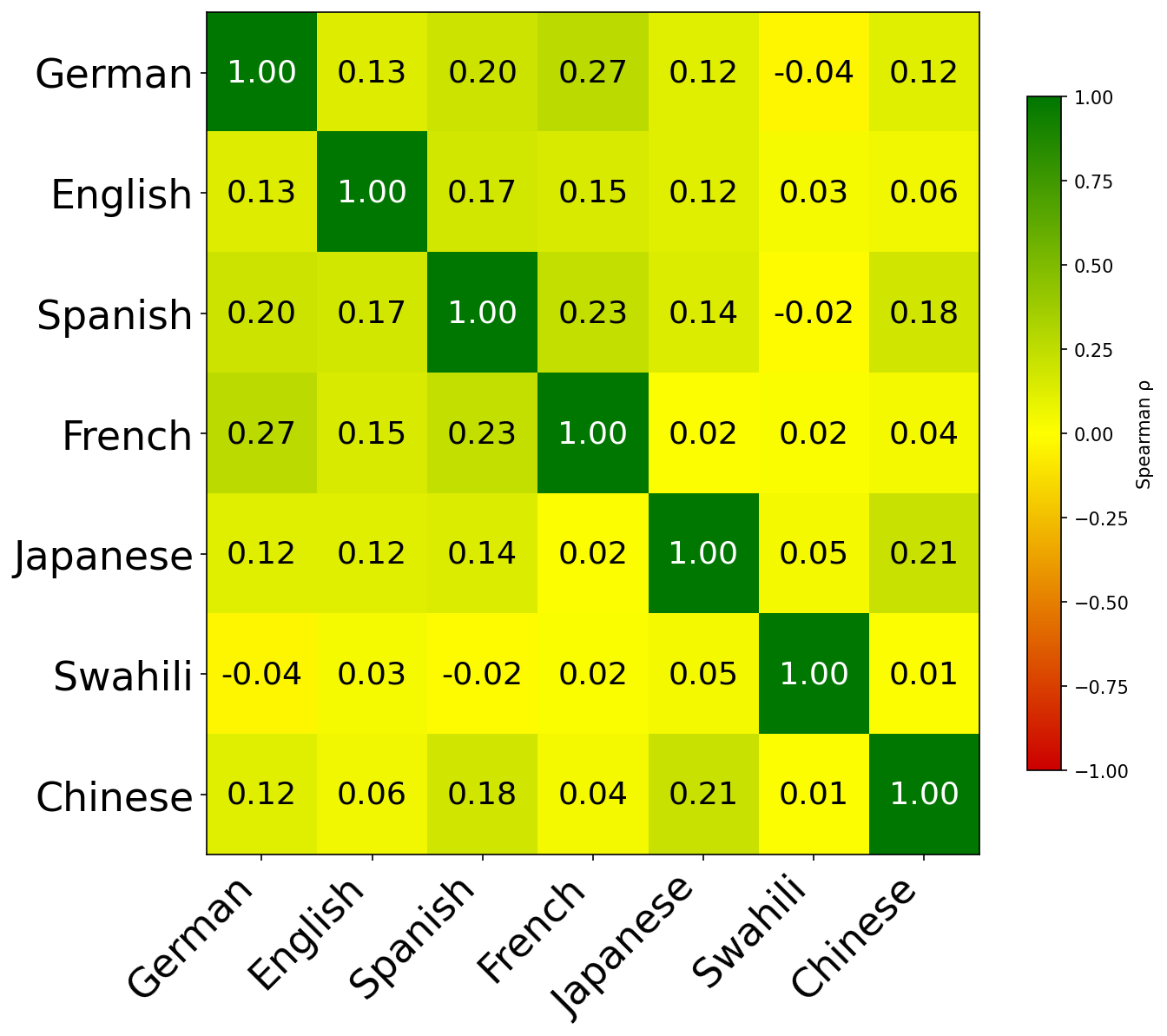}
    \caption{MMLU 3PL }\label{fig:mmlu_3pl_disc}
  \end{subfigure}\hfill
  \begin{subfigure}[t]{0.32\linewidth}
    \includegraphics[width=\linewidth]{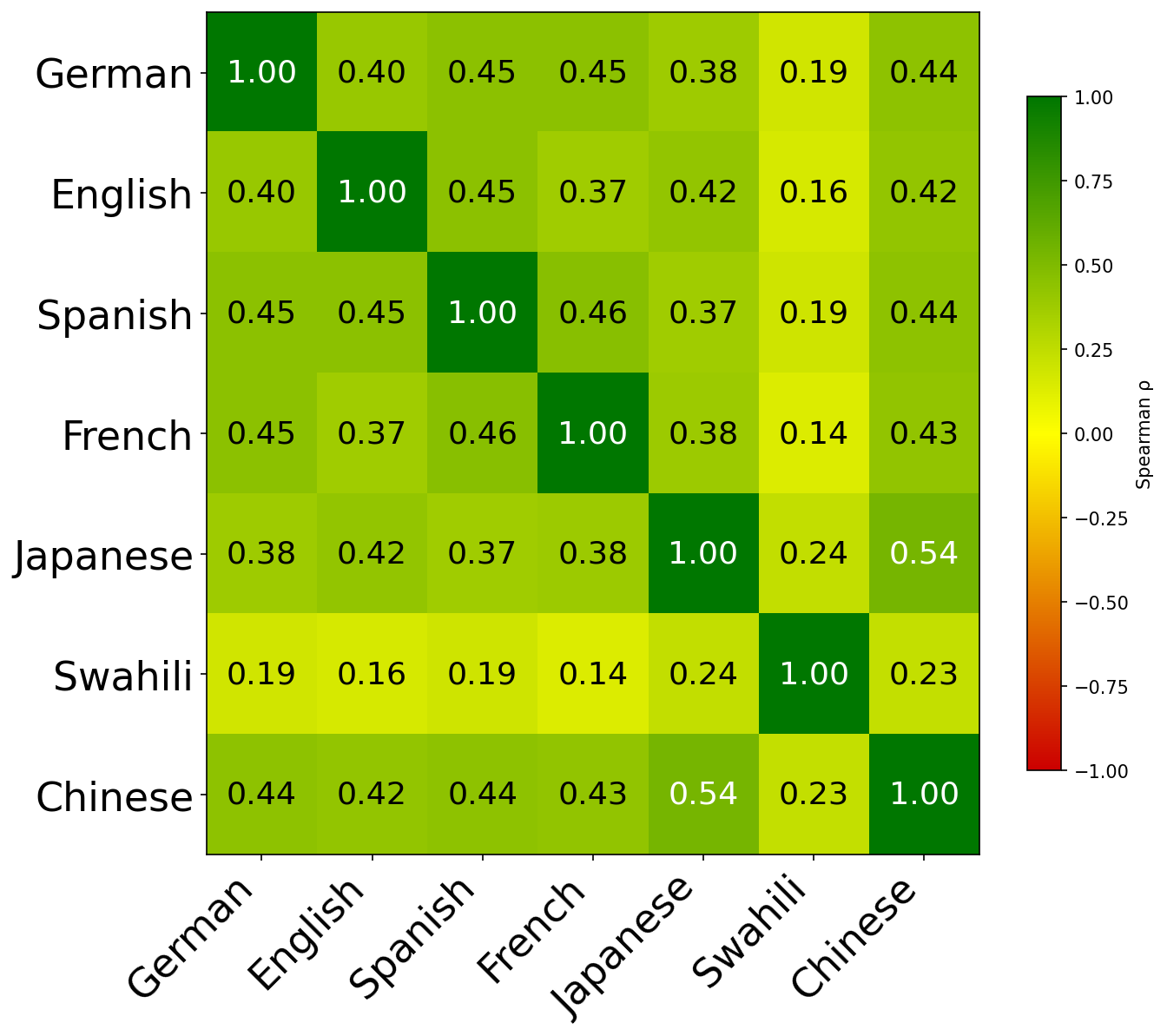}
    \caption{MMLU 3PL}\label{fig:mmlu_3pl_guess}
  \end{subfigure}
  \caption{Pairwise Spearman rank correlation of item \emph{discrimination} ($a$, columns 1--2) and \emph{guessing} ($c$, column 3) across languages for MGSM (top row) and MMLU (bottom row; element-wise average over six domains). Colour encodes $\rho \in [-1,+1]$: red $= -1$, yellow $= 0$, green $= +1$. }
  \label{fig:spearman_disc_lambda}
\end{figure}

\paragraph{Pairwise Spearman Correlations}

\autoref{fig:spearman_heatmaps} shows pairwise item \textit{difficulty} and \autoref{fig:spearman_disc_lambda} shows pairwise item \textit{discrimination} and \textit{guessing }rank correlations across languages for each IRT variant. To ensure these correlations reflect cross-lingual differences rather than within-language estimation noise, we conduct a split-half reliability check that refits 2PL/3PL via MAP on disjoint halves of the model pool and confirms strong within-language parameter agreement (\autoref{app:splithalf}). Under 1PL, all language pairs show positive monotonic agreement on both MGSM (mean $\rho = 0.498$) and MMLU (mean $\rho = 0.524$), with stronger correlations among Indo-European language pairs (e.g.\ German--Russian: $\rho = 0.718$; English--Spanish: $\rho = 0.670$) and weaker but still positive correlations for distant pairs (e.g.\ French--Telugu: $\rho = 0.190$; French--Swahili: $\rho = 0.290$). On MMLU, European languages form a tight cluster ($\rho \in [0.60, 0.67]$), while Swahili is a consistent outlier across both benchmarks. Under 2PL, difficulty correlations show a modest reduction on MGSM (mean $\rho = 0.479$) and MMLU (mean $\rho = 0.466$), with positive values across most language pairs, indicating that 2PL preserves the shared difficulty structure rather than absorbing it into language-specific variance. 
Under 3PL, difficulty correlations remain similarly positive (MGSM mean $\rho = 0.470$; MMLU mean $\rho = 0.513$). The discrimination parameter $a$ is less cross-lingually stable across both 2PL and 3PL as seen in \autoref{fig:spearman_disc_lambda} (detailed discussion in \autoref{app:heatmaps}), indicating it captures language-idiosyncratic signal. The guessing parameter $c$ under 3PL again shows stronger positive correlations across language pairs (mean $\rho = 0.349$ on MGSM; mean $\rho = 0.359$ on MMLU), with the same Swahili isolation pattern persisting, suggesting that chance-level correctness is among the most translation-stable properties of an item. The cross-domain analysis (\autoref{tab:science_vs_nonsci}) shows that scientific domains in MMLU consistently show higher correlation across languages than non-science domains.

\paragraph{Per-Item Rank Standard Deviation}

\begin{table}[t]
  \centering
  \small
  \begin{tabular}{llcc}
    \toprule
    \textbf{Model} & \textbf{Param} & \textbf{MGSM} & \textbf{MMLU} \\
                   &                & mean [min, max] & mean [min, max] \\
    \midrule
    1PL & diff & 48.5\ \ [2.8,\ \ 91.8] & 14.5\ \ [1.6,\ \ 34.3] \\
    \midrule
    2PL & diff & 49.5\ \ [4.2,\ \ 98.1] & 15.5\ \ [1.8,\ \ 40.0] \\
        & disc & 67.1\ \ [29.4,\ \ 95.6] & 19.4\ \ [2.6,\ \ 47.1] \\
    \midrule
    3PL & diff & 49.9\ \ [5.9,\ \ 99.4] & 14.7\ \ [1.4,\ \ 37.6] \\
        & disc & 67.1\ \ [19.1,\ \ 99.3] & 19.6\ \ [2.0,\ \ 52.1] \\
        & $c$  & 56.3\ \ [6.0,\ \ 85.9] & 17.0\ \ [1.5,\ \ 47.0] \\
    \bottomrule
  \end{tabular}
  \caption{Per-item rank standard deviation across languages for each IRT model variant and parameter. MMLU values are pooled across all six domains. Lower values indicate higher cross-lingual stability.}
  \label{tab:rank_std}
\end{table}

While Kendall's $W$ and Spearman $\rho$ capture aggregate ordering agreement, they mask item-level heterogeneity: per-item rank SD reveals which individual items preserve their difficulty rank across language versions of a benchmark, directly informing which items are most stable across languages in a multilingual compressed benchmark. For MGSM, stability is highly heterogeneous under 1PL, with some items near-fixed across languages and others shifting almost the full rank range (\autoref{tab:rank_std}). Under 2PL and 3PL, difficulty rank SD is essentially unchanged from 1PL (MGSM: 49.5 and 49.9 vs.\ 48.5; MMLU: 15.5 and 14.7 vs.\ 14.5), while discrimination rank SD is substantially higher ($\approx$67 on MGSM). The guessing parameter $c$ shows intermediate rank stability (MGSM: 56.3). For MMLU, the pattern mirrors MGSM: wide spread under 1PL, with difficulty stability maintained or improved under 2PL/3PL, and discrimination consistently less stable (\autoref{fig:gsm_item_rank_stability} and \autoref{fig:mmlu_item_rank_stability} in \autoref{app:rank_stability}).

\begin{figure}[t]
  \centering
  \includegraphics[width=\linewidth]{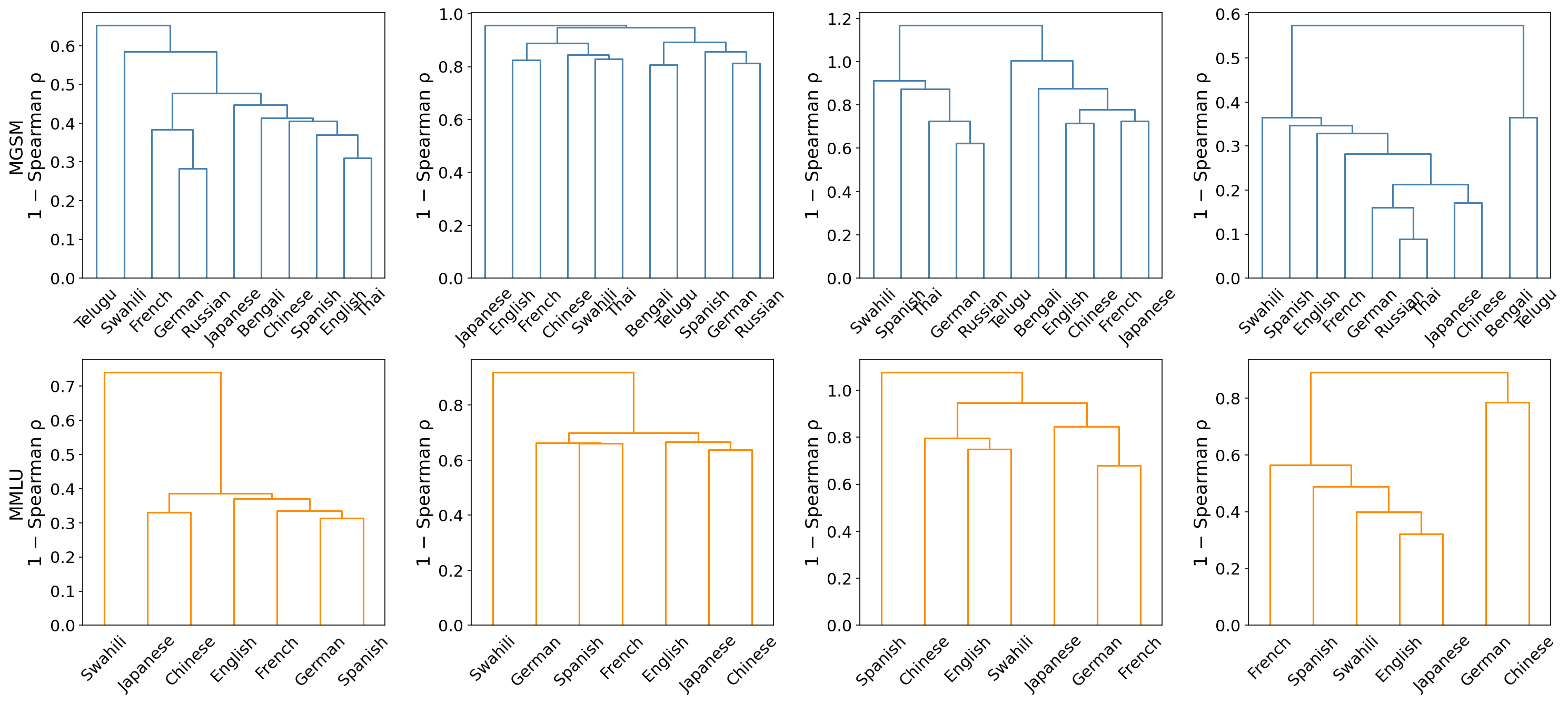}
  \caption{Language clustering by \textit{difficulty} for MGSM (top row) and MMLU (bottom row). Columns 1--3 cluster languages by pairwise Spearman correlation of item
    difficulty rank profiles under 1PL, 2PL, and 3PL respectively.
    Column 4 clusters languages by pairwise Spearman correlation of model
    accuracy vectors.}
  \label{fig:dendrograms}
\end{figure}

\begin{figure}[t]
  \centering
  \includegraphics[width=\linewidth]{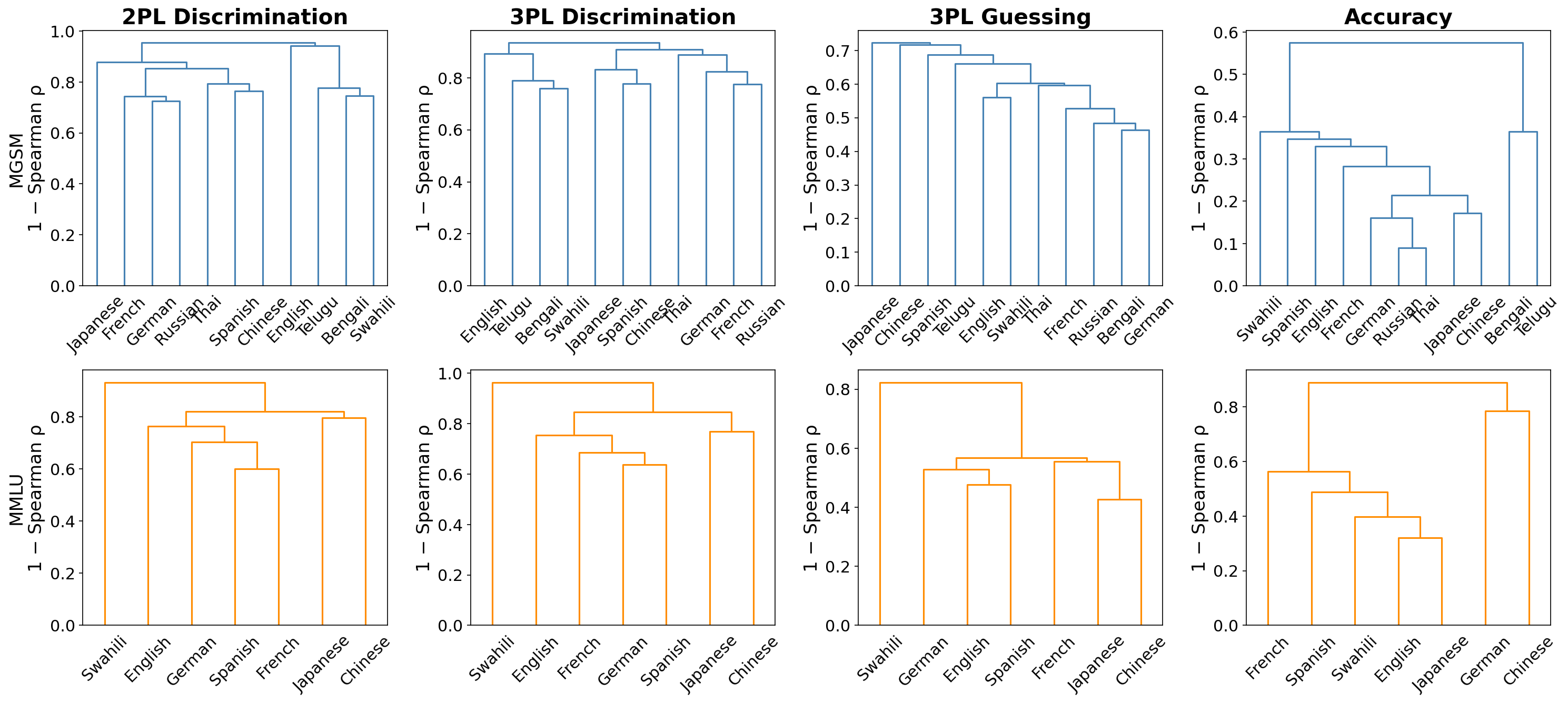}
  \caption{Language clustering by \textit{discrimination} (for 2PL and 3PL),
    and \textit{guessing }(3PL).Rows: MGSM (top), MMLU (bottom). Columns:
    2PL discrimination, 3PL discrimination, 3PL guessing, accuracy (left to right). }
  \label{fig:dendrograms_disc_guess}
\end{figure}

\paragraph{Language Clustering}
\label{sec:lang_clustering}

We cluster languages hierarchically (average linkage on Spearman distance $1-\rho$)
by each IRT parameter profile and compare to accuracy-based clusterings via Mantel
test and mean pairwise Spearman $\bar{\rho}$. For \textit{difficulty} (\autoref{fig:dendrograms}), MGSM dendrograms mirror the accuracy clustering: European languages cluster tightly while Telugu and Swahili are most isolated. A Mantel test quantifies alignment between each IRT difficulty clustering and the accuracy clustering (MGSM: 1PL $r = 0.63$, $p =0.001$; 2PL $r = 0.56$, $p = 0.002$; 3PL $r = 0.53$, $p = 0.003$; all MMLU tests non-significant: 1PL $r = -0.15$, $p = 0.751$; 2PL $r = -0.23$, $p = 0.886$; 3PL $r = -0.17$, $p = 0.786$). For MMLU, the accuracy dendrogram is structurally decoupled from IRT difficulty profiles where German and Chinese cluster as nearest neighbours despite typological distance, while Swahili clusters with European
languages in accuracy space, suggesting MMLU performance might be governed more by
alternate factors like pretraining data distribution than item difficulty structure. For \textit{discrimination} and \textit{guessing}(\autoref{fig:dendrograms_disc_guess}),
the same procedure is applied with mean pairwise $\bar{\rho}$ as an additional
cross-lingual consistency measure. Discrimination is the least cross-lingually
consistent parameter ($\bar{\rho} \approx 0.10$ on both benchmarks), producing
near-equidistant clusterings with no interpretable structure; Mantel tests are
non-significant except MGSM 2PL discrimination ($r = 0.34$, $p = 0.005$), a signal
that disappears under 3PL. Guessing shows substantially higher cross-lingual
consistency ($\bar{\rho} \approx 0.35$ on both benchmarks), visible in lower and more
differentiated merge heights, but does not align with typological or accuracy-based
groupings (all Mantel tests non-significant). 

Thus, \textit{difficulty} is the only parameter whose clustering aligns with accuracy (Mantel tests significant on MGSM, non-significant on MMLU). Neither \textit{discrimination} nor \textit{guessing} aligns with accuracy-based clusters (all Mantel tests non-significant; MGSM 3PL: disc $r = 0.134$, $p = 0.151$; guess $r = 0.074$, $p = 0.341$). However, guessing produces more differentiated dendrogram structure than discrimination: mean pairwise $\rho = 0.349$ vs.\ $0.098$ on MGSM, with lower and more varied merge heights, indicating it captures some consistent cross-lingual signal, though not one that mirrors typological or accuracy-based groupings.

\subsection{Subset Analysis }
\label{sec:iif}

\paragraph{IIF Concordance}

\begin{figure}[t]
  \centering
  \includegraphics[width=\linewidth]{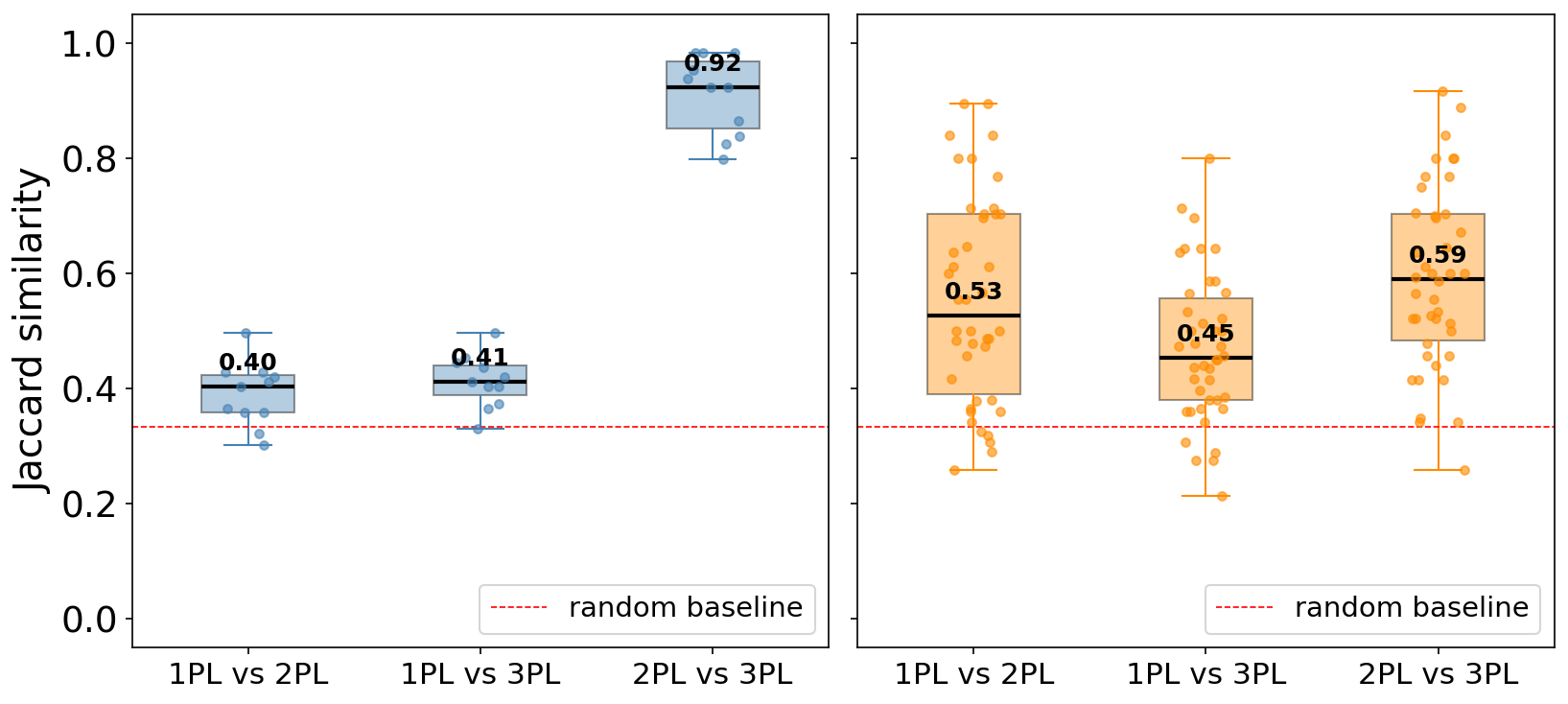}
  \caption{Jaccard similarity of the top-50\% items selected by IIF at $\theta=0$ for each model pair \textbf{(left: MGSM; right: MMLU)}, across all benchmark slices (11 MGSM languages; 42 MMLU language$\times$domain combinations). Each dot is one slice; annotated values are medians. Dashed line marks the random-selection baseline ($J \approx 0.33$).}
  \label{fig:iif_concordance}
\end{figure}

We examine whether the three IRT variants agree on which items are most informative, a choice with direct consequences for the validity of any resulting cross-lingual subset (within-language fit comparisons are in \autoref{app:withinlangfits}). Agreement is operationalised via the Item Information Function (IIF) at $\theta=0$; a sensitivity analysis at $\theta \in \{-1, 0, +1\}$ confirms that the qualitative orderings are robust to this choice (\autoref{app:theta}): 
\begin{align}
  I_j^{(1\text{PL})} &= P_j\,(1 - P_j), \label{eq:iif1pl} \\
  I_j^{(2\text{PL})} &= a_j^2\, P_j\,(1 - P_j), \label{eq:iif2pl} \\
  I_j^{(3\text{PL})} &= \frac{a_j^2\,(P_j - c_j)^2\,(1 - P_j)}{(1-c_j)^2\,P_j},
  \label{eq:iif3pl}
\end{align}
where $b_j$, $a_j$, $c_j$ are difficulty, discrimination, and guessing
parameters; $P_j^{(1\text{PL})} = \sigma(-b_j)$;
$P_j^{(2\text{PL})} = \sigma(a_j(-b_j))$;
$P_j^{(3\text{PL})} = c_j + (1-c_j)\,\sigma(a_j(-b_j))$.
Per slice, items are ranked by $I_j$ and the top 50\% retained
(avg.\ 125 items per MGSM slice; 33 per MMLU slice; sensitivity to this threshold in \autoref{app:compression}); pairwise agreement is
measured by Jaccard similarity $J(A,B) = |A \cap B|/|A \cup B|$, with an
expected baseline of $J \approx 0.33$ for two independent random draws
(\autoref{fig:iif_concordance}).Across both benchmarks, 1PL--2PL agreement (median $J = 0.40$ for MGSM, $J = 0.53$ for MMLU) and 1PL--3PL (median $J = 0.41$ for MGSM, $J = 0.45$ for MMLU) both exceed the random-selection baseline ($J \approx 0.33$). 2PL--3PL shows the strongest concordance overall (median $J = 0.92$ for MGSM, $J = 0.59$ for MMLU), confirming that 2PL and 3PL agree very strongly with each other on which items are most informative. MMLU shows lower 2PL--3PL concordance, reflecting greater variation across the heterogeneous domain slices. 

Thus, switching from 1PL to 2PL or 3PL modifies item selection but does not discard all shared signal, with 2PL and 3PL producing closely related subsets. The low 1PL--2PL/3PL concordance reflects that 1PL selects items by difficulty alone, while 2PL and 3PL IIFs are dominated by the discrimination term, prioritising different, and more language-idiosyncratic, items. The strong 2PL--3PL concordance on MGSM ($J = 0.92$) could reflect near-zero guessing rates on open-ended math items, so the guessing correction in 3PL barely shifts selection from 2PL; the lower concordance on MMLU ($J = 0.59$) confirms that guessing becomes consequential for multiple-choice items where chance-level correctness is non-negligible.

\begin{figure}[t]
  \includegraphics[width=\linewidth]{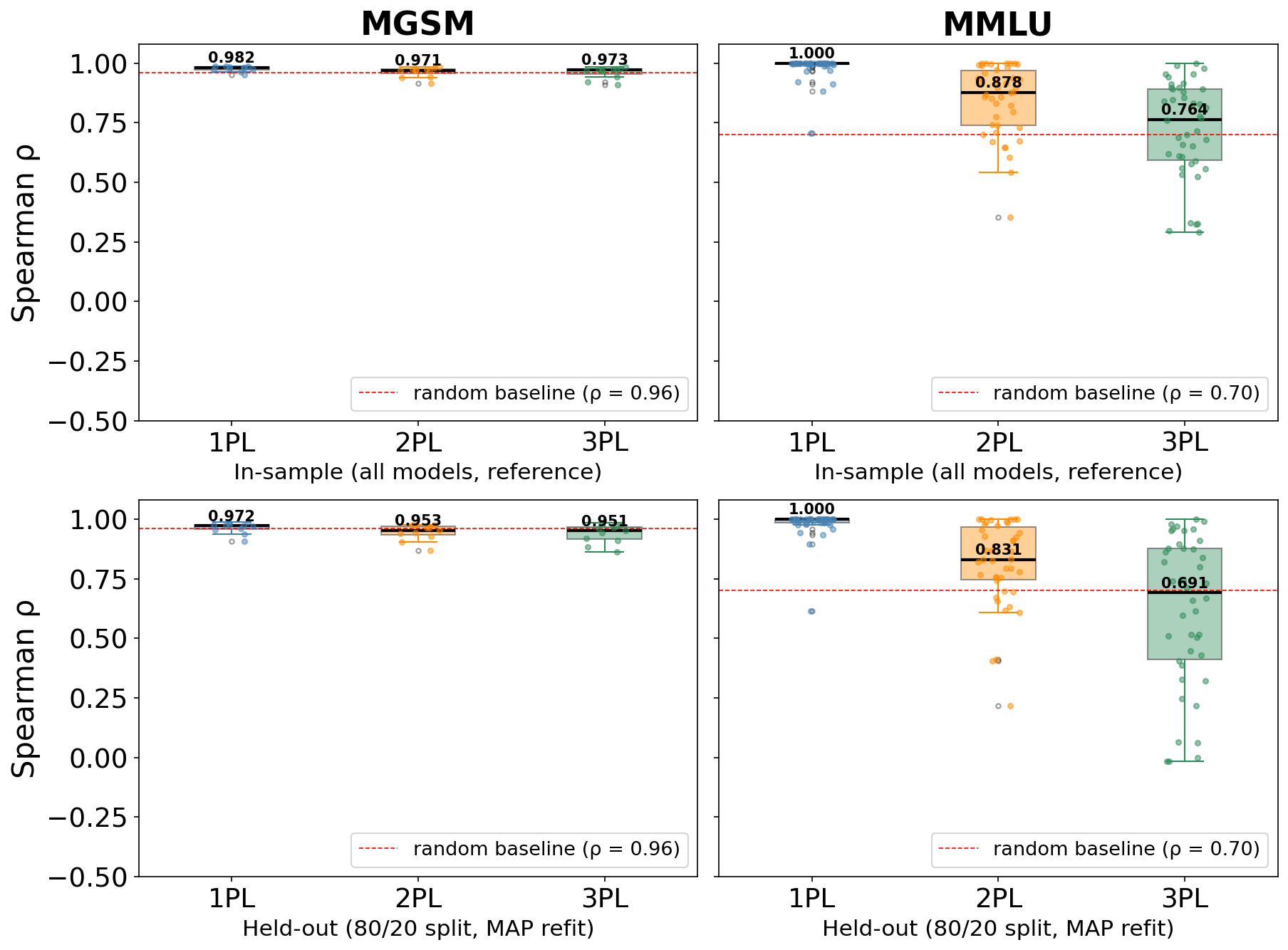}
  \caption{Ranking fidelity (Spearman $\rho$, subset vs.\ full benchmark) for MGSM
    (left) and MMLU (right). \textbf{Top row}: in-sample (all models, upper bound).
    \textbf{Bottom row}: held-out (IRT re-fitted on 80\% of models, evaluated on
    remaining 20\%). Dashed line marks the random-item baseline ($\rho \approx 0.96$ for MGSM, $\rho \approx 0.7$ for MMLU). 1PL fidelity is stable across both splits.}
  \label{fig:heldout_vs_insample}
\end{figure}

\paragraph{Ranking Fidelity}

\begin{table}[t]
  \centering
  \resizebox{\columnwidth}{!}{%
  \begin{tabular}{lcccc}
    \toprule
    & \multicolumn{2}{c}{\textbf{MGSM}} & \multicolumn{2}{c}{\textbf{MMLU}} \\
    \cmidrule(lr){2-3}\cmidrule(lr){4-5}
    \textbf{Selection strategy} & In-samp. & Held-out & In-samp. & Held-out \\
    \midrule
    1PL (difficulty)            & 0.982 & 0.972 & 1.000 & 1.000 \\
    2PL difficulty only         & 0.978 & 0.970 & 0.932 & 0.932 \\
    2PL (diff$+$disc)     & 0.971 & 0.953 & 0.878 & 0.831 \\
    3PL difficulty only         & 0.979 & 0.972 & 0.922 & 0.889 \\
    3PL (diff $+$ disc)           & 0.975 & 0.950 & 0.850 & 0.778 \\
    3PL (diff $+$ disc$+$ guess)    & 0.973 & 0.951 & 0.764 & 0.691 \\
    \bottomrule
  \end{tabular}}
  \caption{Median Spearman $\rho$ (subset vs.\ full benchmark) under progressively
    richer IIF selection strategies; the gap between rows isolates that
    parameter's marginal contribution to fidelity loss.}
  \label{tab:param_decomp}
\end{table}

Since IIF concordance shows that IRT model choice determines which items are selected,
we now ask whether those selections preserve \emph{model rankings} on the full
benchmark. We evaluate four conditions: (i)~in-sample same-language; (ii)~held-out
same-language (IRT fitted on 80\% of models, evaluated on held-out 20\%);
(iii)~in-sample cross-lingual; and (iv)~held-out cross-lingual. The primary metric is
Spearman $\rho$ between per-model subset accuracy and full-benchmark accuracy,
compared against a 200-draw random-item baseline\footnote{A mean-accuracy selection baseline achieves comparable fidelity to 1PL (cross-lingual held-out: MGSM $\rho = 0.954$ vs.\ $0.949$; MMLU $\rho = 0.879$ vs.\ $0.877$), but this does not bear on our research question, which concerns whether IRT \textit{parameters} generalise across languages, not whether IRT outperforms alternatives.} (\autoref{fig:heldout_vs_insample}). Under 1PL, same-language fidelity is high and stable across both splits: in-sample median $\rho = 0.982$ (MGSM) and $1.000$ (MMLU), dropping only marginally on held-out models ($0.972$ and $1.000$), well above the random baselines. On MGSM, 2PL and 3PL also preserve strong held-out fidelity ($\rho = 0.953$ and $0.951$), showing that with stable parameter estimation all three variants are viable. On MMLU, fidelity degrades progressively with model complexity: 2PL held-out median $\rho = 0.831$ and 3PL $\rho = 0.691$, with 3PL showing the widest spread and several low-fidelity slices (per-slice distributions and accuracy scatter plots in Appendix~\ref{app:ranking_fidelity}).

\begin{figure}[t]
  \centering
  \includegraphics[width=\columnwidth]{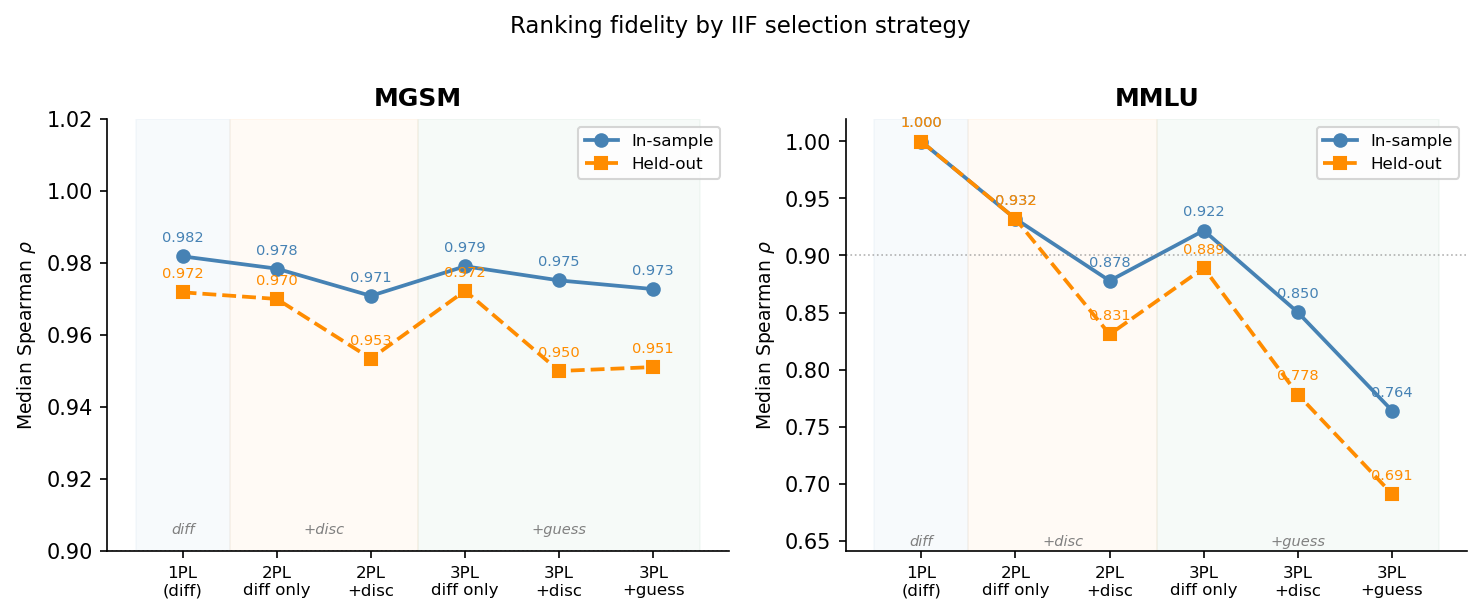}
  \caption{Median Spearman $\rho$ under progressively richer IIF selection
  strategies, isolating each parameter's marginal contribution to fidelity.}
  \label{fig:param_decomp}
\end{figure}

To isolate which parameter drives this degradation, \autoref{tab:param_decomp}
decomposes fidelity by applying progressively richer IIF formulas, also shown in \autoref{fig:param_decomp}: on MGSM all six strategies yield near-identical fidelity, confirming that difficulty alone drives the signal; on MMLU fidelity drops monotonically with each parameter added, with discrimination causing the largest single step down. Selecting items by \textit{difficulty alone} preserves high fidelity across both benchmarks and \textit{discrimination} introduces the primary degradation: $-0.017$ to $-0.022$ on MGSM and $-0.101$ to $-0.111$ on MMLU held-out, consistent with discrimination being language-idiosyncratic. Adding \textit{guessing} further reduces MMLU held-out fidelity by $0.087$ ($0.778 \to 0.691$) while leaving MGSM virtually unchanged ($0.950 \to 0.951$). Together, within-language fidelity loss is driven primarily by \textit{discrimination} rather than model complexity per se; difficulty-based or 1PL selection remains the safest strategy.

\begin{figure}[t]
  \centering
  \includegraphics[width=\columnwidth]{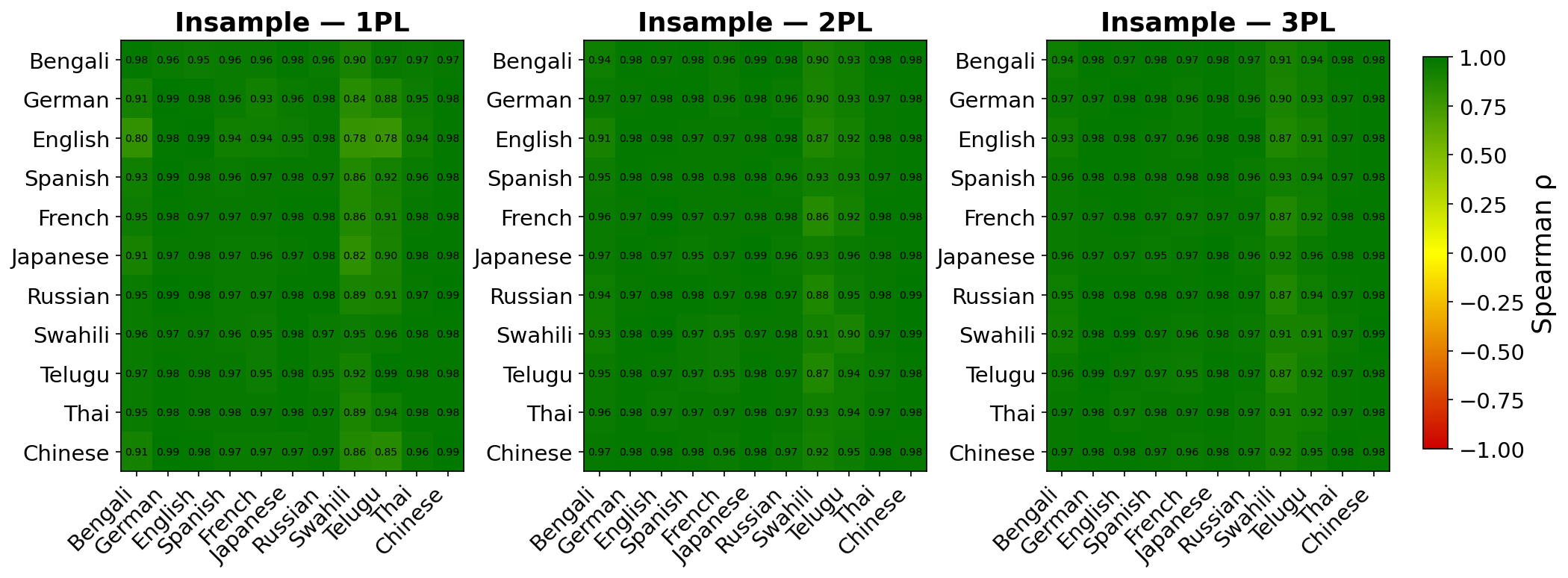}\\[2pt]
  \includegraphics[width=\columnwidth]{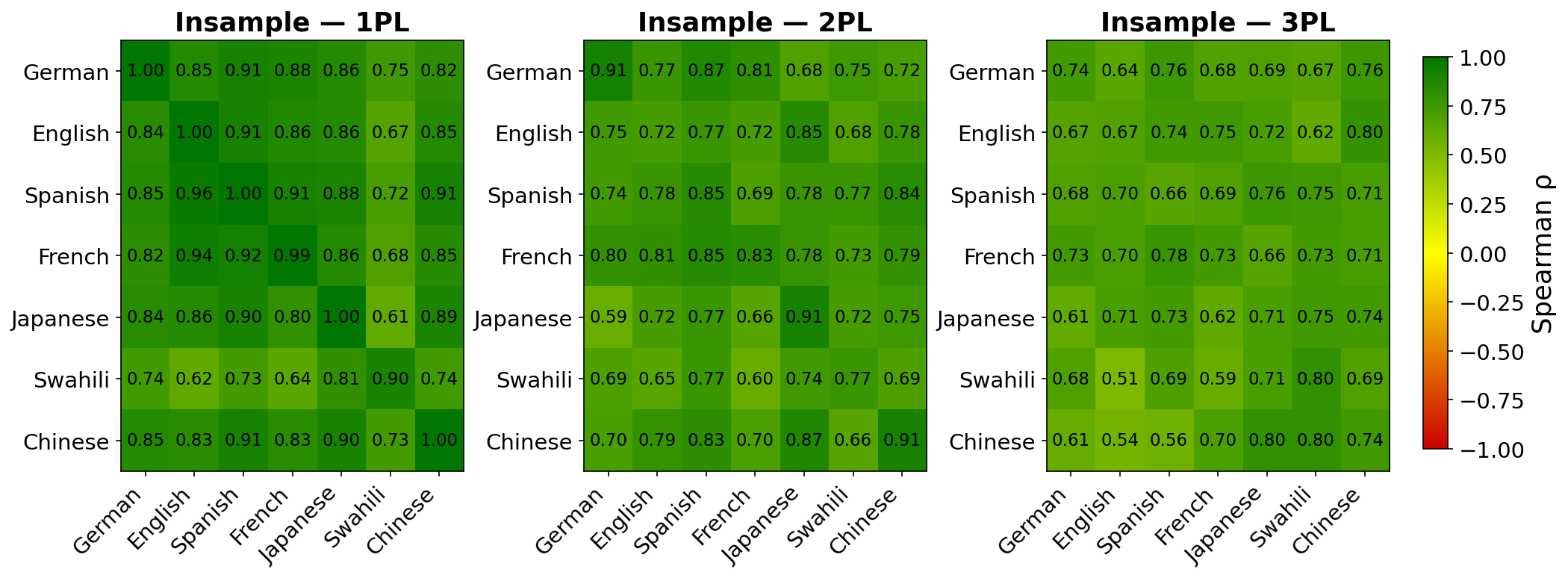}
  \caption{In-sample cross-lingual subset ranking fidelity. Cell $(i,j)$ is Spearman
    $\rho$ between subset accuracy (items from language $i$) and full-benchmark
    accuracy in language $j$. \textbf{Top:} MGSM. \textbf{Bottom:} MMLU (avg.\ over six domains). Columns = 1PL, 2PL, 3PL.}
  \label{fig:crosslang_insample}
\end{figure}

\begin{figure}[t]
  \centering
  \includegraphics[width=\columnwidth]{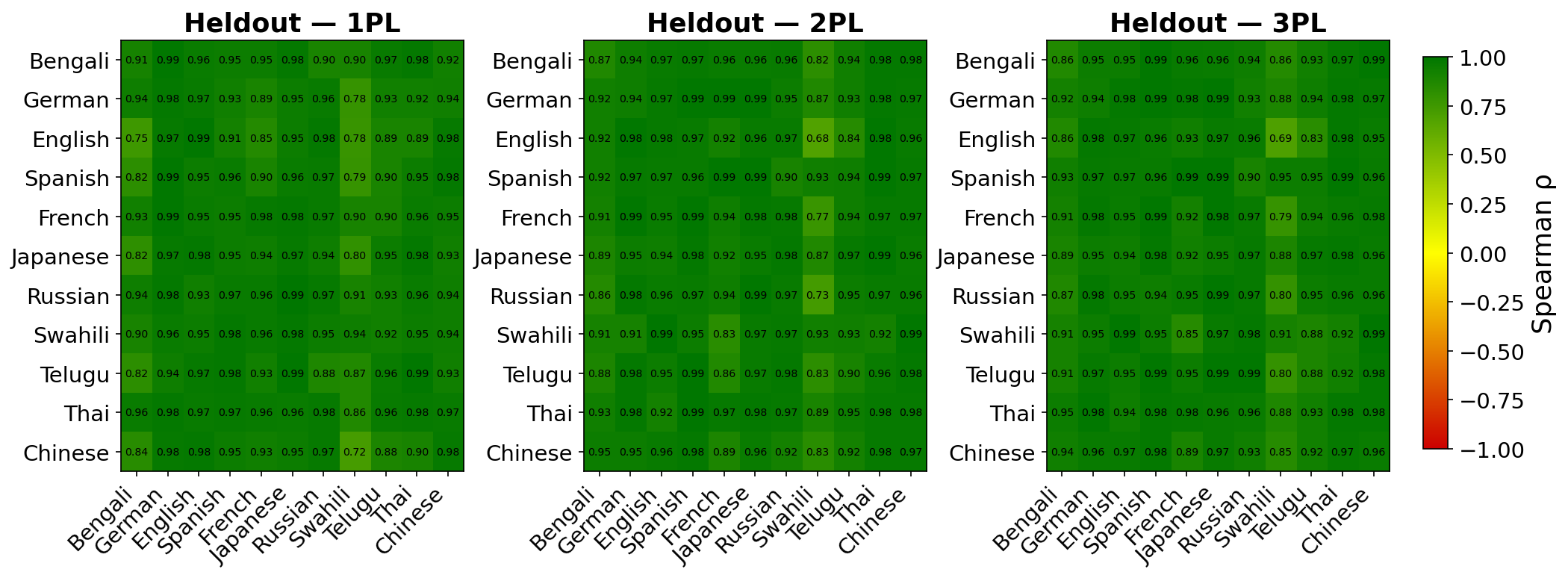}\\[2pt]
  \includegraphics[width=\columnwidth]{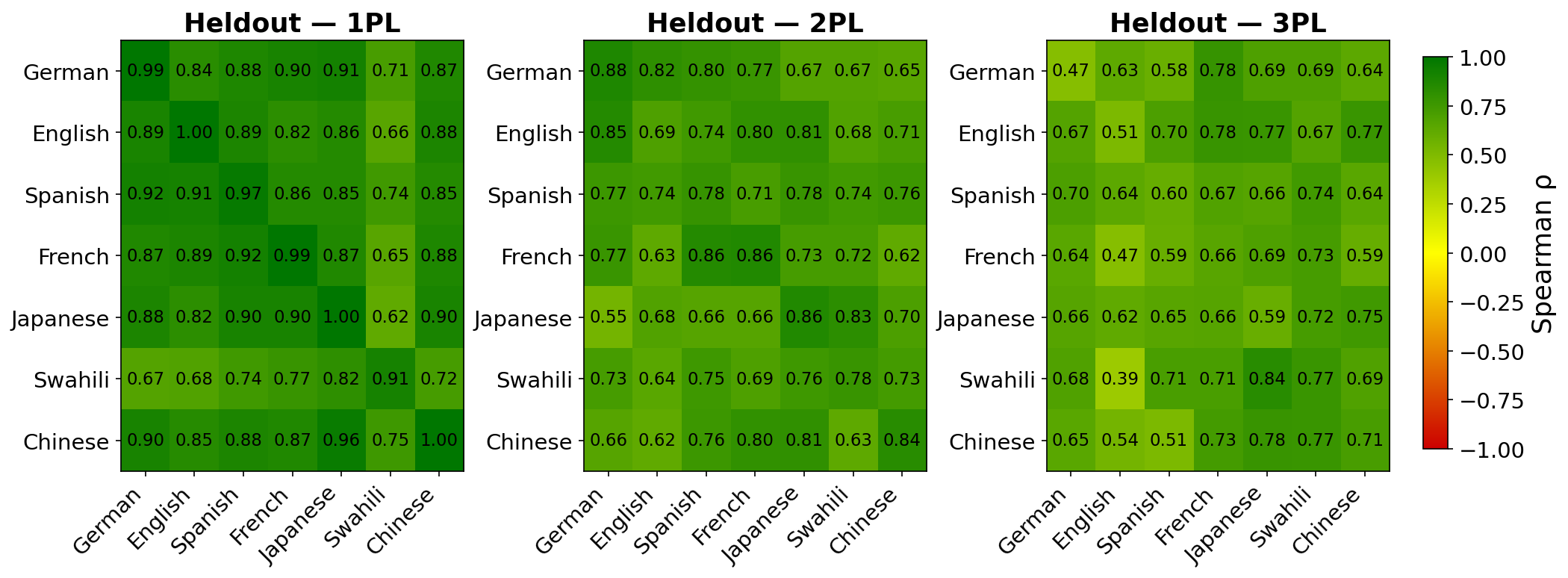}
  \caption{Held-out cross-lingual subset ranking fidelity (IRT re-fitted on 80\%
    of models via MAP, evaluated on 20\%). Same layout as
    Figure~\ref{fig:crosslang_insample}.}
  \label{fig:crosslang_heldout}
\end{figure}

\begin{table}[t]
  \centering
  \resizebox{\columnwidth}{!}{%
  \begin{tabular}{lcccc}
    \toprule
    & \multicolumn{2}{c}{\textbf{MGSM}} & \multicolumn{2}{c}{\textbf{MMLU}} \\
    \cmidrule(lr){2-3}\cmidrule(lr){4-5}
    \textbf{Selection strategy} & In-samp. & Held-out & In-samp. & Held-out \\
    \midrule
    1PL (difficulty)          & 0.968 & 0.949 & 0.854 & 0.877 \\
    2PL difficulty only       & 0.973 & 0.960 & 0.789 & 0.795 \\
    2PL (diff$+$disc)         & 0.973 & 0.962 & 0.765 & 0.777 \\
    3PL difficulty only       & 0.974 & 0.961 & 0.778 & 0.787 \\
    3PL (diff$+$disc)         & 0.972 & 0.958 & 0.761 & 0.749 \\
    3PL (diff$+$disc$+$guess) & 0.972 & 0.959 & 0.714 & 0.729 \\
    \bottomrule
  \end{tabular}}
  \caption{Median cross-lingual Spearman $\rho$ (off-diagonal pairs) under
    progressively richer IIF selection strategies. Items are selected using
    the source language IRT; fidelity is evaluated in the target language.}
  \label{tab:crosslingual_decomp}
\end{table}

Cross-lingual transfer is shown in \autoref{fig:crosslang_insample} (insample) and \autoref{fig:crosslang_heldout} (heldout), with a parameter-wise decomposition in Table~\ref{tab:crosslingual_decomp}. On MGSM, all three variants transfer well (held-out median $\rho = 0.949$--$0.962$), with 2PL and 3PL matching or slightly exceeding 1PL: arithmetic difficulty is translation-invariant and discrimination adds negligible noise. On MMLU, 1PL is the only reliable cross-lingual variant (held-out $\rho = 0.877$); fidelity degrades steadily with model complexity (2PL: $0.777$; 3PL: $0.729$). The parameter decomposition reveals that switching to 2PL \textit{difficulty alone} already accounts for most of this drop ($0.877 \to 0.795$), with discrimination contributing a further but smaller step ($0.795 \to 0.777$). Guessing contributes an additional but smaller $-0.020$ loss.  Inspecting the language-pair breakdown in the figures, this degradation is not uniform: Swahili is consistently the most isolated source and target, mirroring its outlier status in the concordance analysis (\S\ref{sec:cross_lingual}).

\begin{figure}[t]
  \centering
  \includegraphics[width=\columnwidth]{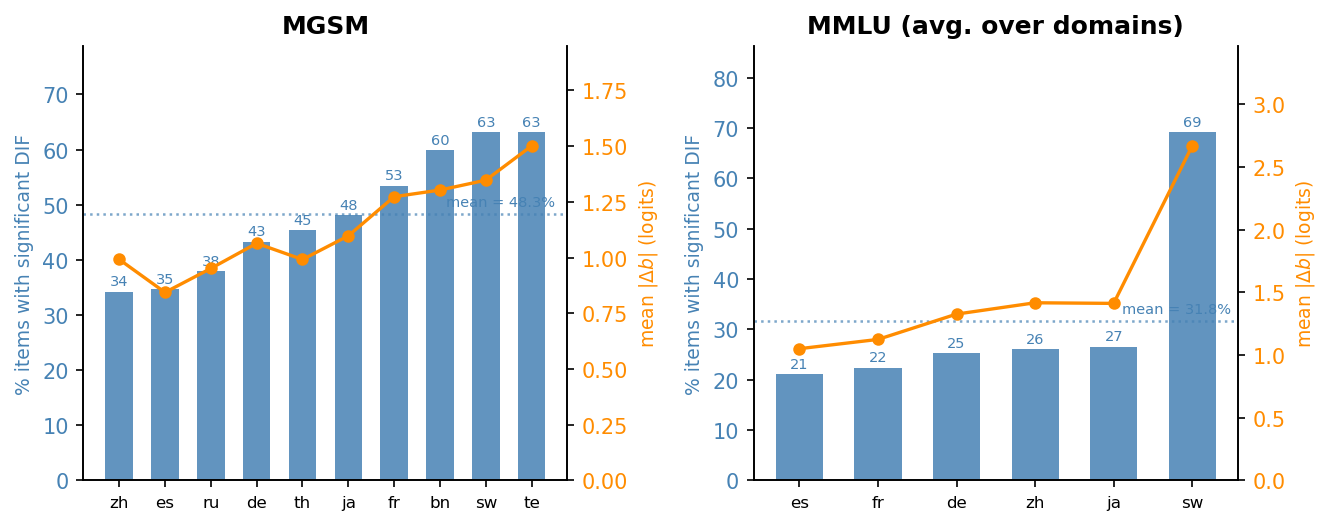}
  \caption{Percentage of items showing significant DIF relative to English (bars, left axis) and mean anchor-corrected difficulty shift $|\Delta b|$ (line, right axis), by language; MMLU averaged over its six domains. Dashed lines mark the benchmark-wide mean.}
  \label{fig:dif_summary}
\end{figure}

\subsection{DIF Analysis}
\label{sec:dif}

DIF effects are consistently strong across both benchmarks (\autoref{fig:dif_summary}). On MGSM, an average of 48.3\% of tested items (range: 34--63\% across 10 languages) show significant DIF relative to English, with a mean anchor-corrected shift of $|\Delta b| = 1.14$ logits. On MMLU, 31.8\% of items show significant DIF on average (range: 21--69\% across 6 focal languages), with a mean shift of $|\Delta b| = 1.50$ logits. Pooled across both benchmarks, this corresponds to an average shift in probability of a correct response of roughly 22 percentage points at the item's own difficulty level, rising to roughly 38 points among items flagged as DIF. Swahili shows the highest rate on MMLU (69\%) and is tied with Telugu for the highest MGSM rate (63\%), consistent with its isolation in the clustering results (\autoref{sec:lang_clustering}). Full per-language and per-domain results are in \autoref{app:dif}.

\section{Discussion}
\label{sec:analysis}

\textbf{Cross-lingual Concordance.} We find that difficulty is consistently the most stable IRT parameter across languages, and this concordance is maintained when we move to 2PL and 3PL models. Discrimination is substantially less stable across all variants and both benchmarks. Guessing sits between the two. The language clustering results on MGSM reinforce this parameter hierarchy: difficulty profiles from all three models significantly track language typology, while discrimination profiles do not. Additionally, STEM domains are consistently more concordant than humanities and social sciences across all variants, suggesting that reasoning-based items are more likely to be translation-stable than culturally situated ones.

\textbf{Subset Analysis.} Within-language, 1PL subsets preserve model rankings reliably on both benchmarks, and 2PL and 3PL are competitive on MGSM but degrade substantially on MMLU. Cross-lingually, MGSM remains robust across all three IRT variants, while on MMLU 1PL is the only reliable variant. The parameter decomposition in Table~\ref{tab:crosslingual_decomp} shows that most of the fidelity drop happens before discrimination even enters selection: switching from 1PL to 2PL difficulty alone already costs substantial fidelity on MMLU (held-out $\rho$: 0.877 $\to$ 0.795). Discrimination contributes a further but smaller drop ($0.795 \to 0.777$). Cross-lingually, therefore, 1PL's advantage lies not only in avoiding discrimination but in producing inherently more transferable difficulty estimates. Finally, the near-identical item selections of 2PL and 3PL ($J = 0.92$) on MGSM but not MMLU show that the additional complexity of 3PL does not diversify the selected subset on reasoning-heavy tasks where guessing is negligible.

\textbf{DIF Analysis.} DIF is widespread and highly item-specific: on MGSM an average of 48.3\% of items and on MMLU 31.8\% of items show significant difficulty shifts relative to English and DIF rates change (highest for Swahili and Telugu on MGSM, Swahili on MMLU), indicating that the same item-level translation pressures accumulate more heavily for languages least related to the source. We restrict this test to \emph{difficulty}: discrimination is already the least cross-lingually stable parameter in \S\ref{sec:cross_lingual}, so no discrimination-invariant anchor set can be identified to equate its scale, and guessing's lower-asymptote estimates require substantially larger per-item response counts than our model pool provides to support individual-item significance testing. Since difficulty is the most cross-lingually stable of the three parameters and still shows this much item-level heterogeneity once tested directly, it is plausible that discrimination and guessing, already less stable in aggregate, conceal comparable or greater item-level heterogeneity that a suitably powered DIF test would need to confirm. The practical implication is that this instability is item-specific rather than an inherent property of a language, and is in principle addressable through targeted re-translation or removal with item-level auditing.


More concretely, for using IRT-based data selection in benchmark filtering on multilingual translated benchmarks: (i)~we must not assume that a subset which works well within one language will generalise to others, i.e., cross-lingual validity should be verified explicitly; (ii)~we must account for domain composition, as STEM domains are reliably more cross-lingually stable than non-STEM, and a single overall fidelity check will mask important variation within mixed-domain benchmarks; (iii)~a subset can look valid by an aggregate concordance check while still containing individual items that behave inconsistently, so item-level DIF screening should complement aggregate validity checks before reusing a subset across languages. Simpler models that rely only on the most stable parameters will produce more transferable subsets.

\section{Conclusion}

We set out to test a basic but largely unexamined assumption in multilingual LLM evaluation: whether IRT, an established tool for benchmark filtering, behaves consistently when the same benchmark is translated across languages. The answer is that it does, but only for some parameters, and even for the parameter that transfers best, only for some items rather than all of them. IRT parameters form a cross-lingual stability hierarchy: difficulty transfers reliably, discrimination does not, and guessing sits in between. Even difficulty, the most stable parameter, still shows significant translation-induced shifts in roughly a third to half of items once each language's overall difficulty is accounted for, showing that cross-lingual instability is not simply a property of a language but of specific items within it. 

As translation-based multilingual benchmarks become common, the psychometric properties of translated items cannot be assumed to carry over across languages. Our findings show that the choice of IRT model, the stability of its parameters, and the presence of item-level differences determine whether a compressed benchmark remains valid across the languages it is meant to evaluate. Cross-lingual validity should be demonstrated, not assumed, and we hope this work, combining aggregate concordance checks with item-level auditing, provides a concrete empirical basis for that standard.

\section*{Limitations}
\label{sec:lim}

We discuss some limitations of our work here. \textit{First}, our model set is restricted to instruction-tuned models with fewer than 20 billion parameters, due to compute constraints. This means our IRT parameter estimates and subset fidelity results may not generalize to larger frontier models, whose ability profiles could alter item difficulty and discrimination rankings in ways not captured here.

\textit{Second}, our experiments are conducted on two benchmarks, MGSM and Global MMLU, which, while covering a broad range of languages and domains, are both derived from English source items via translation. Extending this analysis to benchmarks constructed natively in non-English languages, or to tasks beyond multiple-choice and arithmetic reasoning, remains an important direction for future work. Additionally, the Global MMLU subset used here includes four European languages (German, English, Spanish, French) out of seven, which may inflate cross-lingual concordance estimates relative to a more typologically diverse sample.

\section*{Ethical Considerations}

We do not anticipate any risks in our work. Any data used does not contain any personal details or offensive content. In this study, our use of all artifacts is consistent with the license and intended use of all datasets and models. The benchmarks used, \href{https://huggingface.co/datasets/juletxara/mgsm}{Multilingual GSM} \cite{shi2023language} and \href{https://huggingface.co/datasets/CohereLabs/Global-MMLU}{Global MMLU}\cite{singh2025global} are available on HuggingFace for free use. All large language models used are also open from HuggingFace. The \href{https://pypi.org/project/py-irt/}{py-irt library} used for 1PL model fitting is open source and freely available under MIT License. Our MAP-based implementation for 2PL and 3PL is implemented in PyTorch and will be released alongside the codebase upon publication.

\bibliography{custom}

\begin{thebibliography}{26}
\providecommand{\natexlab}[1]{#1}

\bibitem[{Ahuja et~al.(2023)Ahuja, Diddee, Hada, Ochieng, Ramesh, Jain, Nambi, Ganu, Segal, Ahmed et~al.}]{ahuja2023mega}
Kabir Ahuja, Harshita Diddee, Rishav Hada, Millicent Ochieng, Krithika Ramesh, Prachi Jain, Akshay Nambi, Tanuja Ganu, Sameer Segal, Mohamed Ahmed, and 1 others. 2023.
\newblock Mega: Multilingual evaluation of generative ai.
\newblock In \emph{Proceedings of the 2023 Conference on Empirical Methods in Natural Language Processing}, pages 4232--4267.

\bibitem[{Allalouf et~al.(1999)Allalouf, Hambleton, and Sireci}]{allalouf1999identifying}
Avi Allalouf, Ronald~K Hambleton, and Stephen~G Sireci. 1999.
\newblock Identifying the causes of dif in translated verbal items.
\newblock \emph{Journal of educational measurement}, 36(3):185--198.

\bibitem[{Bingham et~al.(2019)Bingham, Chen, Jankowiak, Obermeyer, Pradhan, Karaletsos, Singh, Szerlip, Horsfall, and Goodman}]{bingham2019pyro}
Eli Bingham, Jonathan~P. Chen, Martin Jankowiak, Fritz Obermeyer, Neeraj Pradhan, Theofanis Karaletsos, Rohit Singh, Paul~A. Szerlip, Paul Horsfall, and Noah~D. Goodman. 2019.
\newblock \href {http://jmlr.org/papers/v20/18-403.html} {Pyro: Deep universal probabilistic programming}.
\newblock \emph{J. Mach. Learn. Res.}, 20:28:1--28:6.

\bibitem[{Chollampatt et~al.(2025)Chollampatt, Pham, Indurthi, and Turchi}]{chollampatt2025cross}
Shamil Chollampatt, Minh~Quang Pham, Sathish~Reddy Indurthi, and Marco Turchi. 2025.
\newblock Cross-lingual evaluation of multilingual text generation.
\newblock In \emph{Proceedings of the 31st International Conference on Computational Linguistics}, pages 7766--7777.

\bibitem[{Doddapaneni et~al.(2025)Doddapaneni, Khan, Venkatesh, Dabre, Kunchukuttan, and Khapra}]{doddapaneni2025cross}
Sumanth Doddapaneni, Mohammed Safi Ur~Rahman Khan, Dilip Venkatesh, Raj Dabre, Anoop Kunchukuttan, and Mitesh~M Khapra. 2025.
\newblock Cross-lingual auto evaluation for assessing multilingual llms.
\newblock In \emph{Proceedings of the 63rd Annual Meeting of the Association for Computational Linguistics (Volume 1: Long Papers)}, pages 29297--29329.

\bibitem[{Gao et~al.(2024)Gao, Tow, Abbasi, Biderman, Black, DiPofi, Foster, Golding, Hsu, Le~Noac'h, Li, McDonell, Muennighoff, Ociepa, Phang, Reynolds, Schoelkopf, Skowron, Sutawika, Tang, Thite, Wang, Wang, and Zou}]{eval-harness}
Leo Gao, Jonathan Tow, Baber Abbasi, Stella Biderman, Sid Black, Anthony DiPofi, Charles Foster, Laurence Golding, Jeffrey Hsu, Alain Le~Noac'h, Haonan Li, Kyle McDonell, Niklas Muennighoff, Chris Ociepa, Jason Phang, Laria Reynolds, Hailey Schoelkopf, Aviya Skowron, Lintang Sutawika, and 5 others. 2024.
\newblock \href {https://doi.org/10.5281/zenodo.12608602} {The language model evaluation harness}.

\bibitem[{Grisay et~al.(2007)Grisay, De~Jong, Gebhardt, Berezner, and Halleux-Monseur}]{grisay2007translation}
Aletta Grisay, JHAL De~Jong, Eveline Gebhardt, Alla Berezner, and Beatrice Halleux-Monseur. 2007.
\newblock Translation equivalence across pisa countries.
\newblock \emph{Journal of Applied Measurement}, 8(3):249.

\bibitem[{Hambleton et~al.(2005)Hambleton, Merenda, Spielberger et~al.}]{hambleton2005issues}
Ronald~K Hambleton, Peter~F Merenda, Charles~D Spielberger, and 1 others. 2005.
\newblock Issues, designs, and technical guidelines for adapting tests into multiple languages and cultures.
\newblock \emph{Adapting educational and psychological tests for cross-cultural assessment}, 1:3--38.

\bibitem[{Harvey and Hammer(1999)}]{harvey1999item}
Robert~J Harvey and Allen~L Hammer. 1999.
\newblock Item response theory.
\newblock \emph{The Counseling Psychologist}, 27(3):353--383.

\bibitem[{Hofmann et~al.(2025)Hofmann, Heineman, Magnusson, Lo, Dodge, Sap, Koh, Wang, Hajishirzi, and Smith}]{hofmann2025fluid}
Valentin Hofmann, David Heineman, Ian Magnusson, Kyle Lo, Jesse Dodge, Maarten Sap, Pang~Wei Koh, Chun Wang, Hannaneh Hajishirzi, and Noah~A. Smith. 2025.
\newblock \href {https://openreview.net/forum?id=mxcCg9YRqj} {Fluid language model benchmarking}.
\newblock In \emph{Second Conference on Language Modeling}.

\bibitem[{Holland and Wainer(2012)}]{holland2012differential}
Paul~W Holland and Howard Wainer. 2012.
\newblock \emph{Differential item functioning}.
\newblock Routledge.

\bibitem[{Hu et~al.(2020)Hu, Ruder, Siddhant, Neubig, Firat, and Johnson}]{hu2020xtreme}
Junjie Hu, Sebastian Ruder, Aditya Siddhant, Graham Neubig, Orhan Firat, and Melvin Johnson. 2020.
\newblock Xtreme: A massively multilingual multi-task benchmark for evaluating cross-lingual generalisation.
\newblock In \emph{International conference on machine learning}, pages 4411--4421. PMLR.

\bibitem[{Lai et~al.(2023)Lai, Ngo, Veyseh, Man, Dernoncourt, Bui, and Nguyen}]{lai2023chatgpt}
Viet~Dac Lai, Nghia Ngo, Amir Pouran~Ben Veyseh, Hieu Man, Franck Dernoncourt, Trung Bui, and Thien~Huu Nguyen. 2023.
\newblock Chatgpt beyond english: Towards a comprehensive evaluation of large language models in multilingual learning.
\newblock In \emph{Findings of the association for computational linguistics: EMNLP 2023}, pages 13171--13189.

\bibitem[{Lalor et~al.(2016)Lalor, Wu, and Yu}]{lalor2016building}
John~P Lalor, Hao Wu, and Hong Yu. 2016.
\newblock Building an evaluation scale using item response theory.
\newblock In \emph{Proceedings of the 2016 Conference on Empirical Methods in Natural Language Processing}, pages 648--657.

\bibitem[{Lalor et~al.(2019)Lalor, Wu, and Yu}]{lalor2019learning}
John~P Lalor, Hao Wu, and Hong Yu. 2019.
\newblock Learning latent parameters without human response patterns: Item response theory with artificial crowds.
\newblock In \emph{Proceedings of the 2019 Conference on Empirical Methods in Natural Language Processing and the 9th International Joint Conference on Natural Language Processing (EMNLP-IJCNLP)}, pages 4249--4259.

\bibitem[{Lord(2012)}]{lord2012applications}
Frederic~M Lord. 2012.
\newblock \emph{Applications of item response theory to practical testing problems}.
\newblock Routledge.

\bibitem[{Milfont and Fischer(2010)}]{milfont2010testing}
Taciano~L Milfont and Ronald Fischer. 2010.
\newblock Testing measurement invariance across groups: Applications in cross-cultural research.
\newblock \emph{International Journal of psychological research}, 3(1):111--130.

\bibitem[{Natesan et~al.(2016)Natesan, Nandakumar, Minka, and Rubright}]{natesan2016bayesian}
Prathiba Natesan, Ratna Nandakumar, Tom Minka, and Jonathan~D Rubright. 2016.
\newblock Bayesian prior choice in irt estimation using mcmc and variational bayes.
\newblock \emph{Frontiers in psychology}, 7:1422.

\bibitem[{Rodriguez et~al.(2021)Rodriguez, Barrow, Hoyle, Lalor, Jia, and Boyd-Graber}]{rodriguez2021evaluation}
Pedro Rodriguez, Joe Barrow, Alexander~Miserlis Hoyle, John~P Lalor, Robin Jia, and Jordan~Lee Boyd-Graber. 2021.
\newblock Evaluation examples are not equally informative: How should that change nlp leaderboards?
\newblock In \emph{Proceedings of the 59th Annual Meeting of the Association for Computational Linguistics and the 11th International Joint Conference on Natural Language Processing (Volume 1: Long Papers)}, pages 4486--4503.

\bibitem[{Rodriguez et~al.(2022)Rodriguez, Htut, Lalor, and Sedoc}]{rodriguez2022clustering}
Pedro Rodriguez, Phu~Mon Htut, John~P Lalor, and Jo{\~a}o Sedoc. 2022.
\newblock Clustering examples in multi-dataset benchmarks with item response theory.
\newblock In \emph{Proceedings of the Third Workshop on Insights from Negative Results in NLP}, pages 100--112.

\bibitem[{Ruder et~al.(2021)Ruder, Constant, Botha, Siddhant, Firat, Fu, Liu, Hu, Garrette, Neubig et~al.}]{ruder2021xtreme}
Sebastian Ruder, Noah Constant, Jan Botha, Aditya Siddhant, Orhan Firat, Jinlan Fu, Pengfei Liu, Junjie Hu, Dan Garrette, Graham Neubig, and 1 others. 2021.
\newblock Xtreme-r: Towards more challenging and nuanced multilingual evaluation.
\newblock In \emph{Proceedings of the 2021 Conference on Empirical Methods in Natural Language Processing}, pages 10215--10245.

\bibitem[{Shi et~al.(2023)Shi, Suzgun, Freitag, Wang, Srivats, Vosoughi, Chung, Tay, Ruder, Zhou, Das, and Wei}]{shi2023language}
Freda Shi, Mirac Suzgun, Markus Freitag, Xuezhi Wang, Suraj Srivats, Soroush Vosoughi, Hyung~Won Chung, Yi~Tay, Sebastian Ruder, Denny Zhou, Dipanjan Das, and Jason Wei. 2023.
\newblock \href {https://openreview.net/forum?id=fR3wGCk-IXp} {Language models are multilingual chain-of-thought reasoners}.
\newblock In \emph{The Eleventh International Conference on Learning Representations}.

\bibitem[{Singh et~al.(2025)Singh, Romanou, Fourrier, Adelani, Ngui, Vila-Suero, Limkonchotiwat, Marchisio, Leong, Susanto et~al.}]{singh2025global}
Shivalika Singh, Angelika Romanou, Cl{\'e}mentine Fourrier, David~Ifeoluwa Adelani, Jian~Gang Ngui, Daniel Vila-Suero, Peerat Limkonchotiwat, Kelly Marchisio, Wei~Qi Leong, Yosephine Susanto, and 1 others. 2025.
\newblock Global mmlu: Understanding and addressing cultural and linguistic biases in multilingual evaluation.
\newblock In \emph{Proceedings of the 63rd Annual Meeting of the Association for Computational Linguistics (Volume 1: Long Papers)}, pages 18761--18799.

\bibitem[{Yen and Fitzpatrick(2006)}]{yen2006item}
Wendy~M Yen and Anne~R Fitzpatrick. 2006.
\newblock Item response theory.
\newblock \emph{Educational measurement}, 4:111--153.

\bibitem[{Zhang et~al.(2024)Zhang, Liu, Huang, Mao, Wang, and Hu}]{zhang2024mela}
Ziyin Zhang, Yikang Liu, Weifang Huang, Junyu Mao, Rui Wang, and Hai Hu. 2024.
\newblock Mela: Multilingual evaluation of linguistic acceptability.
\newblock In \emph{Proceedings of the 62nd Annual Meeting of the Association for Computational Linguistics (Volume 1: Long Papers)}, pages 2658--2674.

\bibitem[{Zhou et~al.(2026)Zhou, Huang, Zhao, Han, Wang, Chen, Yang, Bao, Dong, Xu et~al.}]{zhou2026lost}
Hongli Zhou, Hui Huang, Ziqing Zhao, Lvyuan Han, Huicheng Wang, Kehai Chen, Muyun Yang, Wei Bao, Jian Dong, Bing Xu, and 1 others. 2026.
\newblock Lost in benchmarks? rethinking large language model benchmarking with item response theory.
\newblock In \emph{Proceedings of the AAAI Conference on Artificial Intelligence}, volume~40, pages 35085--35093.

\end{thebibliography}

\appendix

\section{Related Work}
\label{app:related}

We discuss some relevant work here that grounds and motivates our research.

\paragraph{IRT for NLP Benchmark Filtering.} One of the earliest works in Item Response Theory for NLP by \citet{lalor2016building} applied it to annotation quality estimation to identify uninformative items in crowdsourced datasets. Subsequent work used IRT to characterise the difficulty and discriminating power of NLP test sets \cite{lalor2019learning, rodriguez2021evaluation} and to cluster benchmark items by their psychometric profiles \cite{rodriguez2022clustering}. The \texttt{py-irt} library \cite{natesan2016bayesian} made SVI-based IRT estimation tractable for NLP-scale response matrices. Most recently, IRT has been applied directly to LLM benchmarks: \citet{hofmann2025fluid} showed that IRT-selected subsets substantially reduce evaluation cost while preserving validity, and \citet{zhou2026lost} examined ranking fidelity loss under aggressive compression. Our work extends this line to multilingual settings, asking whether item parameters estimated in one language transfer to others, which is a question none of the prior NLP work addresses.

\paragraph{Multilingual LLM Evaluation.}
Large-scale multilingual evaluation suites have proliferated in recent years, including XTREME \cite{hu2020xtreme}, XTREME-R \cite{ruder2021xtreme}, and MEGA \cite{ahuja2023mega}. Translation-based benchmark construction, exemplified by MGSM \cite{shi2023language} and Global MMLU \cite{singh2025global}, is the dominant strategy for cross-lingual coverage. Prior work has documented performance gaps across languages and attributed them to pretraining data distributions \cite{lai2023chatgpt, zhang2024mela}, but the question of whether the \emph{measurement instrument itself} behaves consistently across languages has received little psychometric attention. Our work provides a principled test of this assumption at scale.

\paragraph{Differential Item Functioning.}
Differential Item Functioning (DIF) is the standard psychometric framework for testing whether an item measures the same construct at the same difficulty level across different groups of equal ability \cite{holland2012differential}. An item exhibiting DIF functions differently across groups not because the groups differ in latent ability, but because the item carries group-specific bias or ambiguity that is independent of the underlying construct. IRT-based approaches compare ICC parameters across groups and test for statistically significant differences \cite{lord2012applications}. Our work relates to but differs from classical DIF analysis in several ways. Traditional DIF tests whether a \emph{single item} functions differently across two groups conditional on latent ability, using a significance test. We instead assess whether \emph{rank orderings of all items} by their estimated IRT parameters are consistent across all language versions of a benchmark simultaneously which is a global concordance question rather than an item-level significance test. Furthermore, in our setting the ``groups'' are not distinct human subpopulations but distinct language versions of the same benchmark administered to the same set of LLMs. A formal per-item DIF analysis across language groups would complement our
concordance approach; however, classical DIF requires equating groups on latent
ability, which is ill-defined here because language itself affects overall model
performance distributions. Our global concordance framework sidesteps this
confound by assessing rank-order agreement across all items simultaneously,
rather than conditioning on matched-ability subgroups.

In the context of translated assessments, DIF has been studied extensively in educational measurement. \citet{allalouf1999identifying} showed that cross-lingual DIF in translated scholastic aptitude tests arises from changes in linguistic difficulty, item format, and cultural familiarity under translation. \citet{hambleton2005issues} and \citet{grisay2007translation} extend this to large-scale international assessments such as PISA and TIMSS, providing systematic procedures for identifying cross-lingual DIF and removing items whose difficulty shifts under translation the same validity concern that motivates our cross-lingual concordance analysis. Cross-lingual measurement invariance has been applied to psychological scales and survey instruments \cite{milfont2010testing}, but to our knowledge has not been systematically applied to LLM evaluation benchmarks.

\section{Implementation Details}
\label{app:techdetails}

\paragraph{Artifacts.}
We use LLM models publicly available on HuggingFace, and two multilingual
benchmarks also sourced from HuggingFace: Multilingual GSM
(MGSM; \citealt{shi2023language}) and Global MMLU
\cite{singh2025global}. We create and release two codebases: one for evaluating LLMs on both benchmarks, and one for fitting IRT models on the resulting evaluation data and further analysis based on that data. No new models or datasets are created.

\paragraph{LLM evaluation.} Evaluations are run using \texttt{lm-evaluation-harness} v0.4.10
\cite{eval-harness} with \texttt{vllm} v0.13.0 as the inference backend.
We evaluate 102 models on MGSM and 103 models on Global MMLU, selected
from the HuggingFace Open LLM Leaderboard as the highest-ranked
instruction-tuned open-weight models with fewer than 20 billion
parameters and native compatibility with \texttt{lm-evaluation-harness}.
Each model is called in inference-only mode with a maximum context
length of 12{,}000 tokens and a batch size of 8.
Results are stored as binary correctness labels (item--response pairs),
yielding 568{,}900 evaluated pairs in total.

\paragraph{IRT training.}
One model is fitted per benchmark slice (11 MGSM language slices,
42 MMLU language$\times$domain slices), independently for each of the
three IRT variants (1PL, 2PL, 3PL), giving 159 fitted models in total.
\textbf{1PL} is fitted using \texttt{py-irt} v0.6.6 \cite{natesan2016bayesian}
on top of \texttt{Pyro} v1.9.1 \cite{bingham2019pyro} with \texttt{PyTorch} v2.9.1
as the backend: SVI with Adam optimiser, Trace-ELBO loss, exponential learning rate
schedule ($\eta = 0.1$, $\gamma = 0.9999$/epoch), dropout $p = 0.5$, vague priors,
2{,}000 epochs; parameters at minimum-ELBO epoch retained.
\textbf{2PL and 3PL} are fitted via MAP using our own \texttt{PyTorch} implementation:
Adam optimiser with cosine annealing; 4{,}000 epochs for 2PL and 6{,}000 for 3PL; with best MAP objective retained. Held-out fidelity evaluations use an 80/20 model train/test split, with IRT refitted on the training split only; random-item baselines are averaged over 200 independent draws per slice. Statistical analyses use \texttt{scipy} v1.15.3, \texttt{numpy} v2.2.6, \texttt{pandas} v2.3.3, and \texttt{matplotlib} v3.10.3.

\paragraph{Compute.}
All LLM model evaluations are run on a high performance computing cluster using one NVIDIA A100 GPU per job (120\,GB RAM, CUDA 11.8). IRT fitting and all statistical analyses run on CPU.  Total GPU compute was approximately \textbf{280 A100-GPU-hours} for LLM inference.IRT fitting across all 159 models (53 slices $\times$ 3 variants) required under one CPU-hour in total; statistical analyses are similarly negligible.

\section{SVI vs.\ MAP Inference Comparison}
\label{app:svi_vs_map}

\begin{figure}[t]
  \centering
  \includegraphics[width=\linewidth]{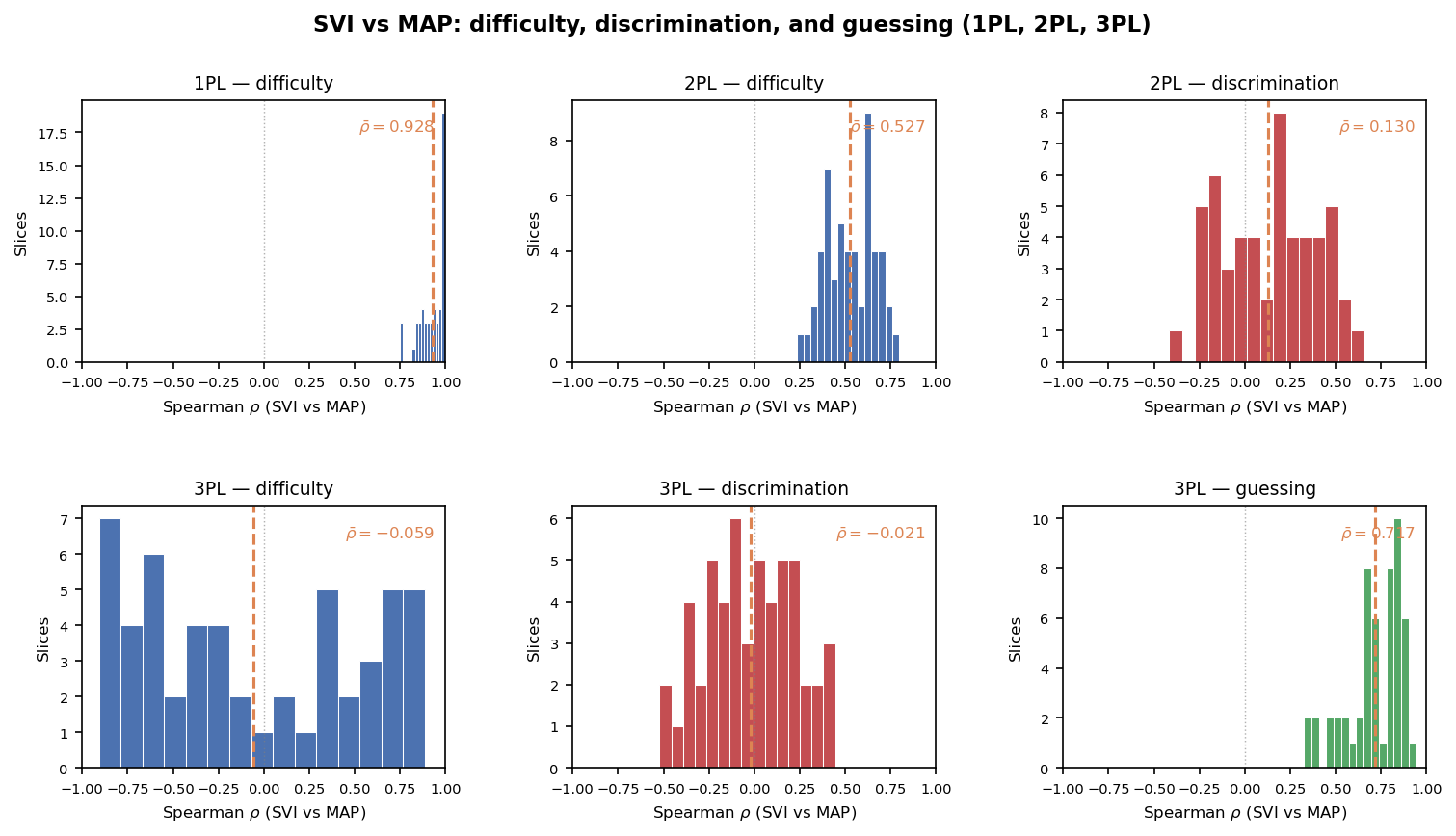}
  \caption{Per-slice Spearman $\rho$ between SVI and MAP estimates for each
    (model, parameter) combination across all 53 benchmark slices. Dashed
    orange lines mark the mean $\rho$ per panel. 1PL difficulty shows strong
    SVI--MAP agreement; discrimination collapses under SVI for 2PL and 3PL.}
  \label{fig:svi_vs_map_all}
\end{figure}

To verify that our choice of inference method does not confound cross-model
comparisons, we fit a MAP-based 1PL to all 53 benchmark slices alongside
MAP-based 2PL and 3PL, and compare difficulty, discrimination, and guessing
estimates against their SVI counterparts (\autoref{fig:svi_vs_map_all}).

For \textbf{1PL}, SVI and MAP difficulty estimates are highly concordant
(mean $\rho = 0.928$, min $= 0.749$, max $= 0.996$), confirming that using
\texttt{py-irt}'s SVI for 1PL does not affect our results.

For \textbf{2PL and 3PL}, the comparison directly illustrates why MAP is
necessary. Discrimination estimates are near-uncorrelated across the two
methods (2PL: mean $\rho = 0.130$; 3PL: mean $\rho = -0.021$): SVI drives
discrimination toward zero (2PL) or negative values (3PL), while MAP with
softplus-constrained discrimination recovers psychometrically meaningful
positive values. Under 3PL the distortion is severe enough to corrupt even
difficulty estimates (mean $\rho = -0.059$). The guessing parameter is
relatively robust across inference methods (mean $\rho = 0.717$), consistent
with its role as a lower-asymptote constraint less sensitive to discrimination
interactions.

\section{Split-Half Analysis}
\label{app:splithalf}

To verify that discrimination and guessing parameters are reliably estimated under MAP rather than under-identified given our model pool size, we conduct a within-language split-half reliability check. For each of the 11 MGSM languages, we randomly partition the full
$N{=}102$ model pool into two disjoint halves ($N{=}51$ each) and independently refit 2PL/3PL via MAP on each half, repeating this for 3 random splits per language (33 fits per model in total).

Agreement between the two halves is strong: mean Spearman $\rho = 0.614$ for 2PL discrimination ($[0.424, 0.754]$), $\rho = 0.621$ for 3PL discrimination ($[0.382, 0.762]$), and $\rho = 0.708$ for 3PL guessing ($[0.517, 0.872]$), despite each half containing roughly half as many models as the full sample used in the paper. Each half also correlates strongly with the full-sample ($N{=}102$) estimates reported in the main text: $\rho \in [0.776, 0.931]$ (mean $0.866$) for 2PL discrimination, $\rho \in [0.748, 0.917]$ (mean $0.860$) for 3PL discrimination, and $\rho \in [0.708, 0.973]$ (mean $0.874$) for 3PL guessing.

Guessing spread at $N{=}51$ is nearly identical to the full sample (mean $\sigma = 0.049$ vs.\ $0.051$); discrimination spread is somewhat reduced at $N{=}51$ (mean $\sigma = 0.445$ vs.\ $0.578$ for 2PL, $0.418$ vs.\ $0.549$ for 3PL), consistent with expected shrinkage
toward the MAP prior under a smaller model pool, though item rankings
by discrimination are preserved, as shown by the correlations above. Together, these results indicate that discrimination and guessing are reliably recoverable within a language at half the model pool size, and that their low cross-lingual concordance (\autoref{app:heatmaps}) reflects cross-lingual instability rather than under-identified noise.

\section{Correlated Raters Analysis}
\label{app:correlatedraters}

Kendall's $W$ aggregates agreement across raters (here, languages); a tight cluster of correlated raters could inflate the statistic even if the remaining raters disagree substantially, analogous to non-independent raters inflating inter-rater reliability.
\autoref{sec:lang_clustering} shows that European languages form exactly such a tight cluster under item difficulty. To test whether this cluster is responsible for the concordance values reported in \S\ref{sec:cross_lingual}, we recompute $W$ after removing all
Indo-European languages: German, English, Spanish, French, Bengali, and Russian for MGSM (5 languages remain: Japanese, Swahili, Telugu, Thai, Chinese), and German, English, Spanish, and French for MMLU (3 languages remain: Japanese, Swahili, Chinese), the latter averaged
over its six domains (results in \autoref{tab:leavefamilyout}).

\begin{table}[t]
  \centering
  \small
  \begin{tabular}{lccc}
    \toprule
    & $W_\text{full}$ & $W_\text{restr.}$ & $\Delta$ \\
    \midrule
    MGSM 1PL & 0.544 & 0.560 & $+0.016$ \\
    MGSM 2PL & 0.526 & 0.553 & $+0.027$ \\
    MGSM 3PL & 0.519 & 0.547 & $+0.028$ \\
    \midrule
    MMLU 1PL & 0.592 & 0.606 & $+0.014$ \\
    MMLU 2PL & 0.540 & 0.574 & $+0.034$ \\
    MMLU 3PL & 0.582 & 0.593 & $+0.011$ \\
    \bottomrule
  \end{tabular}
  \caption{Kendall's $W$ for item difficulty before ($W_\text{full}$)
    and after ($W_\text{restricted}$) removing Indo-European languages.
    MGSM: 11 $\to$ 5 languages. MMLU: 7 $\to$ 3 languages (mean across
    domains). Concordance increases in every case.}
  \label{tab:leavefamilyout}
\end{table}

Removing the Indo-European cluster \emph{increases} $W$ in every case: MGSM rises from $0.544 \to 0.560$ (1PL), $0.526 \to 0.553$ (2PL), and $0.519 \to 0.547$ (3PL); MMLU rises from $0.592 \to 0.606$ (1PL), $0.540 \to 0.574$ (2PL), and $0.582 \to 0.593$ (3PL). If the Indo-European cluster were propping up the reported concordance, its removal should have lowered $W$; instead, the more typologically distinct remaining languages reproduce the same signal on their own. The MMLU restricted set has only three languages and its estimates
therefore carry more sampling variance than the seven-language originals, but the same pattern holds on MGSM's larger five-language restricted set. These results indicate that the
reported concordance values are not an artifact of rater redundancy.

\section{Spearman Rank Correlation Heatmaps}
\label{app:heatmaps}

In main text, \autoref{fig:spearman_heatmaps} shows pairwise Spearman
rank correlations for item \emph{difficulty} while \autoref{fig:spearman_disc_lambda}
provides the corresponding heatmaps for \emph{discrimination} ($a$,
available from 2PL onward) and \emph{guessing} ($c$, 3PL only). Here, we provide a more detailed discussion of findings:

\paragraph{Discrimination ($a$).}Discrimination parameters show low cross-lingual agreement under both 2PL and 3PL. For MGSM 2PL, the mean pairwise $\rho = 0.103$
(range $[-0.068$ (English--Japanese) to $0.275$ (German--Russian)]):
while the estimates are positive and psychometrically meaningful within each language,
they do not agree strongly across languages, indicating that how sharply an item
discriminates between models is a language-specific property of item formulation.
Under MGSM 3PL the mean pairwise $\rho = 0.098$ (range $[-0.085,\, 0.241]$),
consistent with the additional guessing parameter absorbing some residual shared
signal and leaving discrimination with little cross-lingual structure.
MMLU 2PL and 3PL show a similarly low average discrimination agreement
($\bar{\rho} \approx 0.10$ for both), indicating that even with larger per-domain
item pools, discrimination does not generalize cross-lingually.
This provides clear evidence that discrimination captures language-idiosyncratic
surface features of item formulation rather than a shared psychometric property.

\paragraph{Guessing ($c$).}
The guessing parameter tells a markedly different story.
For MGSM 3PL, $\bar{\rho} = 0.349$ (range $[0.186,\, 0.537]$), with
Bengali--German achieving the strongest agreement ($\rho = 0.537$),
a plausible finding if both language versions impose similar chance-level
accessibility on the same underlying items.
English--Telugu is the weakest pair ($\rho = 0.186$); all pairs remain positive.
For MMLU 3PL, guessing agreement is comparably stable:
$\bar{\rho} = 0.359$ (range $[0.138,\, 0.538]$), with every language
pair yielding a positive correlation and Japanese--Chinese the most
consistent ($\rho = 0.538$).
The consistently positive guessing correlations, in contrast to near-zero
or negative discrimination correlations under the same model, suggest that
lower-bound response behavior is a genuine language-invariant property: some items simply impose a floor on the probability of a correct
response regardless of the language of presentation, while discrimination
is driven by language-specific surface features of the item formulation.

\begin{figure*}[!t]
  \centering
  \includegraphics[width=\linewidth]{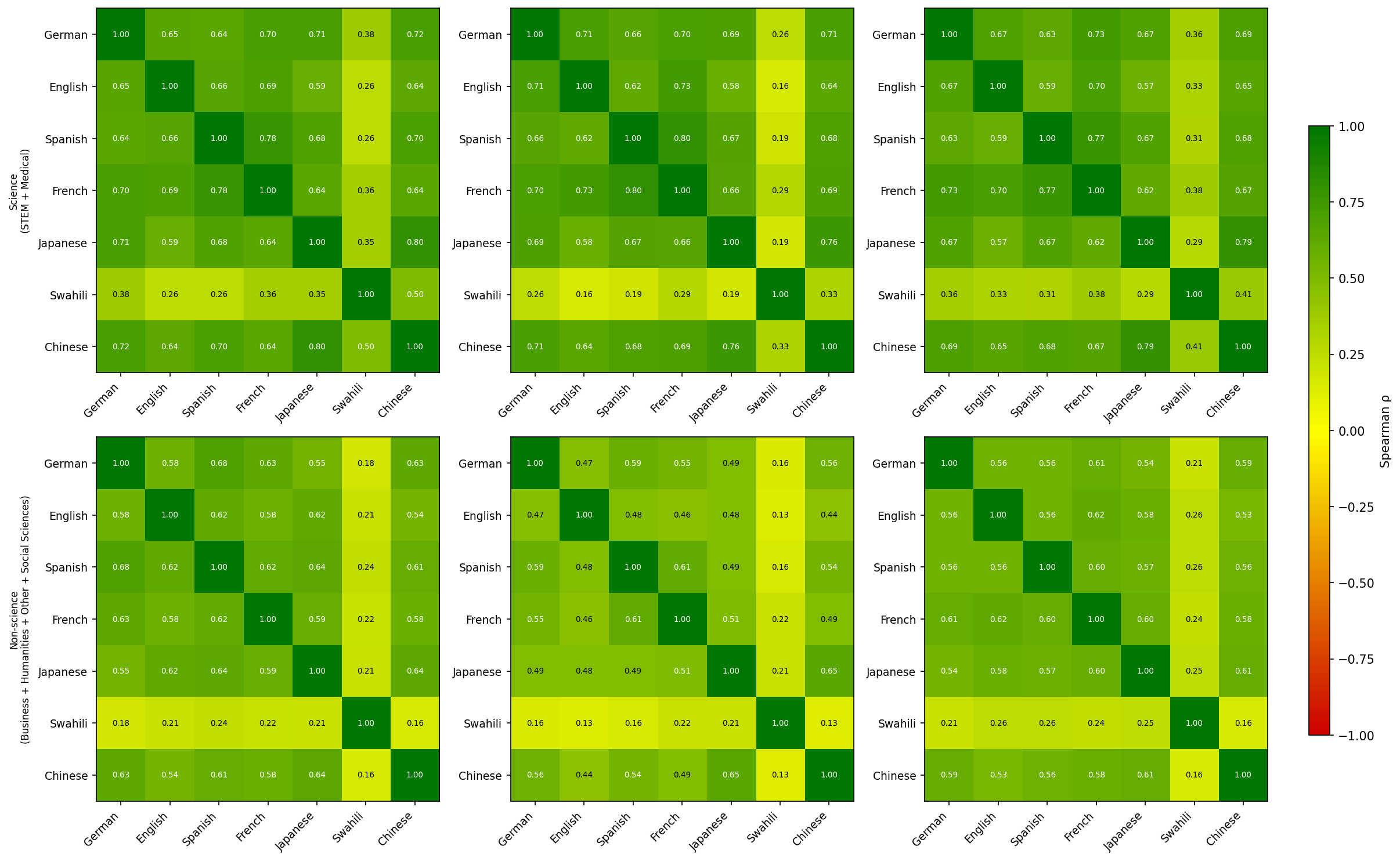}
  \caption{Pairwise Spearman rank correlation of item \emph{difficulty} across
    languages for MMLU, split by domain type (rows) and IRT model variant (columns).
    \textbf{Top row:} Science domains (STEM + Medical), averaged element-wise.
    \textbf{Bottom row:} Non-science domains (Business + Humanities + Other +
    Social Sciences), averaged element-wise.
    Colour encodes $\rho \in [-1,+1]$: red $= -1$, yellow $= 0$, green $= +1$.}
  \label{fig:science_vs_nonscience}
\end{figure*}

\paragraph{Science vs Non-Science Domain}
Figure~\ref{fig:science_vs_nonscience} shows the 2$\times$3 grid of averaged
Spearman difficulty heatmaps for science vs non-science domains across the three
IRT model variants.
Under 1PL, science heatmaps show a predominantly warm palette
(Kendall's $W = 0.647$, $\bar{\rho} = 0.588$), while non-science is already
lower ($\bar{\rho} = 0.492$, $W = 0.564$).
Under MAP-fitted 2PL and 3PL, the science advantage is maintained: science
domains achieve $\bar{\rho} = 0.559$, $W = 0.620$ (2PL) and $\bar{\rho} = 0.581$,
$W = 0.640$ (3PL), while non-science remains somewhat lower at $\bar{\rho} = 0.420$,
$W = 0.501$ (2PL) and $\bar{\rho} = 0.479$, $W = 0.553$ (3PL).
Both domain types maintain moderate cross-lingual concordance across all model
variants, with science items consistently showing stronger agreement.
This reflects that science items test well-defined reasoning steps or unambiguous
facts whose difficulty is partly preserved under translation, whereas non-science
difficulty is dominated by culturally situated knowledge and nuanced phrasing
that shifts substantially across languages.

\section{DIF Analysis}
\label{app:dif}

\autoref{tab:dif_mgsm} and \autoref{tab:dif_mmlu} report the per-language DIF rates and mean anchor-corrected difficulty shifts underlying \autoref{sec:dif}, and \autoref{tab:dif_domain} breaks the MMLU results down by domain. For MGSM, the anchor set is fixed at 63 items (bottom quartile of items by cross-lingual rank standard
deviation) and 187 items are tested per language pair. For MMLU, both the anchor and tested set sizes vary by domain, following each domain's item count.

\begin{table}[t]
  \centering
  \small
  \begin{tabular}{lcc}
    \toprule
    \textbf{Lang.} & \textbf{\% DIF} & \textbf{mean $|\Delta b|$} \\
    \midrule
    \texttt{zh} & 34.2 & 0.992 \\
    \texttt{es} & 34.8 & 0.843 \\
    \texttt{ru} & 38.0 & 0.951 \\
    \texttt{de} & 43.3 & 1.064 \\
    \texttt{th} & 45.5 & 0.991 \\
    \texttt{ja} & 48.1 & 1.097 \\
    \texttt{fr} & 53.5 & 1.273 \\
    \texttt{bn} & 59.9 & 1.303 \\
    \texttt{sw} & 63.1 & 1.347 \\
    \texttt{te} & 63.1 & 1.501 \\
    \bottomrule
  \end{tabular}
  \caption{MGSM: \% of items showing significant DIF relative to
    English (BH-FDR $q=0.05$), and mean anchor-corrected difficulty
    shift $|\Delta b|$, by language. $n_\text{anchor}=63$,
    $n_\text{tested}=187$ for every language pair.}
  \label{tab:dif_mgsm}
\end{table}

\begin{table}[t]
  \centering
  \small
  \begin{tabular}{lcc}
    \toprule
    \textbf{Lang.} & \textbf{\% DIF} & \textbf{mean $|\Delta b|$} \\
    \midrule
    \texttt{es} & 21.0 & 1.051 \\
    \texttt{fr} & 22.4 & 1.124 \\
    \texttt{de} & 25.3 & 1.326 \\
    \texttt{zh} & 26.1 & 1.417 \\
    \texttt{ja} & 26.6 & 1.412 \\
    \texttt{sw} & 69.1 & 2.668 \\
    \bottomrule
  \end{tabular}
  \caption{MMLU: \% of items showing significant DIF relative to
    English, and mean anchor-corrected difficulty shift, by language,
    each averaged across the six MMLU domains.}
  \label{tab:dif_mmlu}
\end{table}

\begin{table}[t]
  \centering
  \small
  \begin{tabular}{lcc}
    \toprule
    \textbf{Domain} & \textbf{\% DIF} & \textbf{mean $|\Delta b|$} \\
    \midrule
    Humanities      & 23.5 & 1.190 \\
    Medical         & 25.9 & 1.441 \\
    Other           & 28.2 & 1.248 \\
    Business        & 32.6 & 1.842 \\
    Social Sciences & 35.3 & 1.621 \\
    STEM            & 45.1 & 1.656 \\
    \bottomrule
  \end{tabular}
  \caption{MMLU: \% of items showing significant DIF and mean
    anchor-corrected difficulty shift, by domain, averaged across the
    six focal languages relative to English.}
  \label{tab:dif_domain}
\end{table}

\section{Per-Item Rank Stability Distributions}
\label{app:rank_stability}

\begin{figure*}[!t]
  \centering
  \begin{subfigure}[t]{0.32\linewidth}
    \includegraphics[width=\linewidth]{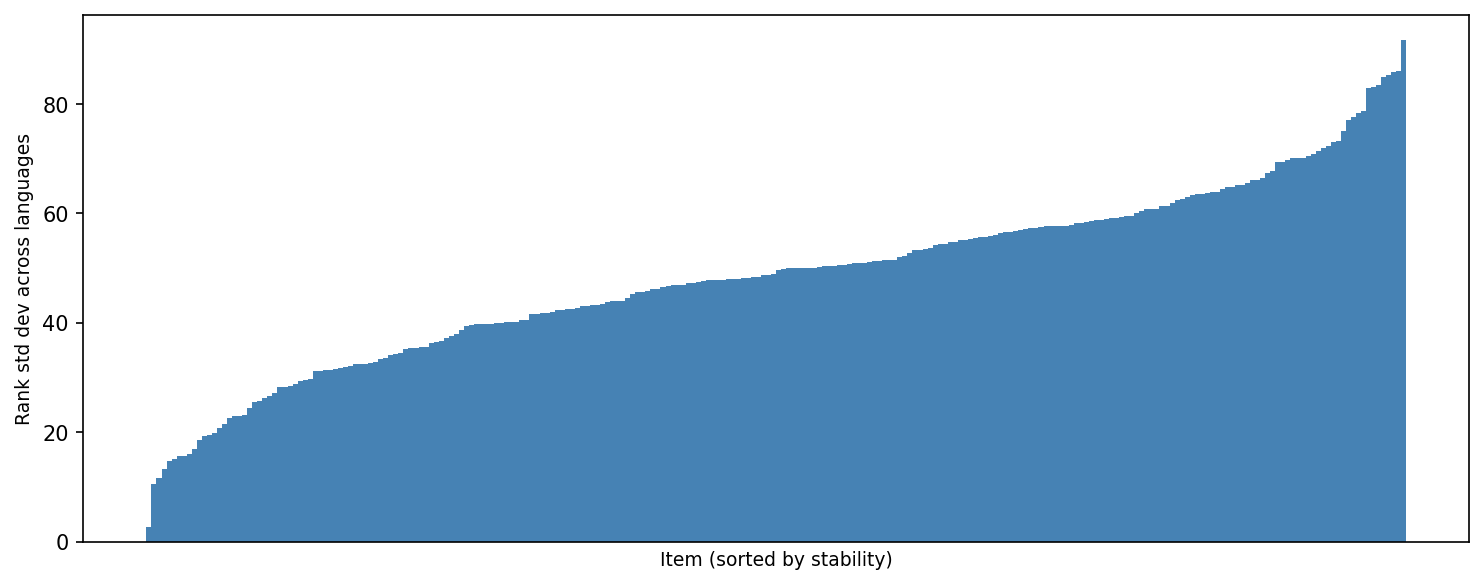}
    \caption{1PL (difficulty)}\label{fig:gsm_1pl_stab}
  \end{subfigure}\hfill
  \begin{subfigure}[t]{0.32\linewidth}
    \includegraphics[width=\linewidth]{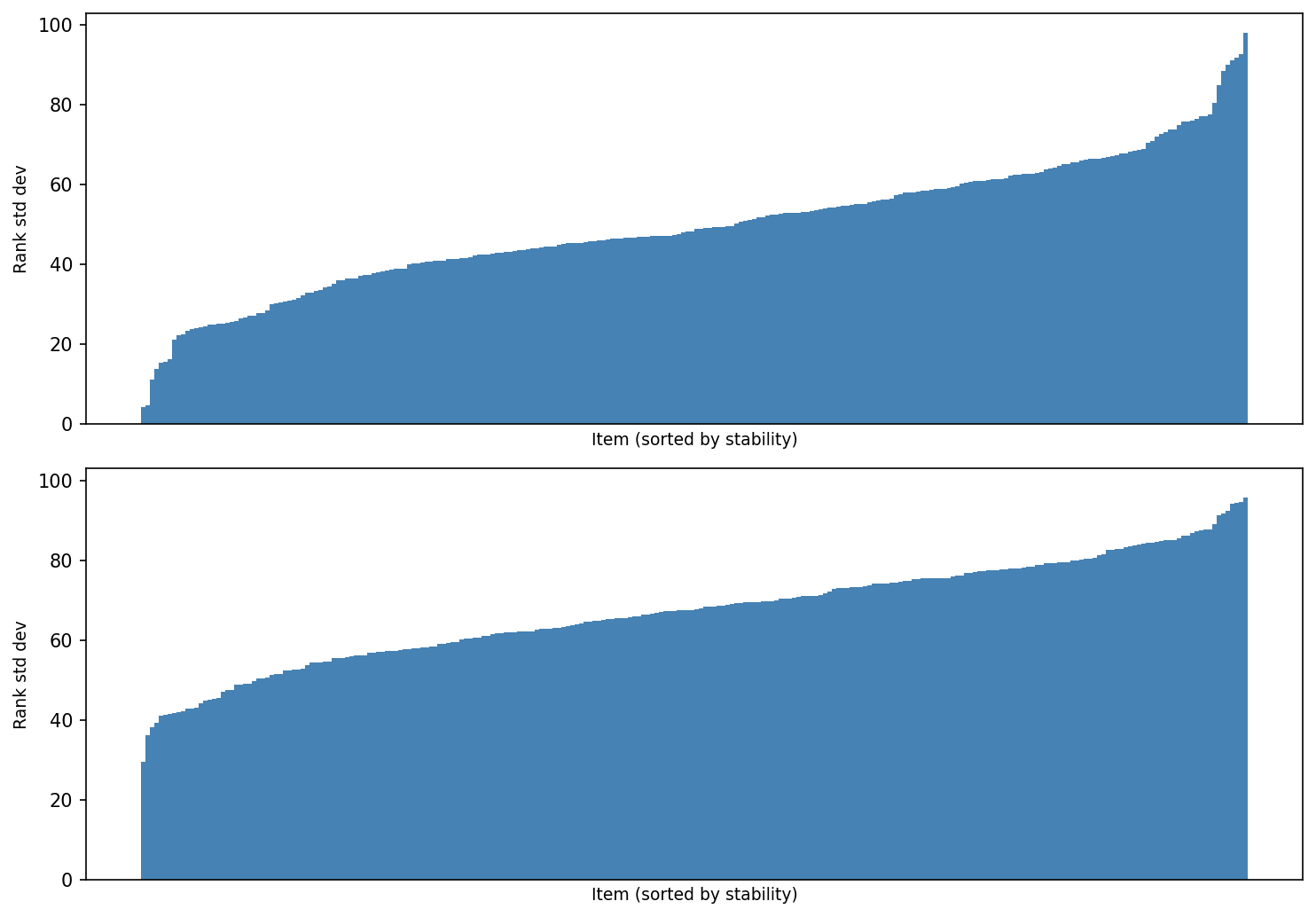}
    \caption{2PL (difficulty + discrimination)}\label{fig:gsm_2pl_stab}
  \end{subfigure}\hfill
  \begin{subfigure}[t]{0.32\linewidth}
    \includegraphics[width=\linewidth]{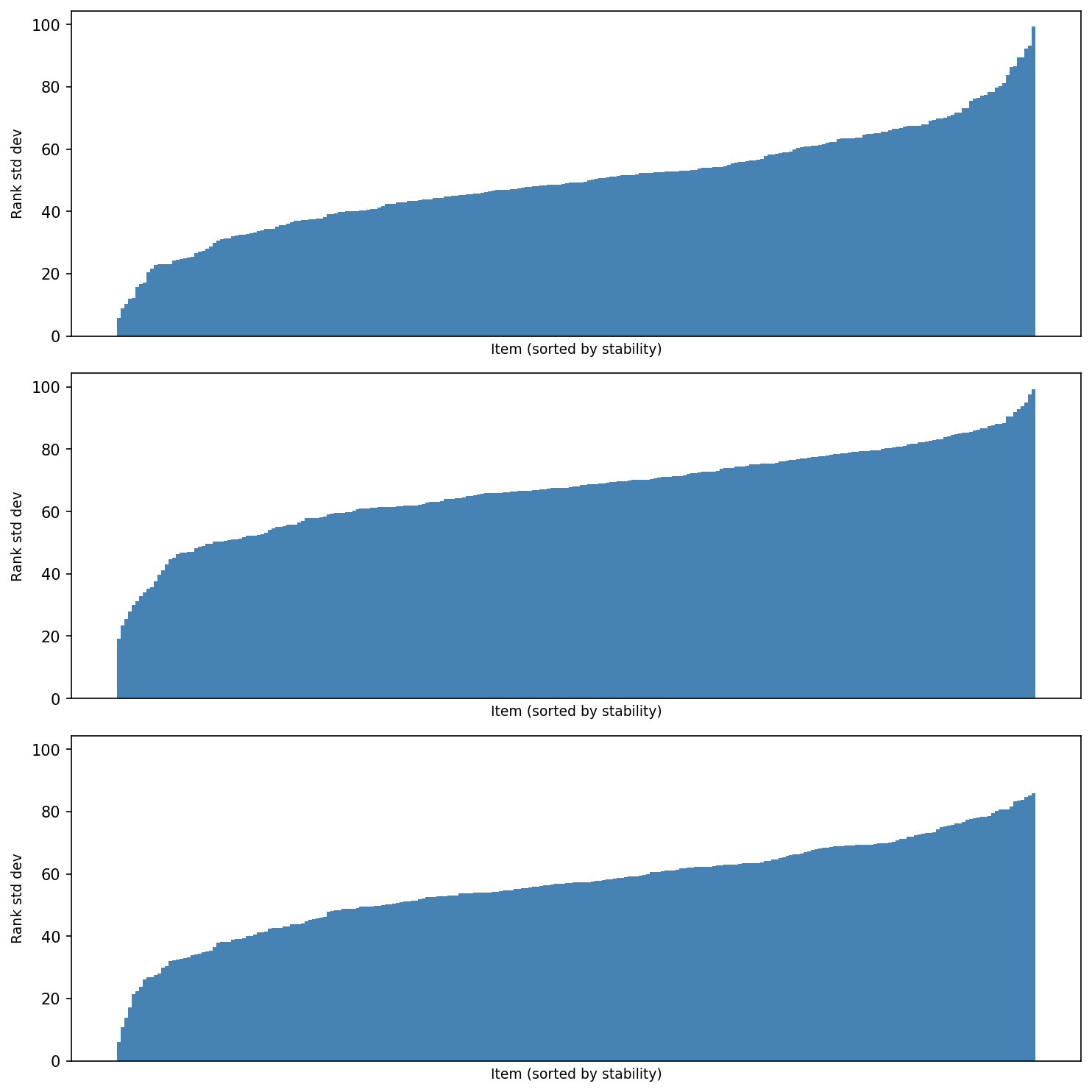}
    \caption{3PL (difficulty + discrimination + guessing)}\label{fig:gsm_3pl_stab}
  \end{subfigure}
  \caption{MGSM per-item rank standard deviation across 11 languages (250 items). Each subfigure shows all parameters for that model variant stacked vertically.}
  \label{fig:gsm_item_rank_stability}
\end{figure*}

\begin{figure*}[!t]
  \centering
  \begin{subfigure}[t]{0.32\linewidth}
    \includegraphics[width=\linewidth]{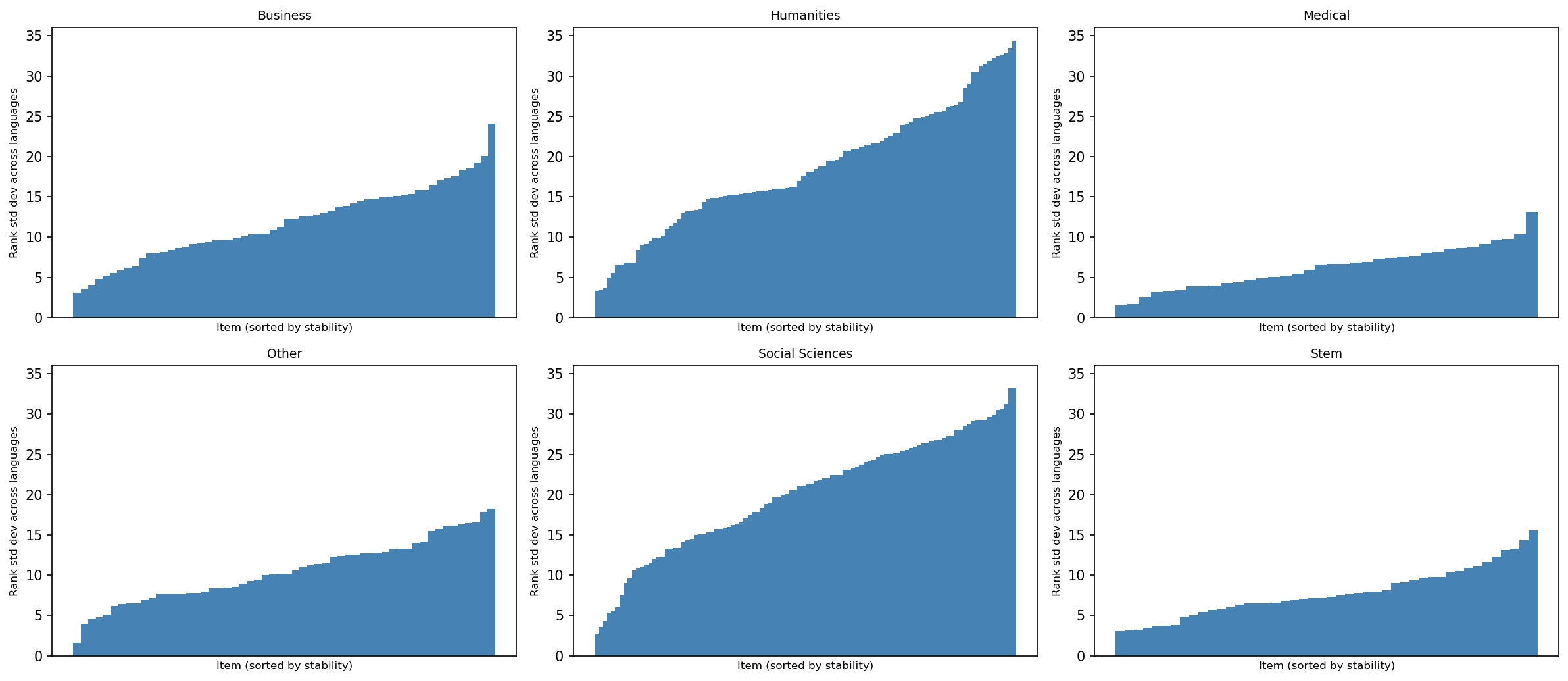}
    \caption{1PL (difficulty)}\label{fig:mmlu_1pl_stab}
  \end{subfigure}\hfill
  \begin{subfigure}[t]{0.32\linewidth}
    \includegraphics[width=\linewidth]{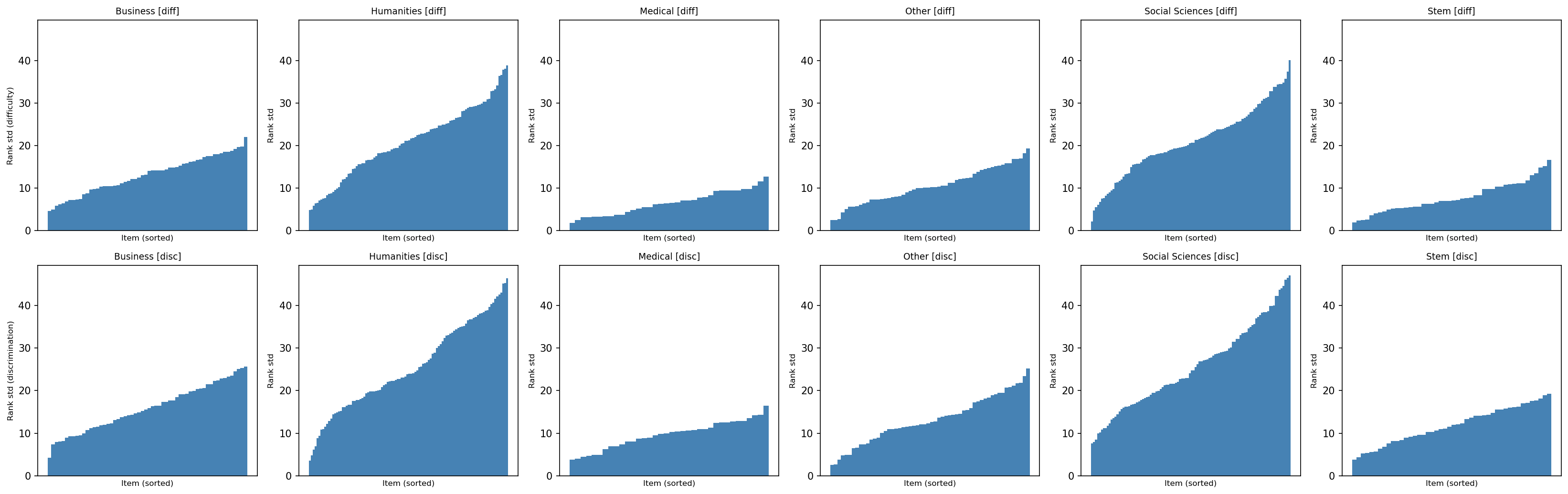}
    \caption{2PL (difficulty + discrimination)}\label{fig:mmlu_2pl_stab}
  \end{subfigure}\hfill
  \begin{subfigure}[t]{0.32\linewidth}
    \includegraphics[width=\linewidth]{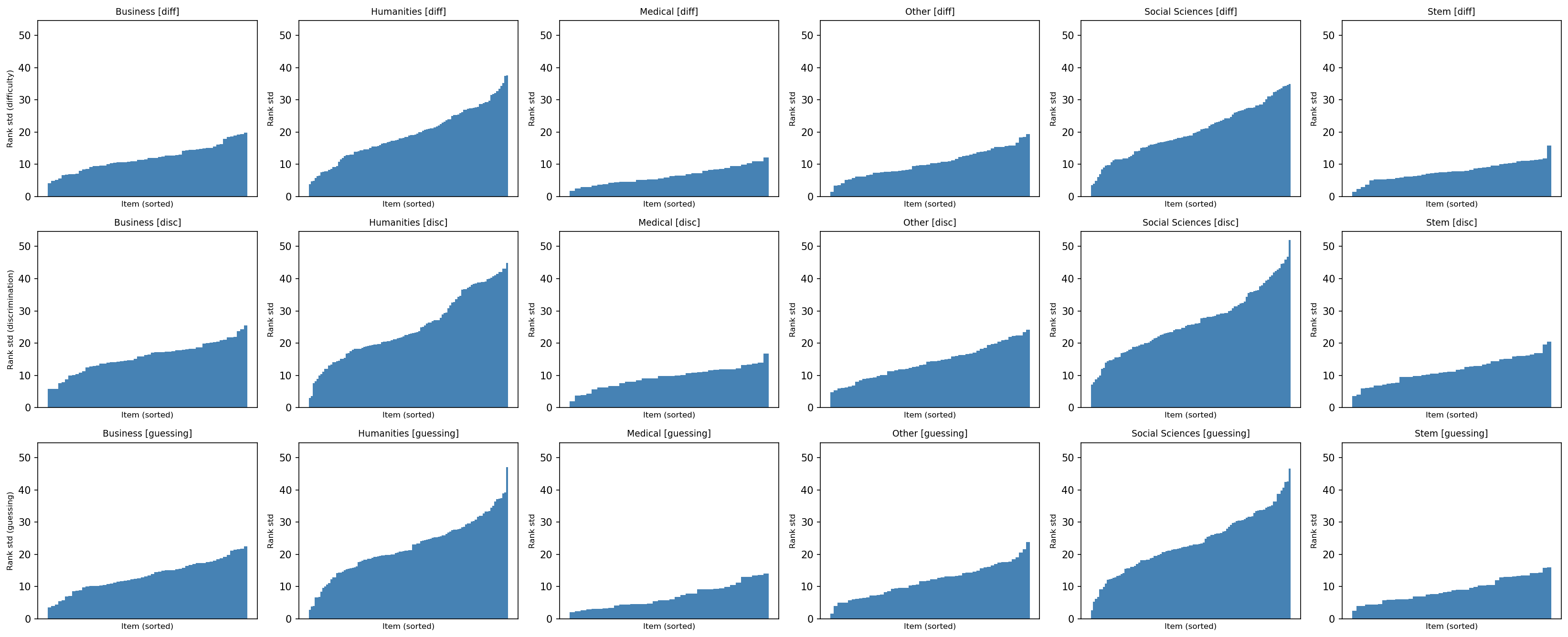}
    \caption{3PL (difficulty + discrimination + guessing)}\label{fig:mmlu_3pl_stab}
  \end{subfigure}
  \caption{MMLU per-item rank standard deviation across 7 languages, pooled
    over six domains ($n \leq 102$ items per domain).}
  \label{fig:mmlu_item_rank_stability}
\end{figure*}

\autoref{fig:gsm_item_rank_stability} and \autoref{fig:mmlu_item_rank_stability} shows per-item rank standard deviation sorted ascending (most to
least consistent left to right). Each bar is one item; bar height is the
standard deviation of its rank position across all languages.
Multi-parameter models show one sub-panel per parameter.

\section{Within-Language Model Fit}
\label{app:withinlangfits}

To assess within-language fit quality for each IRT variant, we report
the MAP negative log-likelihood per observation ($-\ell/\text{obs}$) at convergence,
where lower values indicate better fit.
\autoref{tab:elbo_per_slice} reports these values for every slice.

As expected under correct model fitting (2PL nests 1PL; 3PL nests 2PL), 2PL achieves better fit than 1PL on all 53 slices. 3PL achieves better fit than 2PL on 43/53 slices; on the remaining 10, 3PL's fit is marginally worse (by $0.0001--0.008$ nats/obs), consistent with the added guessing parameter being harder to identify at MAP than genuinely reducing fit.
The improvement is larger on MMLU (mean 1PL $-\ell/\text{obs} = 0.180$ vs 2PL $0.127$ vs 3PL $0.124$) than on MGSM (mean 1PL $0.426$ vs 2PL $0.389$ vs 3PL $0.389$), consistent with MMLU's more heterogeneous item difficulty permitting greater discrimination by 2PL,
and the guessing parameter capturing floor-level response probabilities in knowledge-intensive domains.

\begin{table*}[t]
  \centering
  \small
  \begin{tabular}{llccc}
    \toprule
    \textbf{Benchmark} & \textbf{Slice} & \textbf{1PL} & \textbf{2PL} & \textbf{3PL} \\
    \midrule
    MGSM & \texttt{bn} & 0.360 & 0.311 & 0.313 \\
         & \texttt{de} & 0.453 & 0.416 & 0.415 \\
         & \texttt{en} & 0.420 & 0.383 & 0.381 \\
         & \texttt{es} & 0.492 & 0.464 & 0.463 \\
         & \texttt{fr} & 0.454 & 0.420 & 0.420 \\
         & \texttt{ja} & 0.466 & 0.436 & 0.435 \\
         & \texttt{ru} & 0.479 & 0.445 & 0.442 \\
         & \texttt{sw} & 0.405 & 0.366 & 0.367 \\
         & \texttt{te} & 0.284 & 0.236 & 0.244 \\
         & \texttt{th} & 0.445 & 0.412 & 0.412 \\
         & \texttt{zh} & 0.427 & 0.391 & 0.391 \\
    \midrule
    MMLU  & \texttt{de}\_business        & 0.176 & 0.137 & 0.138 \\
          & \texttt{de}\_humanities      & 0.139 & 0.104 & 0.104 \\
          & \texttt{de}\_medical         & 0.122 & 0.077 & 0.077 \\
          & \texttt{de}\_other           & 0.150 & 0.102 & 0.101 \\
          & \texttt{de}\_social\_sciences & 0.166 & 0.132 & 0.129 \\
          & \texttt{de}\_stem            & 0.222 & 0.151 & 0.146 \\
          & \texttt{en}\_business        & 0.108 & 0.066 & 0.062 \\
          & \texttt{en}\_humanities      & 0.112 & 0.078 & 0.081 \\
          & \texttt{en}\_medical         & 0.113 & 0.092 & 0.095 \\
          & \texttt{en}\_other           & 0.155 & 0.122 & 0.119 \\
          & \texttt{en}\_social\_sciences & 0.151 & 0.113 & 0.112 \\
          & \texttt{en}\_stem            & 0.178 & 0.125 & 0.122 \\
          & \texttt{es}\_business        & 0.163 & 0.124 & 0.119 \\
          & \texttt{es}\_humanities      & 0.126 & 0.085 & 0.088 \\
          & \texttt{es}\_medical         & 0.153 & 0.088 & 0.078 \\
          & \texttt{es}\_other           & 0.186 & 0.137 & 0.133 \\
          & \texttt{es}\_social\_sciences & 0.137 & 0.109 & 0.106 \\
          & \texttt{es}\_stem            & 0.213 & 0.151 & 0.145 \\
          & \texttt{fr}\_business        & 0.174 & 0.134 & 0.135 \\
          & \texttt{fr}\_humanities      & 0.123 & 0.093 & 0.093 \\
          & \texttt{fr}\_medical         & 0.141 & 0.091 & 0.095 \\
          & \texttt{fr}\_other           & 0.175 & 0.124 & 0.117 \\
          & \texttt{fr}\_social\_sciences & 0.146 & 0.113 & 0.111 \\
          & \texttt{fr}\_stem            & 0.231 & 0.170 & 0.165 \\
          & \texttt{ja}\_business        & 0.162 & 0.086 & 0.077 \\
          & \texttt{ja}\_humanities      & 0.140 & 0.095 & 0.093 \\
          & \texttt{ja}\_medical         & 0.158 & 0.086 & 0.083 \\
          & \texttt{ja}\_other           & 0.182 & 0.128 & 0.120 \\
          & \texttt{ja}\_social\_sciences & 0.184 & 0.137 & 0.134 \\
          & \texttt{ja}\_stem            & 0.201 & 0.121 & 0.114 \\
          & \texttt{sw}\_business        & 0.332 & 0.260 & 0.255 \\
          & \texttt{sw}\_humanities      & 0.319 & 0.246 & 0.241 \\
          & \texttt{sw}\_medical         & 0.251 & 0.161 & 0.155 \\
          & \texttt{sw}\_other           & 0.294 & 0.233 & 0.231 \\
          & \texttt{sw}\_social\_sciences & 0.337 & 0.258 & 0.251 \\
          & \texttt{sw}\_stem            & 0.305 & 0.224 & 0.219 \\
          & \texttt{zh}\_business        & 0.137 & 0.079 & 0.078 \\
          & \texttt{zh}\_humanities      & 0.136 & 0.095 & 0.090 \\
          & \texttt{zh}\_medical         & 0.125 & 0.077 & 0.071 \\
          & \texttt{zh}\_other           & 0.198 & 0.132 & 0.123 \\
          & \texttt{zh}\_social\_sciences & 0.140 & 0.105 & 0.103 \\
          & \texttt{zh}\_stem            & 0.178 & 0.113 & 0.109 \\
    \bottomrule
  \end{tabular}
  \caption{MAP negative log-likelihood per observation ($-\ell/\text{obs}$) for each
    IRT model variant and benchmark slice. Lower values indicate better within-language fit.}
  \label{tab:elbo_per_slice}
\end{table*}

\section{IIF Theta Sensitivity}
\label{app:theta}

The main analysis evaluates IIF at $\theta = 0$ (median ability). To assess
sensitivity to this choice, we recompute item selections at $\theta \in \{-1,
0, +1\}$ and measure the Jaccard similarity of the resulting top-50\% subsets
across theta values within each model, and cross-model Jaccard at each theta.

\paragraph{Within-model sensitivity.}
Under MAP, 2PL item selection shows meaningful theta sensitivity on both benchmarks:
MGSM 2PL mean $J = 0.639$ between $\theta=-1$ and $\theta=0$; $J = 0.429$ between
$\theta=0$ and $\theta=+1$; $J = 0.232$ between $\theta=-1$ and $\theta=+1$.
This reflects that properly-identified discrimination parameters vary across items,
so IIF rankings depend on theta. 3PL shows similar sensitivity (MGSM: $J = 0.617$,
$0.400$, $0.208$ for the same pairs).
1PL shows moderate theta sensitivity on MGSM (mean $J = 0.727$ between
$\theta=-1$ and $\theta=0$; $J = 0.835$ between $\theta=0$ and $\theta=+1$)
but near-invariance on MMLU (mean $J \geq 0.929$). The greater sensitivity on
MGSM reflects the wider difficulty range of arithmetic items relative to the
narrower MMLU domain slices.

\paragraph{Cross-model concordance across theta.}
\autoref{tab:theta_sensitivity} shows median Jaccard for each model pair at
each $\theta$. The qualitative ordering is preserved across all theta values:
1PL--2PL and 1PL--3PL agreement both remain above the random baseline ($J \approx 0.33$)
at every $\theta$, and 2PL--3PL agreement is consistently high ($J \geq 0.87$).
The conclusion that 2PL and 3PL select overlapping but not identical subsets
relative to 1PL therefore holds across all evaluated ability levels.

\begin{table}[h]
  \centering
  \small
  \begin{tabular}{llcc}
    \toprule
    \textbf{Pair} & $\theta$ & \textbf{MGSM} & \textbf{MMLU} \\
                  &          & median $J$    & median $J$ \\
    \midrule
    1PL vs 2PL & $-1$ & 0.553 & 0.268 \\
               &  $0$ & 0.404 & 0.528 \\
               & $+1$ & 0.506 & 0.672 \\
    \midrule
    1PL vs 3PL & $-1$ & 0.524 & 0.273 \\
               &  $0$ & 0.412 & 0.454 \\
               & $+1$ & 0.543 & 0.672 \\
    \midrule
    2PL vs 3PL & $-1$ & 0.923 & 0.688 \\
               &  $0$ & 0.923 & 0.590 \\
               & $+1$ & 0.866 & 0.895 \\
    \bottomrule
  \end{tabular}
  \caption{Median Jaccard similarity of top-50\% IIF-selected subsets across
    model pairs, evaluated at three theta values. The random-selection baseline
    is $J \approx 0.33$. Qualitative orderings are stable across all theta values.}
  \label{tab:theta_sensitivity}
\end{table}

\section{Compression Ratio Sensitivity}
\label{app:compression}

The main analysis uses a top-50\% IIF cutoff. To assess sensitivity to this
choice, \autoref{tab:compression} reports in-sample mean Spearman $\rho$
(subset vs.\ full benchmark) at top-30\%, top-50\%, and top-70\% for each
IRT variant and benchmark.

The qualitative ordering is preserved across all ratios. 1PL is reliable at
50\% and above on both benchmarks; at 30\%, fidelity dips slightly (MMLU mean $\rho = 0.931$).
2PL is competitive on MGSM at all compression levels (mean $\rho \geq 0.939$), but shows
substantially lower fidelity on MMLU at all thresholds (mean $\rho = 0.771$--$0.875$),
remaining below 1PL. 3PL follows a similar pattern, with MMLU fidelity
lower than 2PL and 1PL at all ratios (mean $\rho = 0.674$--$0.721$).
The 50\% default is the point where 1PL clears the random baseline
reliably across all slices while achieving meaningful compression.

\begin{table}[h]
  \centering
  \small
  \begin{tabular}{llcc}
    \toprule
    \textbf{Model} & \textbf{Top-$k$} & \textbf{MGSM} & \textbf{MMLU} \\
                   &                  & mean $\rho$ & mean $\rho$ \\
    \midrule
    1PL & 30\% & 0.939 & 0.931 \\
        & 50\% & 0.977 & 0.983 \\
        & 70\% & 0.994 & 0.999 \\
    \midrule
    2PL & 30\% & 0.939 & 0.771 \\
        & 50\% & 0.963 & 0.844 \\
        & 70\% & 0.974 & 0.875 \\
    \midrule
    3PL & 30\% & 0.939 & 0.674 \\
        & 50\% & 0.961 & 0.721 \\
        & 70\% & 0.975 & 0.717 \\
    \bottomrule
  \end{tabular}
  \caption{In-sample ranking fidelity (mean Spearman $\rho$, subset
    vs.\ full benchmark) at three compression ratios. The random-selection
    baseline is $\rho = 0.96$ for MGSM and $= 0.70$ for MMLU.
    1PL conclusions are stable across all ratios; 2PL and 3PL fail on MMLU
    at every threshold.}
  \label{tab:compression}
\end{table}

\section{Subset Ranking Fidelity}
\label{app:ranking_fidelity}

\begin{figure*}[t]
  \centering
  \includegraphics[width=\linewidth]{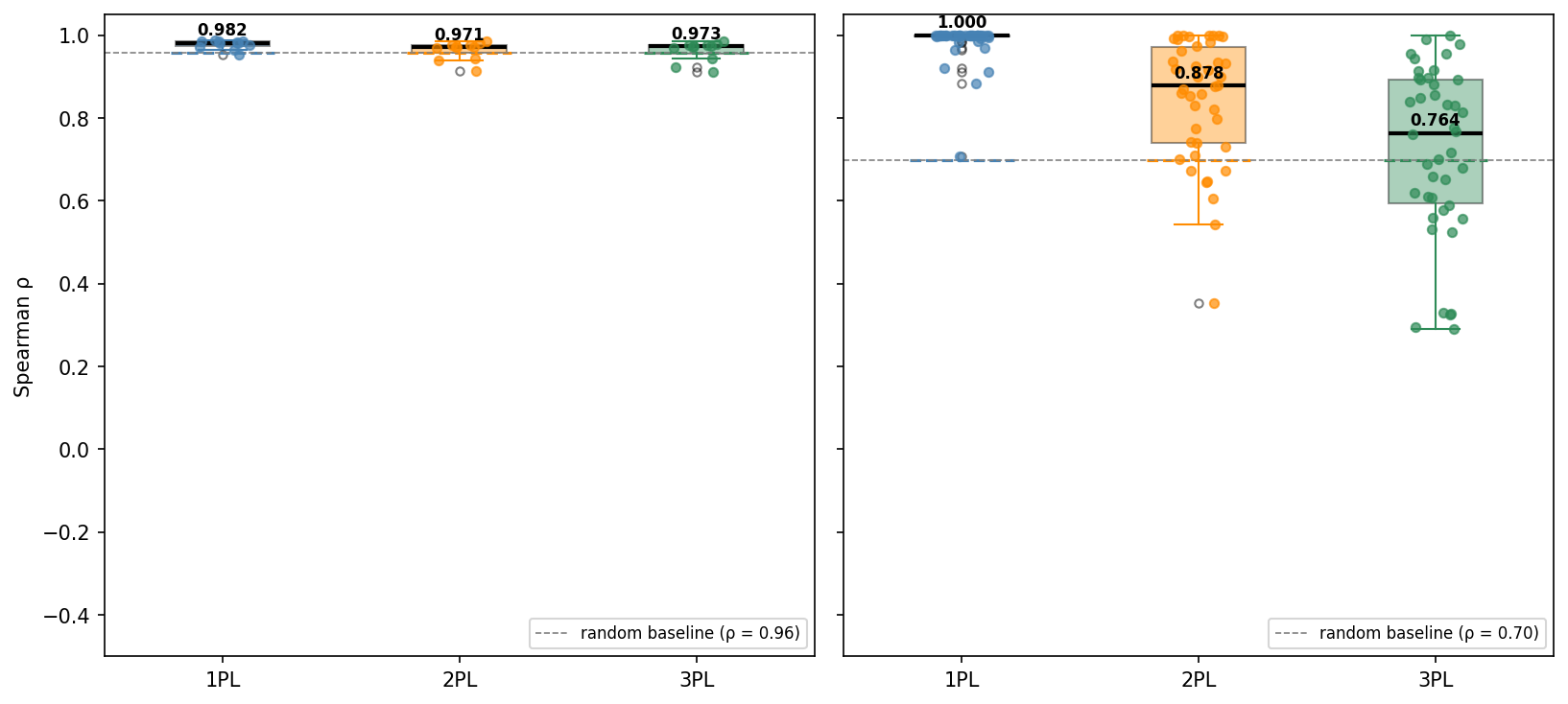}
  \caption{Distribution of Spearman $\rho$ (subset vs.\ full benchmark) across all
    in-sample benchmark slices: 11 MGSM language slices (left) and 42 MMLU
    language$\times$domain slices (right).
    Each dot is one slice; annotated values are medians.
    Dashed lines mark for the random-item baseline ($\rho \approx 0.96$ for MGSM, $\approx 0.70$ for MMLU).}
  \label{fig:same_lang_boxplot}
\end{figure*}

\begin{figure*}[t]
  \centering
  \includegraphics[width=\linewidth]{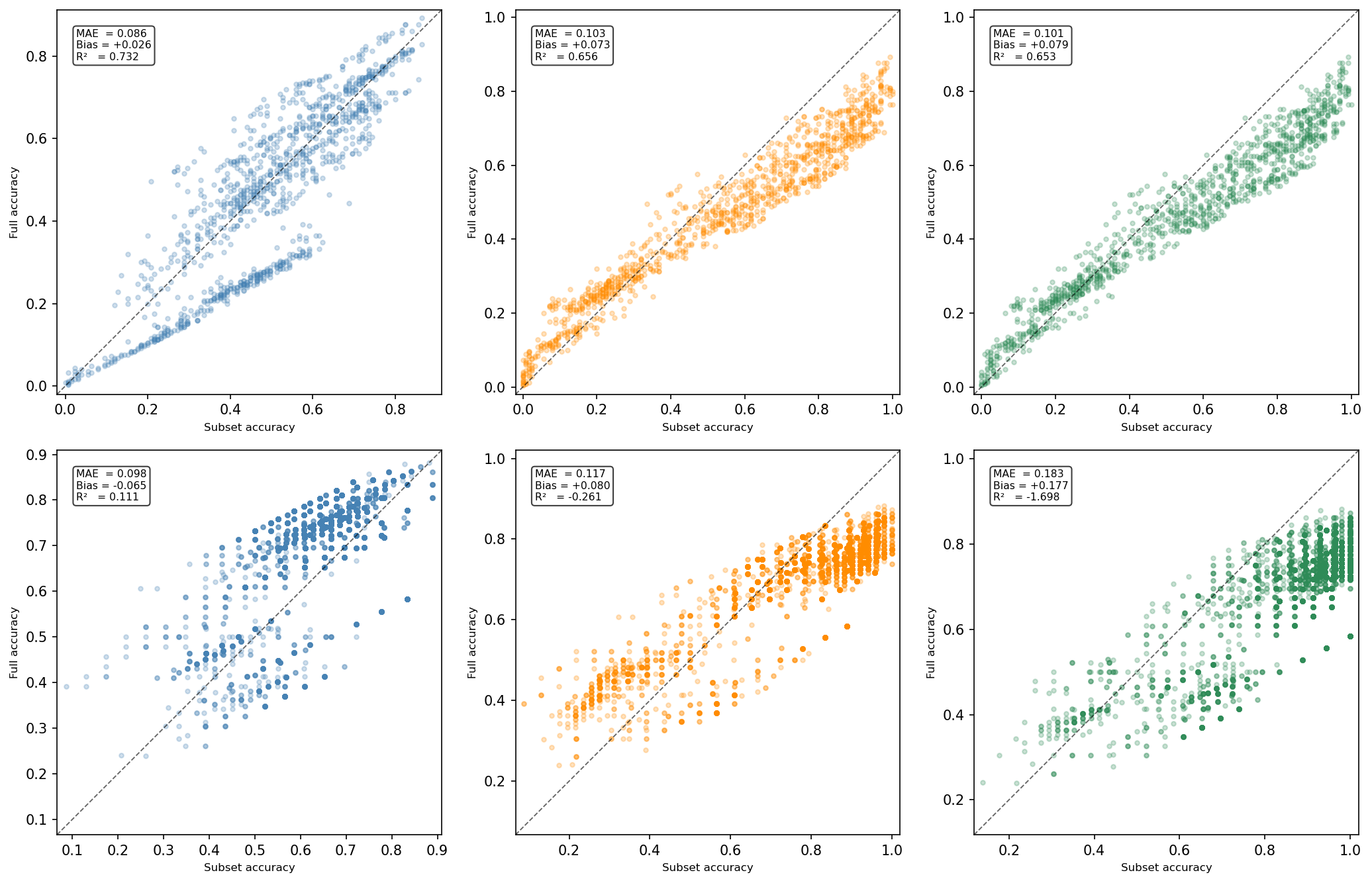}
  \caption{In-sample subset accuracy vs.\ full-benchmark accuracy for each
    (model, benchmark-slice) pair. Rows: MGSM (top) and MMLU (bottom).
    Columns: 1PL, 2PL, 3PL.
    Dashed line is $y = x$ (ideal); each panel is annotated with mean absolute error
    (MAE), Spearman $\rho$, and mean signed bias (subset $-$ full).
    1PL points track the diagonal tightly with low bias; 2PL and 3PL scatter widely
    and show systematic overestimation on MMLU.}
  \label{fig:insample_scatter}
\end{figure*}

\begin{figure*}[t]
  \centering
  \includegraphics[width=\linewidth]{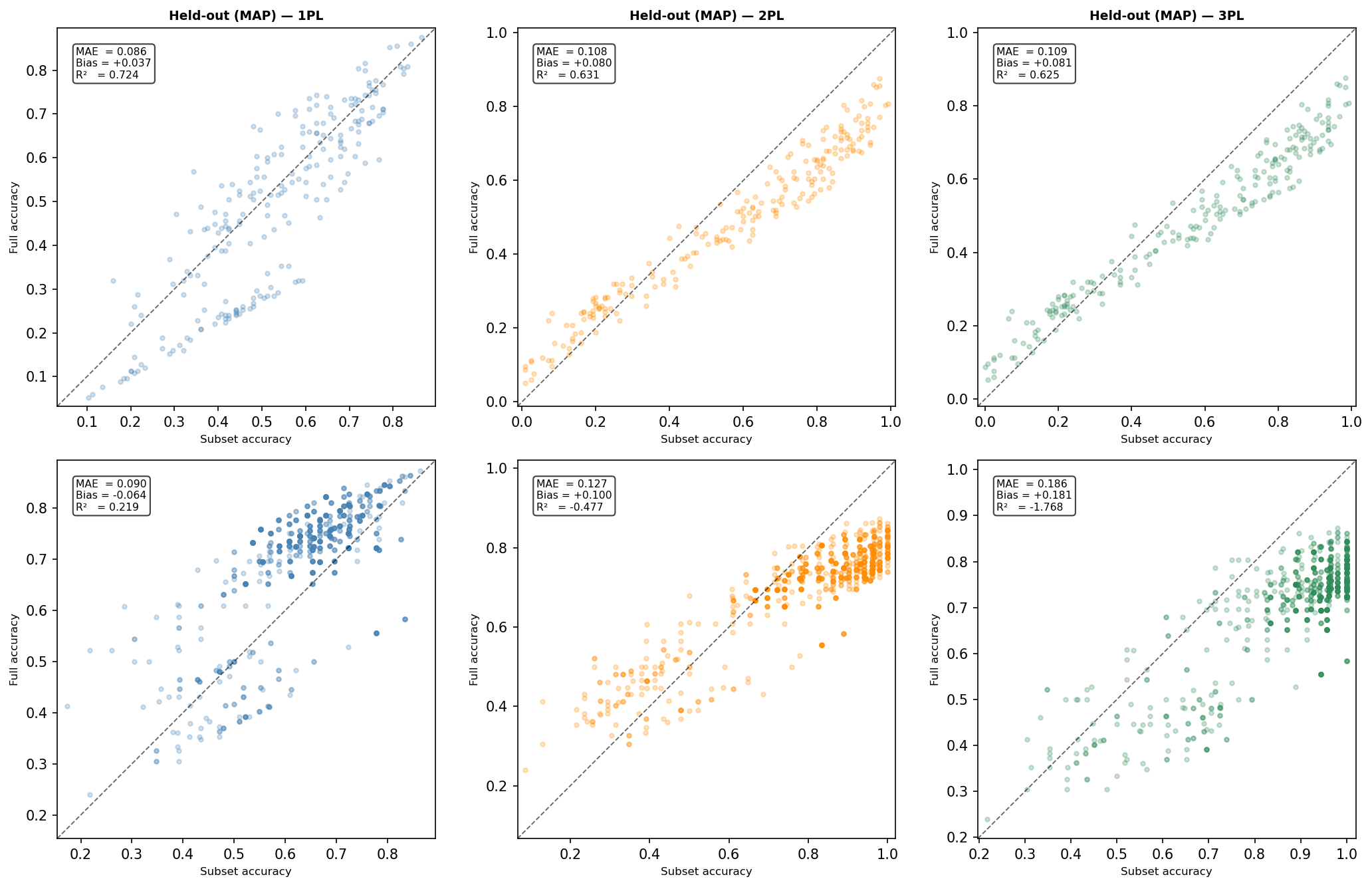}
  \caption{Held-out subset accuracy vs.\ full-benchmark accuracy.
    IRT re-fitted on 80\% of models; each point is one (held-out model,
    benchmark-slice) pair ($n = 220$ for MGSM; $n = 876$ for MMLU).
    Layout and annotation as in Figure~\ref{fig:insample_scatter}.
    1PL MAE and bias are essentially unchanged from in-sample;
    2PL and 3PL retain their large positive bias on MMLU.}
  \label{fig:heldout_scatter}
\end{figure*}

Here, we provide: per-slice fidelity across all benchmark variants (Figure~\ref{fig:same_lang_boxplot}), and per-model accuracy-prediction scatter for in-sample (Figure~\ref{fig:insample_scatter}) and held-out (Figure~\ref{fig:heldout_scatter}) evaluations.

\paragraph{Per-slice Fidelity (Figure~\ref{fig:same_lang_boxplot}).}
Each dot is one benchmark slice — 11 MGSM language slices and 42 MMLU
language$\times$domain slices.
For MGSM, 1PL is uniformly strong across all 11 languages (median $\rho = 0.982$,
IQR entirely above $\rho = 0.95$), with no language slice falling below the
random baseline. 2PL is similarly strong (median $\rho = 0.971$, IQR $0.957$--$0.976$),
with the minimum at $\rho = 0.915$, well above the random baseline; no slice fails badly.
3PL is comparable (median $\rho = 0.973$, IQR $0.955$--$0.976$, min $0.911$).
For MMLU, the contrast between 1PL and the multi-parameter models sharpens considerably.
1PL achieves median $\rho = 1.000$ — the majority of the 42 slices reach perfect
rank correlation — with all slices well above the random baseline ($\rho \approx 0.70$).
2PL falls to median $\rho = 0.878$ (IQR $0.740$--$0.970$, min $0.354$): most MMLU slices
retain useful fidelity, but the lower tail extends below the random baseline on a minority of
slices.
3PL is further reduced (median $\rho = 0.764$, IQR $0.594$--$0.892$, min $0.290$),
with more slices falling below the reference line than under 2PL.

\paragraph{In-sample accuracy prediction (Figure~\ref{fig:insample_scatter}).}
Each point is one (model, benchmark-slice) pair ($n = 1{,}122$ for MGSM;
$n = 4{,}320$ for MMLU).
The dashed $y = x$ line marks perfect accuracy prediction; deviations from it
indicate systematic bias in how the subset estimates a model's full-benchmark score.
Under 1PL, MGSM points cluster tightly near the diagonal
(MAE $= 0.086$, bias $= +0.026$): the subset very slightly overestimates
full accuracy, consistent with IIF at $\theta = 0$ selecting items near median
difficulty, which form a marginally easier sub-pool than the benchmark as a whole.
Under 1PL, MMLU shows the opposite bias (MAE $= 0.098$, bias $= -0.065$):
the IIF-selected items are slightly harder than the MMLU average because
IIF peaks for items with difficulty $b \approx 0$, which are harder than
the typical MMLU item (most items are easier than chance for these models).
Consequently, subset accuracy understates full-benchmark accuracy by about
six percentage points on average.
Under 2PL and 3PL, MGSM accuracy prediction is close to 1PL
(MGSM 2PL MAE $= 0.103$, bias $= +0.073$; 3PL MAE $= 0.101$, bias $= +0.079$).
On MMLU, 2PL IIF introduces a moderate positive bias ($+0.080$, MAE $= 0.117$): the
discrimination-weighted selection still over-selects items where models score high,
inflating subset accuracy estimates relative to the full benchmark by about 8 percentage points.

\paragraph{Held-out accuracy prediction (Figure~\ref{fig:heldout_scatter}).}
The held-out evaluation re-fits IRT on 80\% of models and measures accuracy
prediction on the remaining 20\%.
For 1PL, the held-out MAE is nearly identical to in-sample on both benchmarks
(MGSM: $0.086$ held-out vs.\ $0.086$ in-sample; MMLU: $0.090$ vs.\ $0.098$),
and the directional bias is unchanged, confirming that 1PL item selection
generalizes to unseen models without any degradation in accuracy-prediction quality.
For MGSM 2PL, held-out MAE ($0.108$) is comparable to in-sample ($0.103$ in-sample is
for all models; held-out evaluates on the 20\% unseen split), reflecting that MAP
discrimination estimates generalize reasonably to unseen models within MGSM.
For MMLU, 2PL and 3PL retain moderate-to-large positive bias ($+0.100$ and $+0.181$
respectively) in held-out evaluation, confirming that the positive accuracy bias
is a structural property of discrimination-weighted item selection on knowledge-intensive
benchmarks rather than an in-sample artifact.

\end{document}